\documentclass[10pt,journal,cspaper,compsoc]{IEEEtran}
\ifCLASSOPTIONcompsoc
\usepackage[compress]{cite}
\else
\usepackage{cite}
\fi
\ifCLASSINFOpdf
\usepackage[pdftex]{graphicx}
\usepackage{amssymb}
\usepackage{rotating}
\else
\fi
\usepackage[cmex10]{amsmath}

\usepackage[font={small}]{caption}
\begin{document}
\title{Template-Based Active Contours}
%
%
%

\author{Jayanth~Krishna~Mogali$^{1}$,
         Adithya~Kumar~Pediredla$^{1}$, and
         Chandra~Sekhar~Seelamantula$^{2}$,~\IEEEmembership{Senior Member,~IEEE}
\IEEEcompsocitemizethanks{
\IEEEcompsocthanksitem $^{1}$Both J.~K.~Mogali and A.~K.~Pediredla contributed equally to this paper 
\IEEEcompsocthanksitem $^{1, 2}$The authors are with the Department of Electrical Engineering, Indian Institute of Science, Bangalore - 560 012, Karnataka, India. Email: chandra.sekhar@ieee.org. Phone: +91-80-2293-2695; Fax: +91-80-2360-0444.
}

\thanks{}
}

\IEEEcompsoctitleabstractindextext{
\begin{abstract}
We develop a generalized active contour formalism for image segmentation based on shape templates. The shape template is subjected to a restricted affine transformation (RAT) in order to segment the object of interest. RAT allows for translation, rotation, and scaling, which give a total of five degrees of freedom. The proposed active contour comprises an inner and outer contour pair, which are closed and concentric. The active contour energy is a contrast function defined based on the intensities of pixels that lie inside the inner contour and those that lie in the annulus between the inner and outer contours. We show that the contrast energy functional is optimal under certain conditions. The optimal RAT parameters are computed by maximizing the contrast function using a gradient descent optimizer. We show that the calculations are made efficient through use of Green's theorem. The proposed formalism is capable of handling a variety of shapes because for a chosen template, optimization is carried with respect to the RAT parameters only. The proposed formalism is validated on multiple images to show robustness to Gaussian and Poisson noise, to initialization, and to partial loss of structure in the object to be segmented. 
\end{abstract}
}

\maketitle
\IEEEpeerreviewmaketitle
\section{Introduction}
\IEEEPARstart{T}{he} problem of segmentation in computer-aided image analysis relates to outlining the boundaries of the object of interest. The optimal boundary/contour cannot often be solved in closed form. Instead, it is solved for iteratively using optimization techniques. The contour evolves from a user-specified initialization, and converges to the object boundaries. Such contours are referred as {\it active contours} or {\it snakes}, a term coined by Kass et al. \cite{Kass}.\\
\indent Snakes are evolutionary contours that move under the influence of certain forces: gradients or functions derived therefrom, which direct the contour to the boundary of interest --- internal forces, which ensure certain degree of smoothness or regularity of the curve, these forces are referred as {\it Snake energies}. The novelty in snake design lies in the specification of suitable energies and in representation of the curve \cite{Staib,GVF,Cootes,Pentland,Blake,Brigger}. 
The key challenges in segmentation are low signal-to-noise ratio, occlusion, non-uniformity of regional intensities, broken or diffuse boundaries or a combination of them.
To mitigate these problems, a common approach is to incorporate prior shape information relating to the object of interest in the optimization framework \cite{tsai,Paragois,Levelsetprior, Levelprior,Cremerslevelprior}. The effectiveness of incorporating shape priors finds application in medical image segmentation \cite{tsai,Paragois,snakuscule,ovuscule,Adithya,Adithya1}, where the objects of interest exhibit a certain degree of shape specificity. 
\subsection{Prior Art}
Cootes et al. \cite{Cootes} proposed a method to incorporate shape-prior information in the form of a model derived from a set of training samples, which is assumed to reliably capture variations in the shape of the object of interest. However, this technique relies on image training data, which may be expensive, difficult, and in some cases not reliable. In the absence of such data, segmentation using a single shape prior is an alternate option. We briefly review previous literature on segmentation using a single shape prior, follow by our approach.

\indent Techniques on single shape prior based segmentation employ regularization energies in addition to image-derived energies for optimization \cite{Levelsetprior, Levelprior,Cremerslevelprior,Paragois,Cremers,Werlberger,snakuscule,ovuscule}. Regularization energies penalize segmenting curve for deviating from the geometric transformation space (GTS) of the reference curve, thereby imposing a shape constraint. The regularization functional is a shape distance metric (SDM) that measures the deviation in shape between the segmenting curve and the reference. Optimization within this regularization framework is carried out iteratively and in two stages  \cite{Levelsetprior, Levelprior,Cremerslevelprior,Cremers,Werlberger}. In the first stage, the curve is evolved to optimize the image derived energies, while in the second stage, the curve is modified to satisfy the shape constraint by minimizing the SDM between the evolving curve and the reference in the GTS space governing the reference curve. As gradient descent is employed for minimization, the local optimality of the solution outputted by the algorithm may not satisfy the shape constraint.
Th\'evenaz et al. proposed a technique that mitigates the two stage approach with in the regularization framework, for segmenting circular \cite{snakuscule} and elliptical blobs \cite{ovuscule}. The active contour comprises two concentric circles (snakuscule) or ellipses (ovuscule) as a shape template for segmentation. In both cases, the snake energy is defined as the contrast between the pixel intensities in the inner contour and those lying in the annular region between the inner and outer contours. Snakuscules are parametrized by a pair of diametrically opposite points, which are constrained to always possess the same ordinate by means of a regularization functional. Ovusucules are specified by three points on the inner ellipse. However, the shape specific parameterization and regularization employed is not easily generalizable to any shape. A regularization-free approach was proposed by Chen et al. \cite{Chen}. They combined image registration techniques with segmentation. This algorithm exhaustively search for optimal similarity transform (ST) parameters globally, to match reference shape and object of interest. The technique is iterative and each iteration consists of two steps. In the first step, the translation parameters are updated following a global search in the image pixel coordinates. 
In the second step, rotation and scaling parameters are updated using similar global search in the log polar representation of the image. The computations for the search are made efficient by the use of fast Fourier transform (FFT). \\
\indent In \cite{Adithya}, we extended the formalism of Th\'evenaz et al. \cite{snakuscule,ovuscule} to concentric rectangular templates for segmentation of gel electrophoresis images. In \cite{Adithya1}, we proposed a unified approach for designing snakuscules and ovuscules from a pair of concentric circles and optimizing for a restricted affine transformation (RAT). RAT has five degrees of freedom allowing for different scaling along horizontal and vertical directions, rotation, and translation. This unified approach is simpler to formulate and optimize, while retaining the shape specificity.
\subsection{This Paper}
\indent In this paper, we consider a template based active contour formalism for segmentation. Specifically, we extend the formalism in \cite{snakuscule,ovuscule,Adithya,Adithya1} to a generic shape template, which may even be non-convex. The salient features of our formulation are:
\begin{enumerate}
\item{} The mathematical formalism is shape-independent, devoid of explicit regularization energies and requires calculation of a fixed five parameters irrespective of the shape.
\item{} Using Green's theorem, we convert region integrals into contour integrals. As contour integrals require less number of computations compared to region integrals, this step will result in computational savings.
\item We will show through experiments that proposed active contour formalism is fast enough to enable real time deployment.
\end{enumerate}  
We retained the {\it nested structure} in \cite{snakuscule,ovuscule} for specifying the shape template. The inner contour of the snake lies completely inside the outer contour. The user-defined shape template serves as the inner contour while the outer contour is obtained either by scaling the inner contour (for convex shapes) or by a manual specification.  The contours are parametrized using B-splines; linear B-splines are used for representing polygonal shapes and cubic B-splines are used for designing smooth closed contours. The snake energy is a normalized contrast function, which measures the difference in average intensities between the inner contour and the annular region between the inner and outer contours. We modify the energy proposed by Th\'evenaz et al. \cite{snakuscule,ovuscule}, by adding a normalization using the areas enclosed by the inner and outer contours to ensure stability of fit and to eliminate explicit regularization. Upon optimization, the inner contour locks on to the object of interest. The outer contour guides the computation of the local contrast metric. The optimization of the snake energy is carried out with respect to the RAT parameters of the shape template. However, the proposed formalism can also be adapted to a generic transformation.\\
\indent The rest of the paper is organized as follows. In Section~\ref{activeshape}, we briefly review some key aspects of B-spline kernel based curve parameterization. We also give a formulation of the active contour and introduce the snake energy. In Section~\ref{Affineopt}, we address the active contour optimization problem. The computational aspects are discussed in Section~\ref{optimization}. In Section~\ref{validation}, we present experimental results on synthesized images, medical, and biological images. Robustness to Gaussian, Poisson noise conditions, and to initialization are also reported.
\section{Active Contour Formulation}
Active contour design involves two major steps: design of the contour, and design of the energy functional. We design the contour using a B-spline based parameterization which have some interesting properties such as non-negativity, compact support, minimum curvature. Shifted cubic B-splines also form Riesz basis. We briefly review some of those properties to justify their usage in the proposed formalism. \\
\indent Usage of B-splines will lead to a case where we can use the parameters of the curve (referred as spline knots) for optimizing the RAT parameters, instead of the curve itself. This will be shown in Section \ref{Affineopt}, when we derive the derivatives of RAT parameters. As we are going to work on the spline coefficients, we review the Riesz basis property satisfied by B-splines, which allows us to operate on the discrete spline coefficients.
\subsection{Riesz basis property}
Consider a signal $f(t)$ in the space spanned by the shifted cubic B-spline. We can write $f(t) = \sum_k c_k\beta(t-k) $, where $\beta$ is a cubic-B-spline. As the shifted cubic B-splines are linearly independent, they form Riesz basis and satisfy the following Riesz basis conditions:
\begin{align}
A\|\mathbf{c}\|^2 \le  \Big\|\sum_k c_k\beta(t-k) \Big\|^2       \le B\|\mathbf{c}\|^2 , \nonumber \\
\mathrm{where}\,\,0<A\,\le\,B\,<\infty
\label{Riesz basis2}
\end{align}
This property ensures that a significant change in the value of coefficients ($\mathbf{c}$) leads to a significant change in the shape of the curve and vice versa.
\label{activeshape}
\subsection{B-spline-based template parametrization}
\label{parameterization}
\begin{figure}[lt]
\centering
\includegraphics[width=2.in]{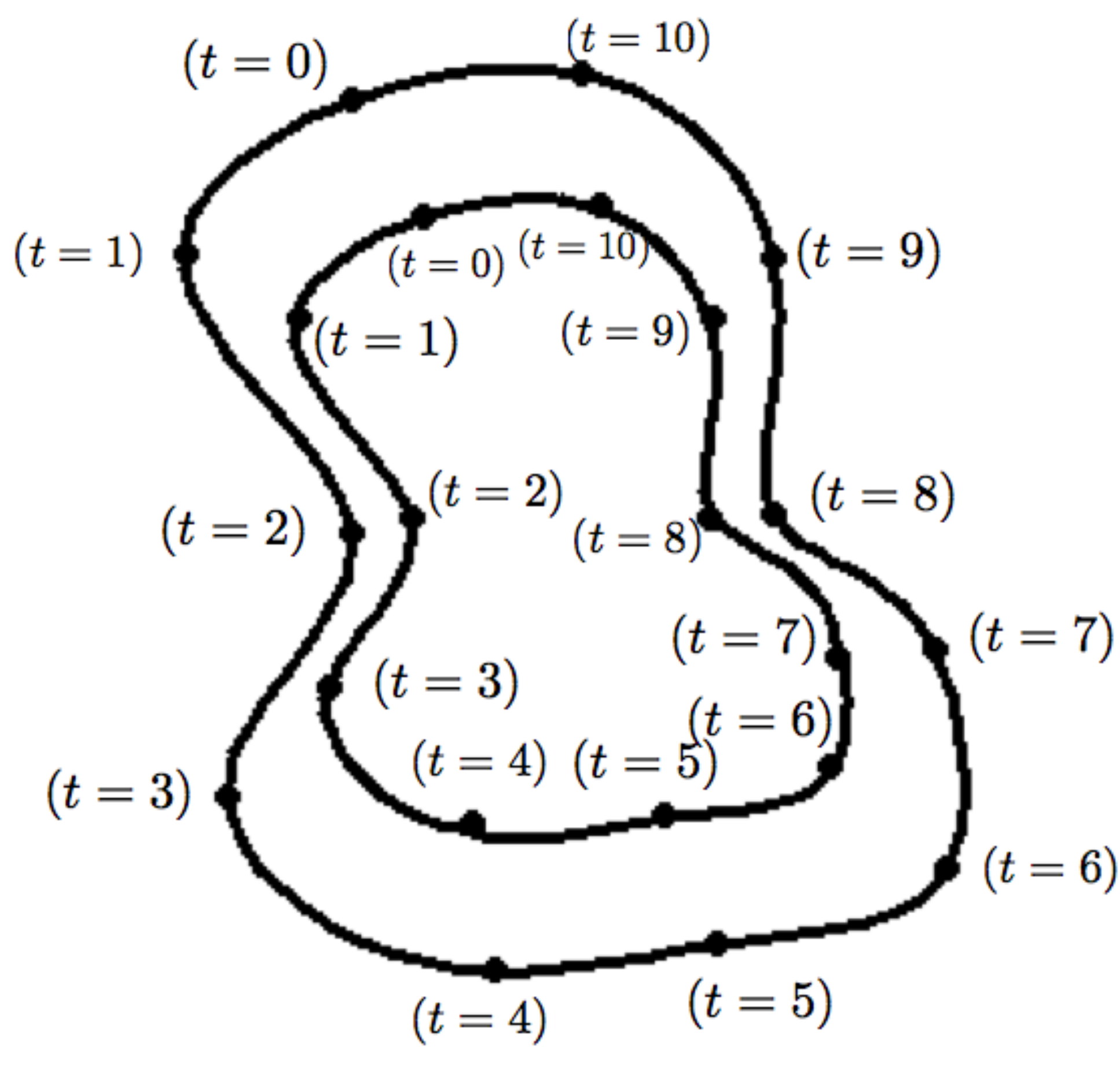}\\
\caption{Example of a shape template. The Inner and outer contours have been parametrized using cubic B-splines with 11 knots.}
\label{Figshapetemplate}
\end{figure}

\indent Consider a pair of inner and outer contours parametrized as $\mathbf{r_0}(t)=(x_0(t),y_0(t))^T$ and $\mathbf{r_1}(t)=(x_1(t),y_1(t))^T$, respectively, where $t$ is the independent variable. The subscripts $0$ and $1$ indicate the inner and outer contours, respectively. The functions $x_0(t)$, $x_1(t)$, $y_0(t)$, and $y_1(t)$, are expressed in terms of a basis function $\beta(t)$ and its integer translates. 
The structure of the parametrization is along the lines of \cite{Brigger, Jacob}, and results in a vector representation for the shape template:
\begin{eqnarray}
\mathbf{r}_i(t)&=&\left(
\begin{array}{c}
x_i(t) \\
y_i(t) \\
\end{array}
\right)
=
\sum\limits_{k=-\infty}^{\infty}\, \mathbf{c}_{i,k} \beta(t-k),
\label{shapetemplate}
\end{eqnarray}
where 
$\mathbf{c}_{i,k}=     
\left(
\begin{array}{c}
c_{x_i,k} \\
c_{y_i,k} \\
\end{array}
\right)$ 
$i=0, 1$, are the coefficient vectors, which are referred to as {\it spline coefficients} in spline literature. The Riesz basis property (\ref{Riesz basis2}) of the B-splines allows us to operate on the knots to control the contour. Closed curves result in periodic $x_i(t)$ and $y_i(t)$, for which we have the equivalent representation:
\begin{eqnarray}
\mathbf{r}_i(t)&=&\sum\limits_{k=0}^{M-1}\, \mathbf{c}_{i, k} \beta_p(t-k),
\label{parametric}
\end{eqnarray}
where $\beta_{p}(t)= \sum\limits_{k=-\infty}^{\infty}\,\beta(t-kM)$ is $M$-periodic; $c_{x_i,k}=c_{x_i,k+M}$, $c_{y_i,k}=c_{y_i,k+M}$, $i=0, 1$, are the periodized coefficient sequences with period $M$ ($M$ being the number of knots).  In Figure~\ref{Figshapetemplate}, we show a shape template designed using cubic B-splines. \\
\indent For convex shapes centered at the origin, the outer contour ($\mathbf{r_0}(t)$) may be chosen as a scaled version of the inner contour ($\mathbf{r_0}(t)$) that is, $\mathbf{r_1}(t) = {\alpha}\,\mathbf{r_0}(t)$, $\alpha$ is a scalar greater than unity. In this case, the coefficients of the outer contour are ${\alpha}$ times those of the inner contour. The parameter $\alpha$ gives direct control over the width of the annular region and determines the region over which the local contrast function is computed. For non-convex shapes, the outer contour must be parametrized independently of the inner contour. 
\subsection{From shape templates to active contours}
\label{affine}
\indent Template based active contours are derived from the shape template in (\ref{shapetemplate}) according to the transformation:
\begin{eqnarray}
\small
\left(\begin{array}{cc} 
X_i(t)  \\
Y_i(t) 
\end{array}\right) 
=
{\left(\begin{array}{ccc}
\label{RATransform}
\quad A\cos\theta & B\sin\theta & x_c \\
-A\sin\theta & B\cos\theta & y_c  
\end{array}\right)}
\left(\begin{array}{ccc}
x_i(t)  \\ 
y_i(t) \\
1
\end{array}\right),
\end{eqnarray}
where $i=0,1$, and $(X_0(t),Y_0(t))$ and $(X_1(t),Y_1(t))$ are the inner and outer contours of the template based active contour respectively. $A$ and $B$ represent the scale parameters along axial and perpendicular directions, respectively; $\theta$ represents the angle of rotation (clockwise), $x_c$ and $y_c$ are the translation parameters.  The nested contour design is general as the outer and inner contours can be chosen differently. 

\subsection{Snake energy specification}
\indent In this section, we first provide a short structural overview of snake energies in general and discuss the snake energies that are relevant to our work. We then propose a new snake energy and establish the connection shared by the proposed formulation with other snake energies.\\
\indent Most snake energies \cite{Kass,Staib,GVF,Cootes,Pentland,Blake,Brigger,Levelprior,Cremerslevelprior,snakuscule,ovuscule,Adithya,Adithya1} can be represented by (\ref{RegionGradient}). The snake energy ($E$) comprises image and contour derived functions. The contour related functions ($E_\mathrm{Regularization}$) are aimed at ensuring stability and smoothness of the evolving contour, while, the image related functions are responsible for driving the snake towards the object boundaries:
\begin{align}
\small
\label{RegionGradient}
\mbox{$E$}=\underbrace{\left(\mbox{$\alpha_1\,E_\mathrm{Region}$}+\mbox{$\alpha_2\,E_\mathrm{Gradient}$}\right)}_{Image\, energy}+\underbrace{\left(\mbox{$\alpha_3\,E_\mathrm{Regularization}$}\right)}_{Contour\,energy}\,.
\end{align}
\indent The image related energies are broadly classified into region ($E_{Region}$) and gradient energies ($E_{Gradient}$). Region energies rely on image statistical measures for segmentation, whereas gradient energies finely segment object boundaries by capturing steep transitions in pixel intensity using derivative operators. Typically, a linear combination of the energies is used for segmentation. Assigning the weight factors ($\alpha_1,\alpha_2,\alpha_3$) for the combination is not straightforward and may require tuning on an image by image basis.\\
\indent Recently, there is growing interest to use only region based energies as the image derived component of the snake energy \cite{Chakraborthy,VeseChan,Chesnaud,Yezzi}. The motivation is due to the fact that gradient energies highlight noise and minor artifacts in the image, thereby degrading the segmentation performance of the snake. Also, it is known that region energy-based snakes are more robust to noise and initialization than gradient energy snakes as they rely on non-local measures. In view of these observations, we favor the use of region based energies in our formulation. We briefly review a few popular region based energies and gradually introduce our snake energy.\\
\indent For an active contour initialized on an image $f$, Chan and Vese \cite{VeseChan} proposed the following energy model, which is a special case of Mumfard-Shah functional\cite{Mumford}:
\begin{eqnarray} 
\label{Chan Vese}
\mbox{$E$}=\underbrace{
\begin{aligned}
\displaystyle \left ( \,\,\iint\limits_{\text{inside(C)}}\right.&{(f-\mu_\mathrm{foreground})^2\,\mathrm{d}x\mathrm{d}y}  \\
\displaystyle +&\left .\iint\limits_{\text{outside(C)}}{(f-\mu_\mathrm{background})^2\,\mathrm{d}x\mathrm{d}y}\right)
\end{aligned}
}_{E_{Fitting\,term}}\nonumber\\
+\underbrace{\left({\lambda_{1}\,\text{Length(C)}}+{\lambda_{2}\,\text{Area(C)}}\right)}_{E_{Reg}}\,.
\end{eqnarray}
$E$ comprises two terms $E_{Fitting\,term}$ and $E_{Reg}$, where $E_{Reg}$ represents curve regularization energies: length and area of the evolving contour C. The region within the evolving contour (C) is considered as foreground and outside the contour as background. $E_{Fitting\,term}$ penalizes the variances of foreground and background regions. $\mu_{\mathrm{foreground}}$ and $\mu_\mathrm{background}$ are the mean intensity values of the foreground and background respectively. Segmentation is achieved based on the assumption that the foreground and background regions are distinguishable based on $\mu_\mathrm{foreground}$ and $\mu_\mathrm{background}$. Yezzi et al. \cite{Yezzi} proposed the mean separation energy ($E_{MS}$) shown in (\ref{mean sep}) for segmentation. Instead of approximating the foreground and background regions with constant pixel intensities, the formulation aims to segment the image by maximally separating their mean intensity values ($\mu_\mathrm{foreground}$ and $\mu_\mathrm{background}$). The formulation includes an arc length regularizer, penalizing the active contour for encircling small noisy regions:
\begin{eqnarray} 
\label{mean sep}
\mbox{$E_{MS}$}=-\frac{1}{2}{(\mu_\mathrm{foreground}-\mu_\mathrm{background})^2}+{\alpha_{1}\,\int_{C}{\mathrm{d}s}} .
\end{eqnarray}
\indent Th\'evenaz et al. proposed active contours to segment locally bright blob like structures. They considered concentric circles\cite{snakuscule} and ellipses\cite{ovuscule}. They used a local contrast function that is similar to the mean separation energy.
\begin{align} 
\label{philip}
\small
\mbox{$E$}=\underbrace{\left(\frac{1}{|\Re_0|}\Bigg\{ {\int\int_{{\Re_1} \backslash {\Re_0}}{f\,\mathrm{d}x\,\,\mathrm{d}y}-\int\int_{\Re_0}{f\,\mathrm{d}x\,\,\mathrm{d}y}} \Bigg\}\right)}_\mathrm{Local\,contrast} \nonumber \\
+\mbox{$J_{Reg}$},
\end{align}

\begin{figure}
\centering
\includegraphics[width=2.in]{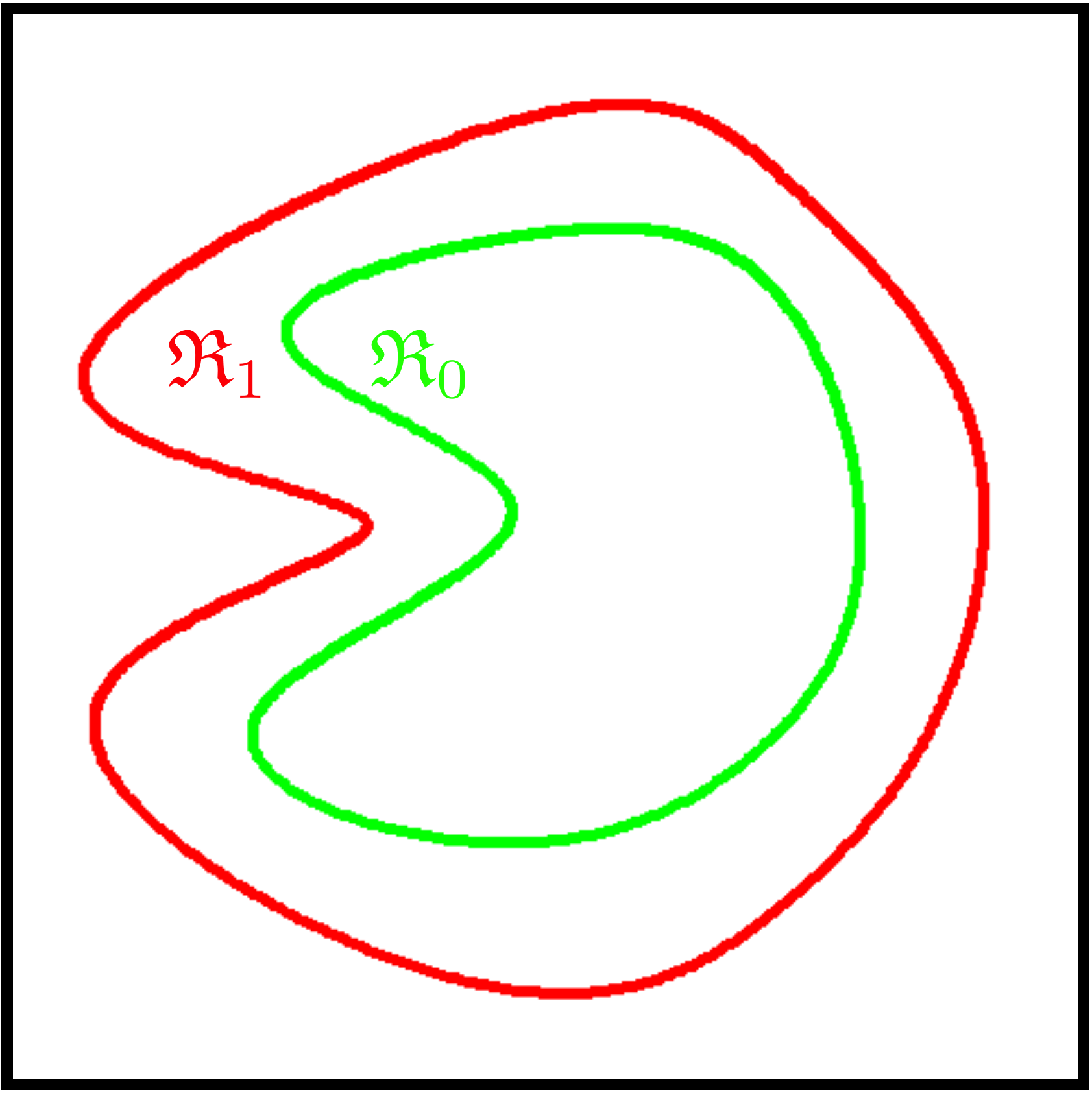}\\
\caption{Figure to illustrate the regions $\Re_0$ and $\Re_1$ of energy functional.}
\label{EnergyFunctionExplanation}
\end{figure}
\begin{figure}[t]
\centering
$
\begin{array}{ccc}
\includegraphics[width=1.0in]{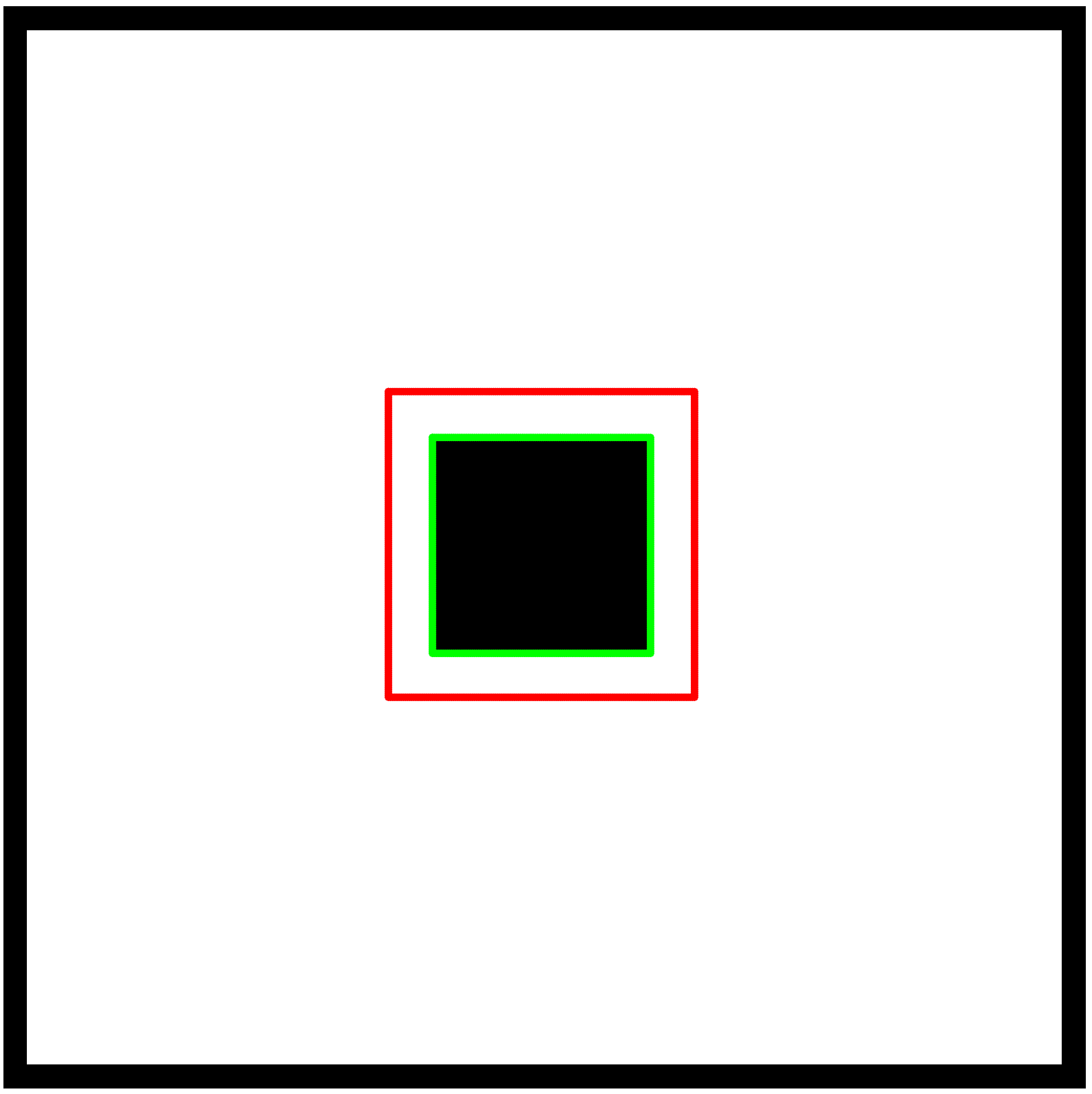} &
\includegraphics[width=1.0in]{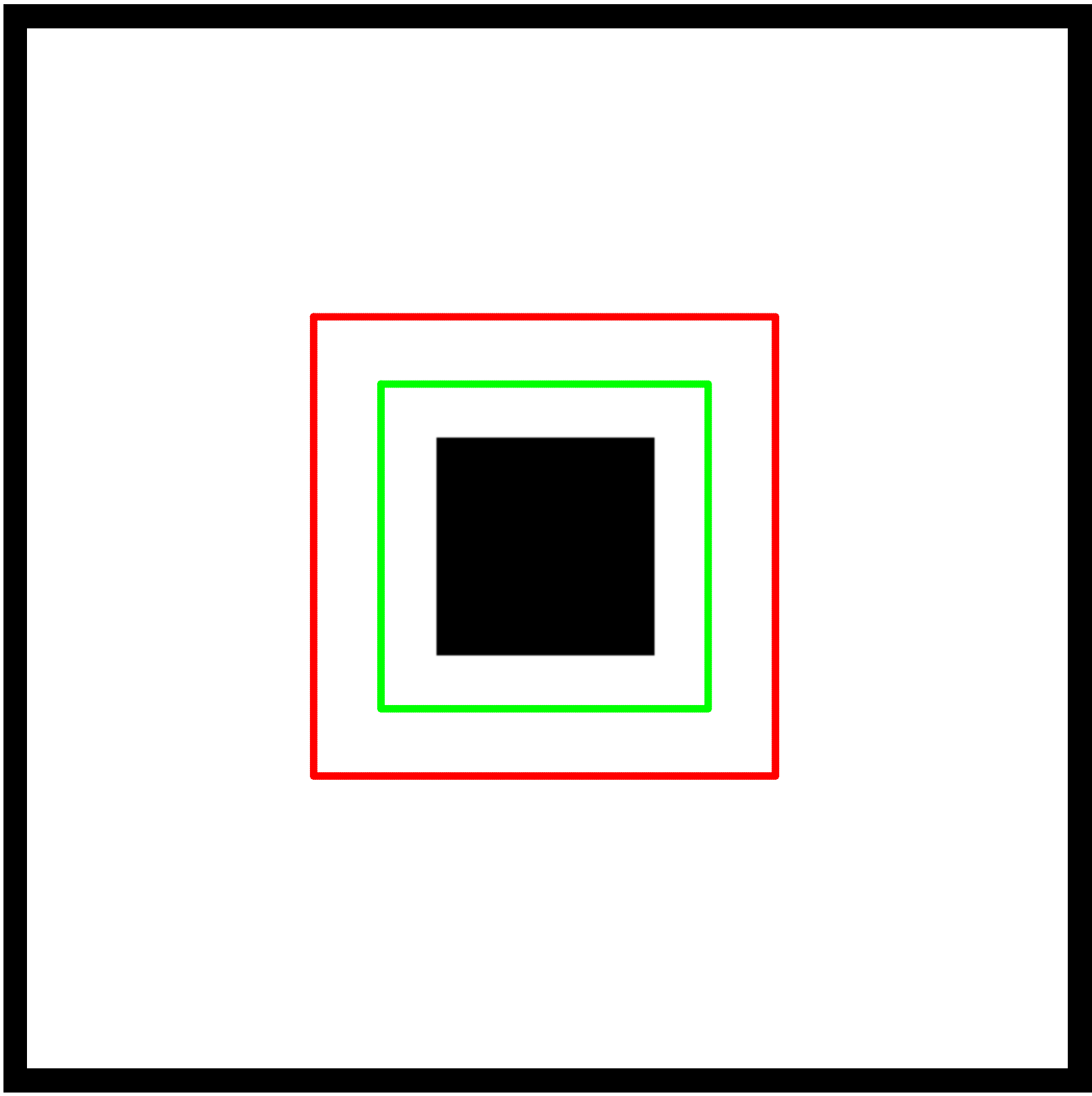} &
\includegraphics[width=1.0in]{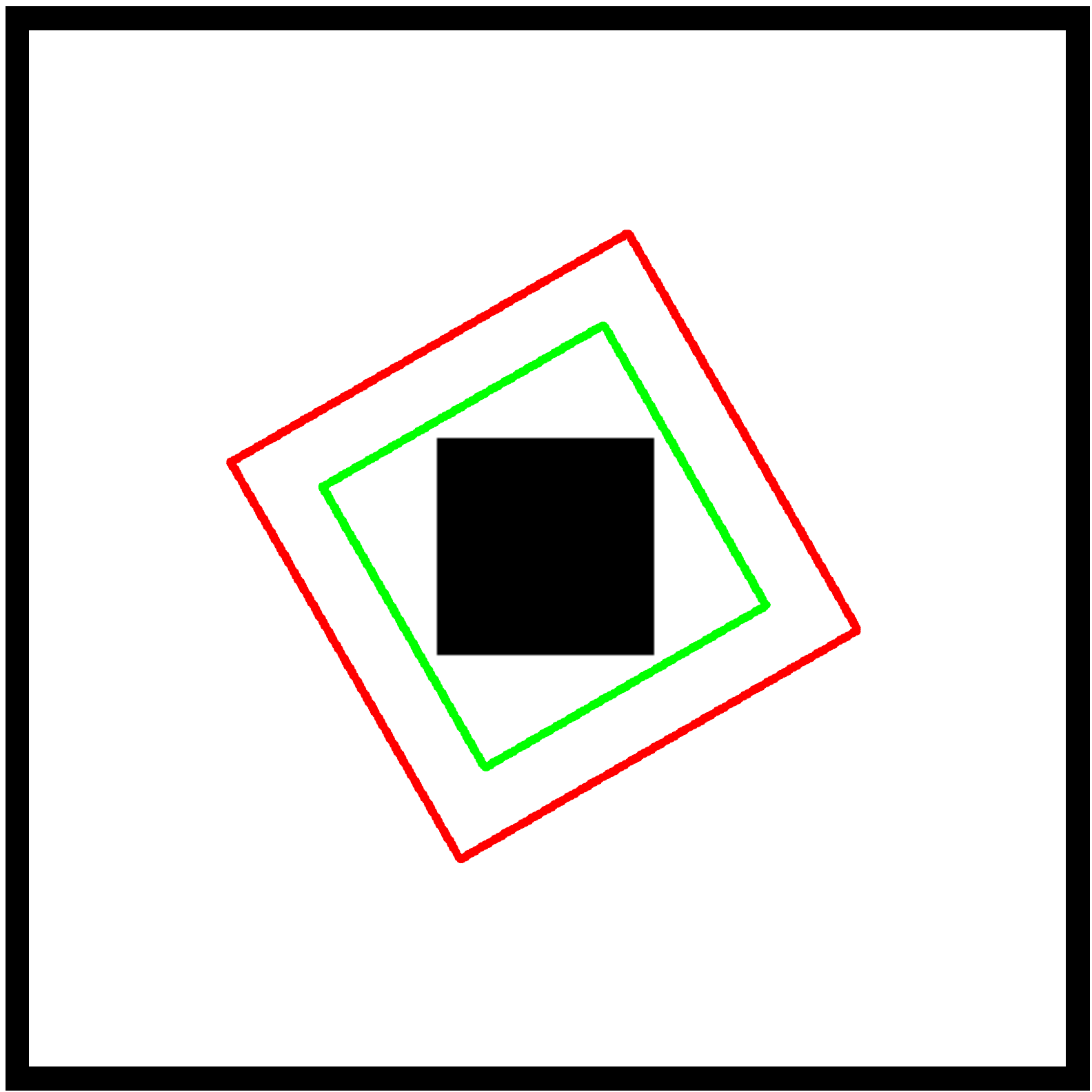}\\
(a) & (b) & (c)
\end{array}
$
\caption{To illustrate the need for normalization by area ($AB$) of the bounding box containing the shape template; $\alpha=2$; (a) Result with normalization; (b) and (c) suboptimal configurations obtained in the absence of normalization.}
\label{blackbox}
\end{figure}

In (\ref{philip}), $\Re_0$ is region bound by the inner contour and $\Re_1$, by the outer contour, with $|\Re_1|=2\,|\Re_0|$. The contrast term in (\ref{philip}) forces the snake to maximize the difference in average intensity between the annular region and inner contour. The parameterization adopted by Th\'evenaz et al. yields non-unique solutions, which are resolved with the help of the regularization term ($J_{Reg}$). The shape specific nature of the parameterization leads to a regularization functional that is also shape dependent. The contrast function considers the area bounded by the inner contour as foreground and the annular region as background. The area occupied by these regions on the image varies as the snake evolves. This is in contrast to the techniques in \cite{VeseChan,Yezzi} where the union of the foreground and background region is the constant user specified bounding box on the image.\\ 
\indent Motivated by the definition of local contrast proposed by Th\'evenaz et al., we propose a modified version of local contrast. Consider the nested snake design introduced in Section \ref{affine}. For a  snake initialized on an image $f$, let ${\Re_0}$ and ${\Re_1}$ be the regions enclosed by $(X_0(t),Y_0(t))$ and $(X_1(t),Y_1(t))$, respectively (as shown in Figure~\ref{EnergyFunctionExplanation}). We define our snake energy as follows:    
\normalsize
\begin{eqnarray} 
\label{Energy Formulation}
\mbox{$E$}&=&\frac{1}{AB}\Bigg\{ {a_0\int\int_{{\Re_1} \backslash {\Re_0}}{f\,\mathrm{d}x\,\,\mathrm{d}y}-a_r\int\int_{\Re_0}{f\,\mathrm{d}x\,\,\mathrm{d}y}} \Bigg\}\nonumber  \\
&=&\frac{1}{AB}\Bigg\{ a_0\underbrace{{\int\int_{\Re_1}{f\,\mathrm{d}x\,\,\mathrm{d}y}}}_{E_1}-\,a_1\underbrace{\int\int_{\Re_0}{f\,\mathrm{d}x\,\,\mathrm{d}y}}_{E_0}\Bigg\}, \nonumber \\
\end{eqnarray}
\normalsize
where $a_0$ and $ a_1$ are the areas enclosed by the inner and outer contours of the shape template respectively, $a_r$ is the area of the annular region of the shape template. Minimizing $E$ enables the active contour to lock on to objects that are brighter than their immediate surroundings, and maximizing it would have the opposite effect. The normalization term $AB$ in the denominator removes the ambiguity suffered by active contours. To explain the benefit of normalization, consider the example shown in Figure~\ref{blackbox}, which consists of a dark square in a bright background. In order to capture the dark square, $E$ needs to be maximized. Optimizing $E$ with the help of the normalization term results in the outline shown in Figure~\ref{blackbox}(a), whereas (b) and (c) show the results obtained without normalization. For all three cases, the value of $E=a_0E_1-a_1E_0$ is the same, but the fit is optimal when ${E}$ is normalized by $AB$, and suboptimal otherwise. It is also desirable for the snake to remain stationary when placed on a constant region. We see that if $f(x,y) = k$ (a constant), then $\displaystyle E=\frac{1}{AB}\big\{a_0ka_1AB-a_1ka_0AB\big\} = 0$, which means that the snake is inert on constant regions. Due to shape-specific nature of the proposed snake and one-one relationship between shape and parameters, our formulation does not require additional regularization energies. For symmetric shapes this one-one relationship does not hold. However, there are finite number of configurations of RAT for the same shape. All these local minima are strict local minima and the gradient descent technique will converge to one of the shapes without oscillating. The only exception is circular active contour, for which uncountably-infinite configurations of $\theta$ correspond to the same shape; we resort to the method proposed by us in \cite{Adithya1} for segmenting circular shapes.
\subsection{Fischer ratio interpretation of the snake energy}
\label{FischerInterpretation}
\indent We next show that optimizing $E$ in (\ref{Energy Formulation}) is analogous to maximizing the Fischer ratio \cite{duda} under certain conditions. Consider an image $f$ that contains a homogeneous bright object to be segmented in the presence of a relatively darker background. Assume that the image $f$ is corrupted with Gaussian noise with mean $\mu$ and variance $\sigma^2$. The distributions of $f$ within the bright object and outside of it are both Gaussian, with same variances, but with different means. The problem of outlining is therefore equivalent to a Gauss-Gauss detection problem \cite{kay}, which can be solved using a Fischer ratio approach.
The energy function in (\ref{Energy Formulation}) can be modified as
\begin{align}
\label{Fischer}
E&=\frac{a_0E_r-a_rE_0}{AB}=a_ra_0\left(\frac{E_r}{a_rAB}-\frac{E_0}{a_0AB}\right) \nonumber\\
&\propto \left ( \hat{E_r} - \hat{E_0} \right ).
\end{align}
For a snake initialized on an image, if $\hat{E_0} > \hat{E_r}$, minimizing the snake energy (\ref{Fischer}) is equivalent to maximizing $\displaystyle \left(\hat{E_r}-\hat{E_0}\right)^2$, where $\displaystyle \hat{E_r}\,\text{and}\,\hat{E_0}$ are the mean intensities of $f$ in the annular region and inner contour, respectively. The Fischer discriminant for this Gauss-Gauss detection problem is $F  = \frac{\displaystyle (\hat{E_r}-\hat{E_0})^2}{\displaystyle \sigma_{r}^2+\sigma_{0}^2}$, where $\sigma_{r}$, $\sigma_{0}$ are variances of $f$ in the annular region and the inner contour, respectively. Since $\sigma_r = \sigma_0$ under uniform Gaussian noise conditions, the proposed snake energy optimization is directly related to Fischer ratio maximization.

\subsection{Advantages of contrast energy over squared contrast energy}
\label{SignAgnosticSnakes}
\indent We assumed that the object is locally brighter or darker than its immediate surroundings and correspondingly selected the sign of the energy in (\ref{Energy Formulation}) for minimization. By squaring the energy and adding a negative sign in (\ref{Energy Formulation}), we would obtain the sign-agnostic version of the contrast energy, called squared contrast energy ($\underline{E}^2$), similar to mean separation energy in (\ref{mean sep}). Though $\underline{E}^2$ would overcome the problem of choosing the sign of the energy, we show that the energy in (\ref{Energy Formulation}) provides certain advantages over $\underline{E}^2$. We perform two experiments to compare local contrast function ($E$) in (\ref{Energy Formulation}) and local squared contrast energy ($\underline{E}^2$). We show that local contrast function is preferable to local squared contrast energy. 

\begin{figure}[t]
\centering 
$
\begin{array}{ccc}
\includegraphics[width=1.0in]{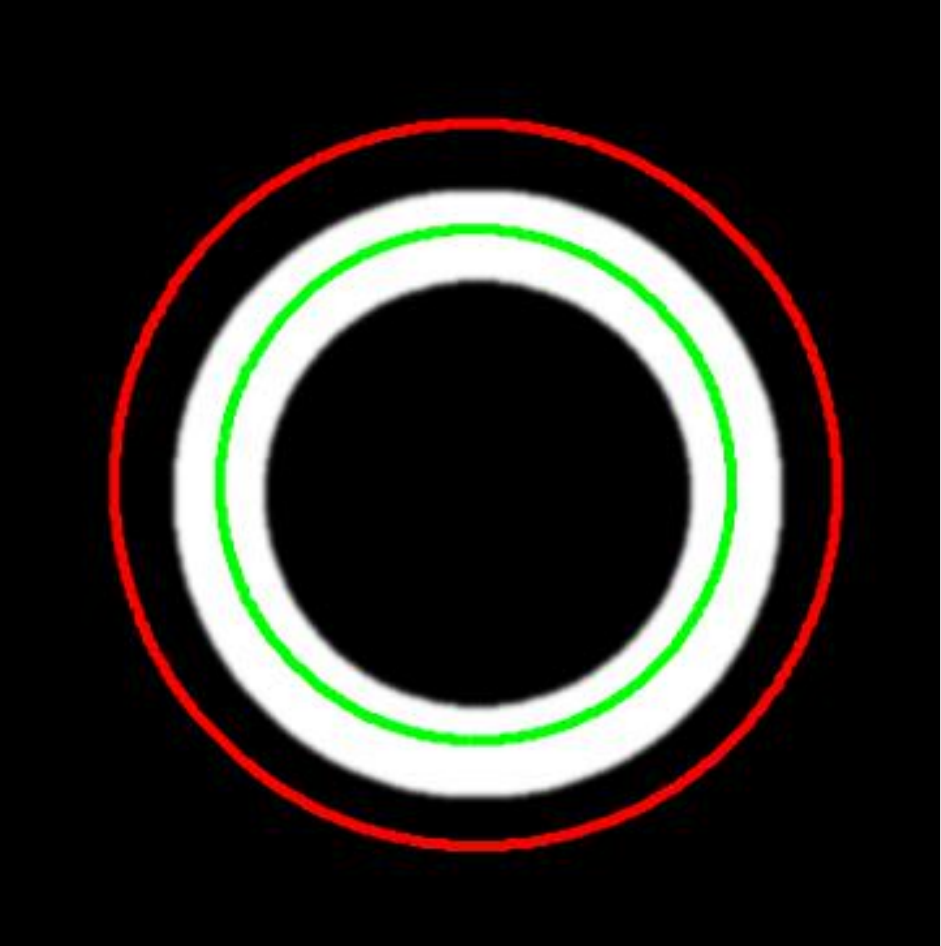} &
\includegraphics[width=1.0in]{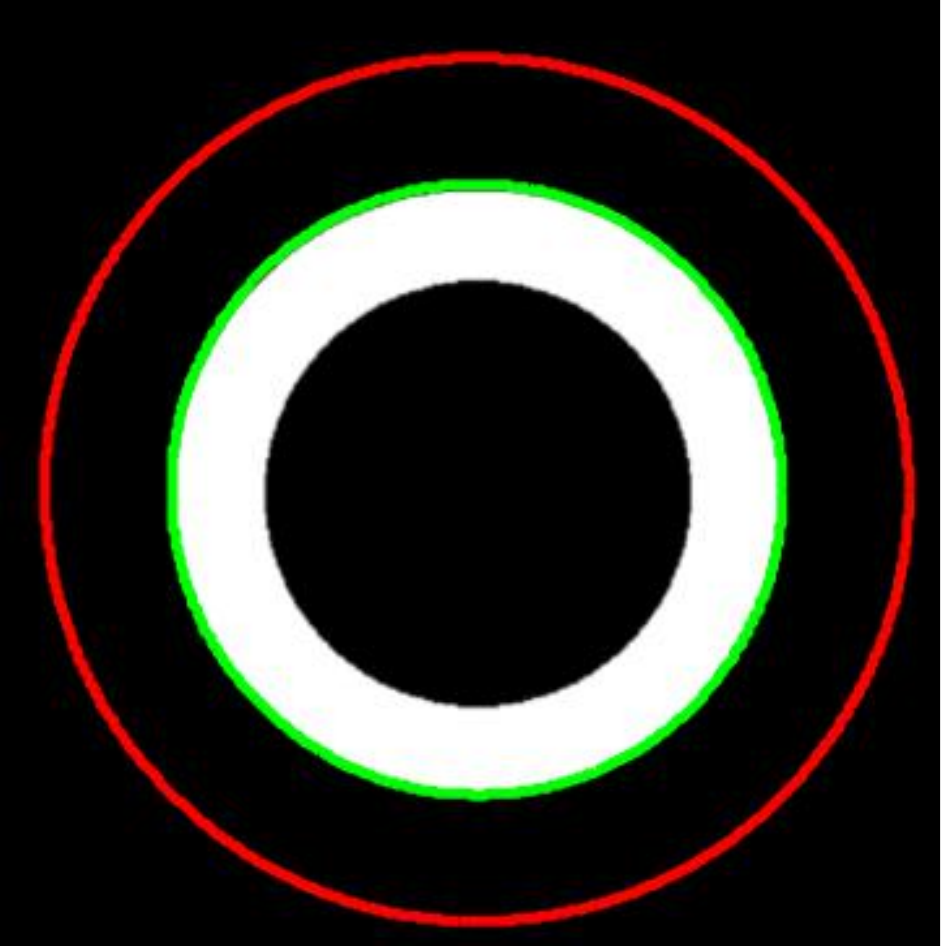} &
\includegraphics[width=1.0in]{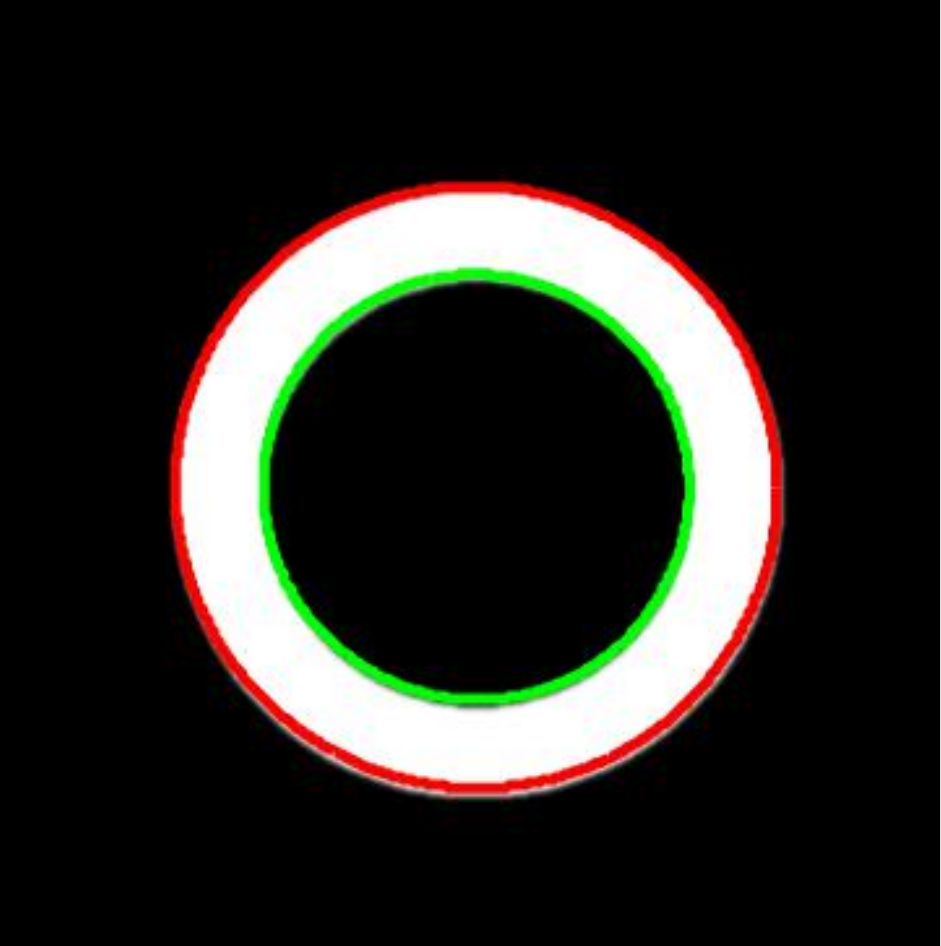}\\
(a) & (b) & (c) \\
\includegraphics[width=1.0in]{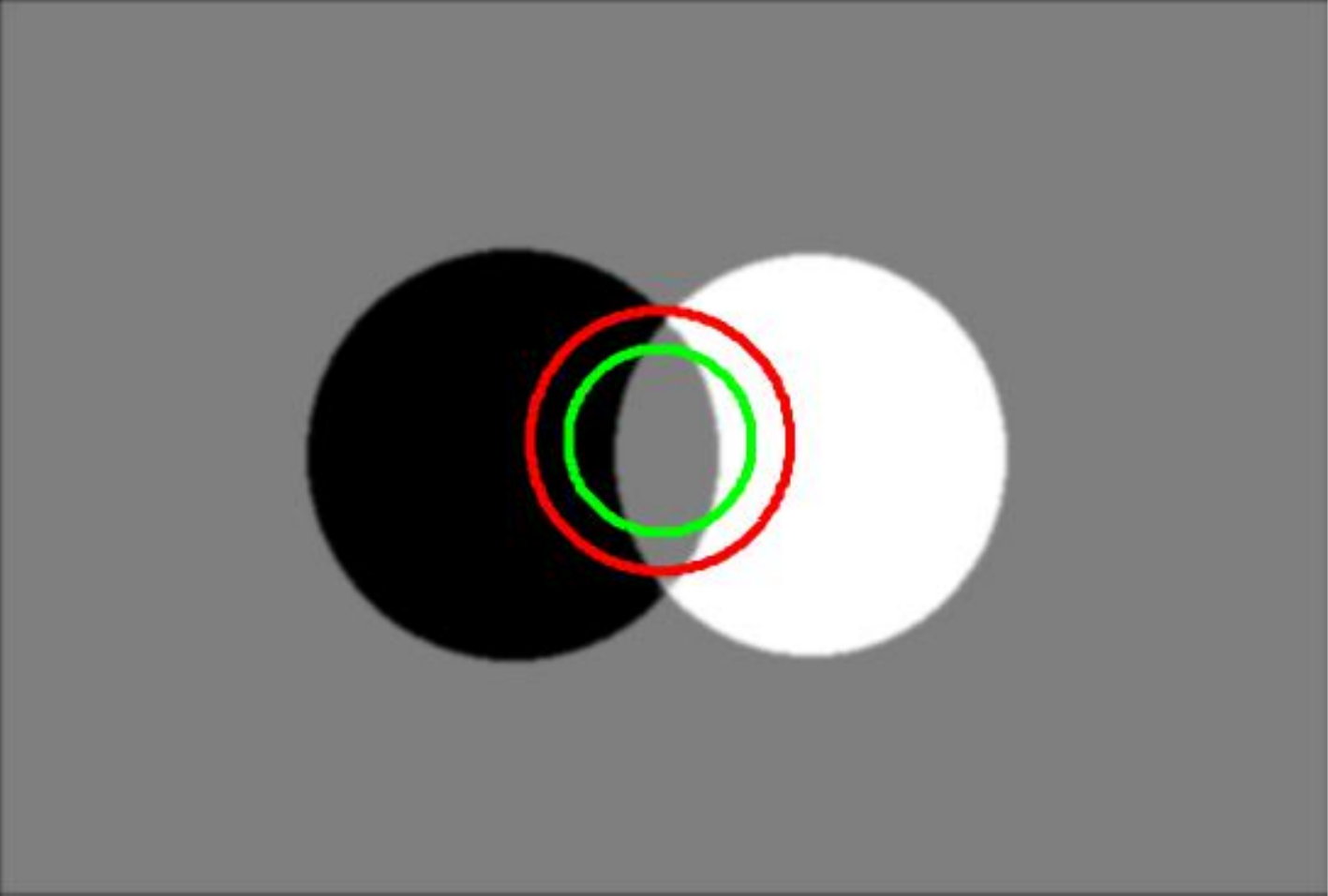} &
\includegraphics[width=1.0in]{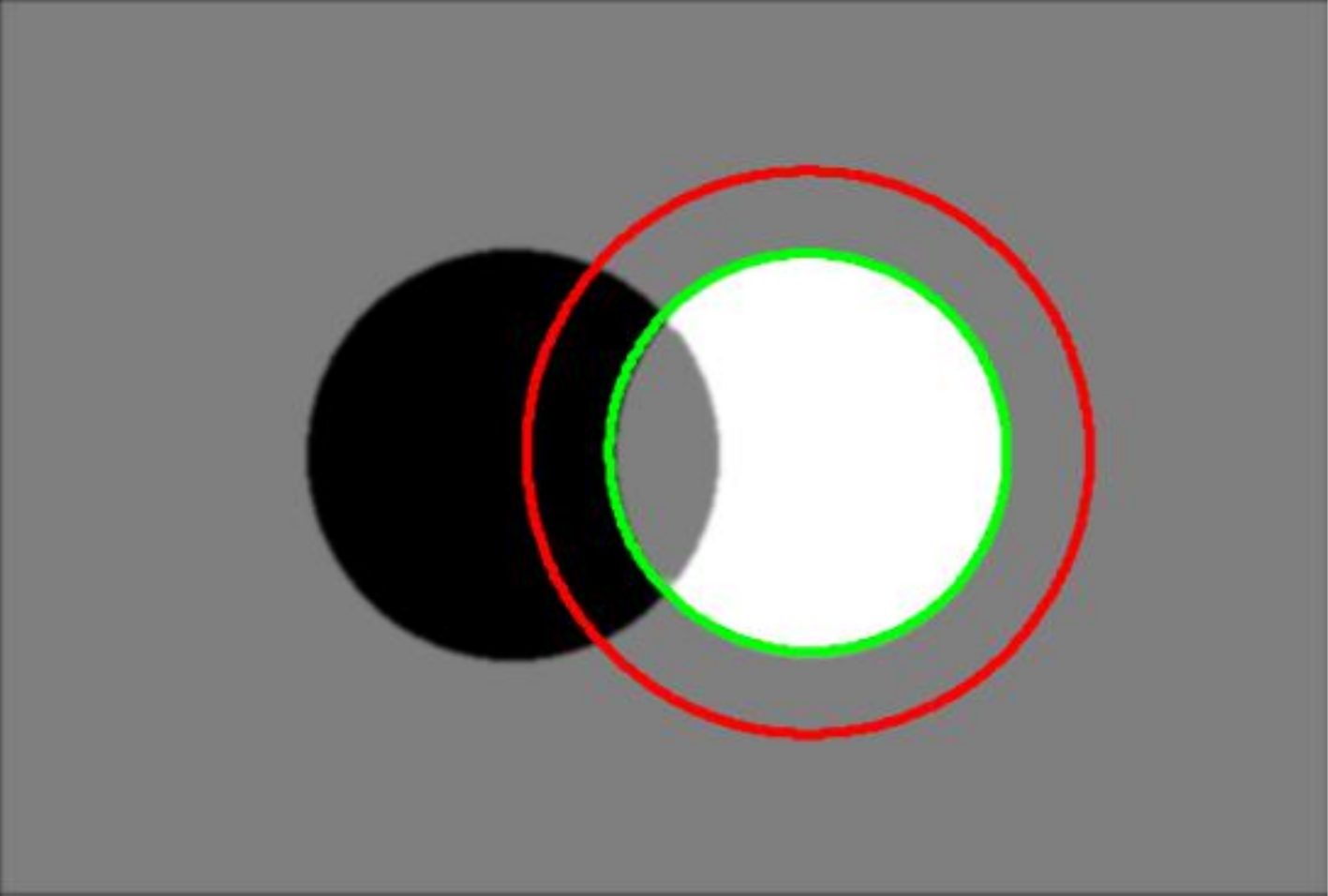} &
\includegraphics[width=1.0in]{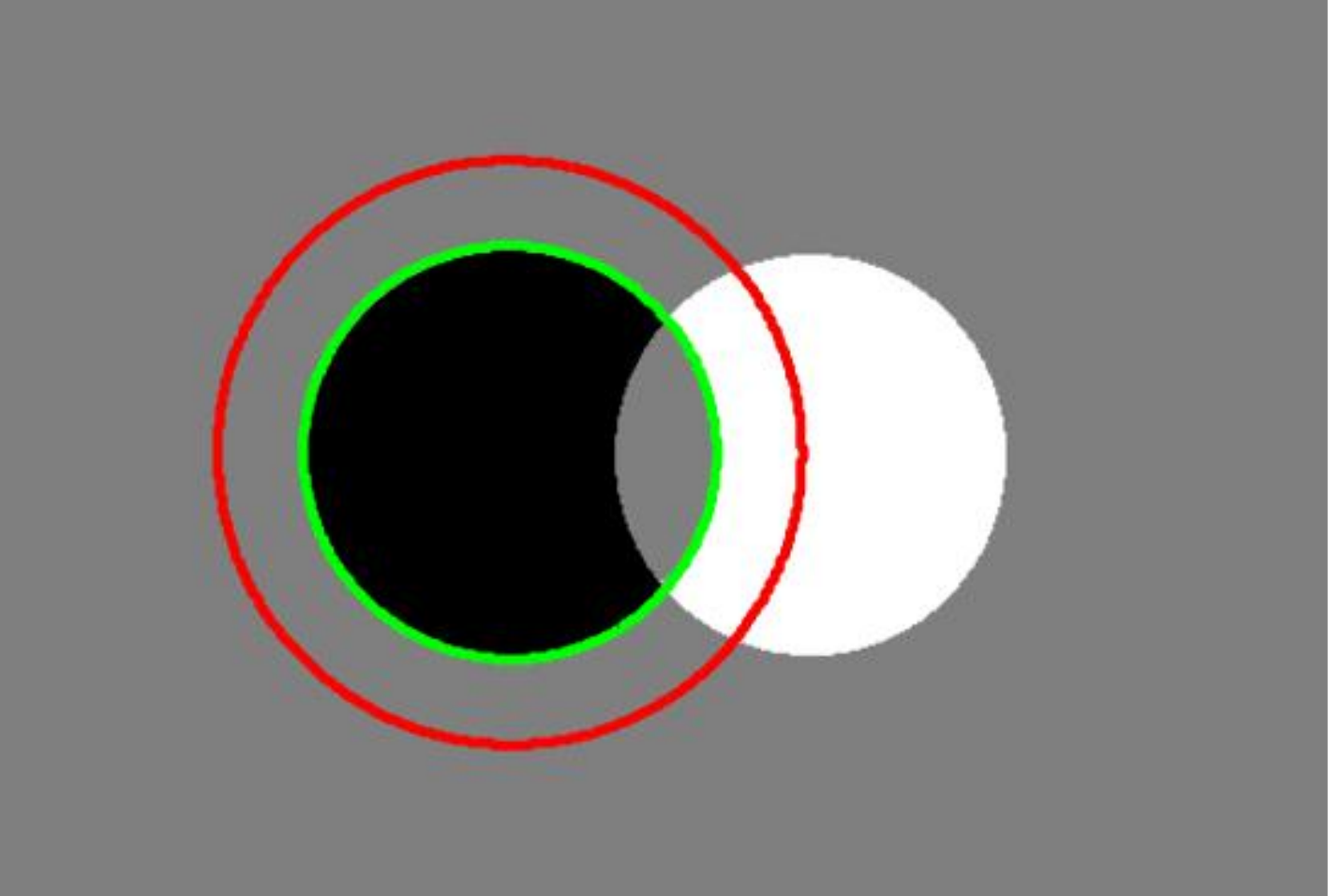}\\
 (d) & (e) & (f)
\end{array}
$
\caption{Comparison between $E$ and $\underline{E}^2$ snakes; Top row illustrates the first experiment; Bottom row illustrates the second experiment}
\label{Contrast}
\end{figure}
\begin{figure}[t]
\centering 
$
\begin{array}{c}
\includegraphics[width=3.0in]{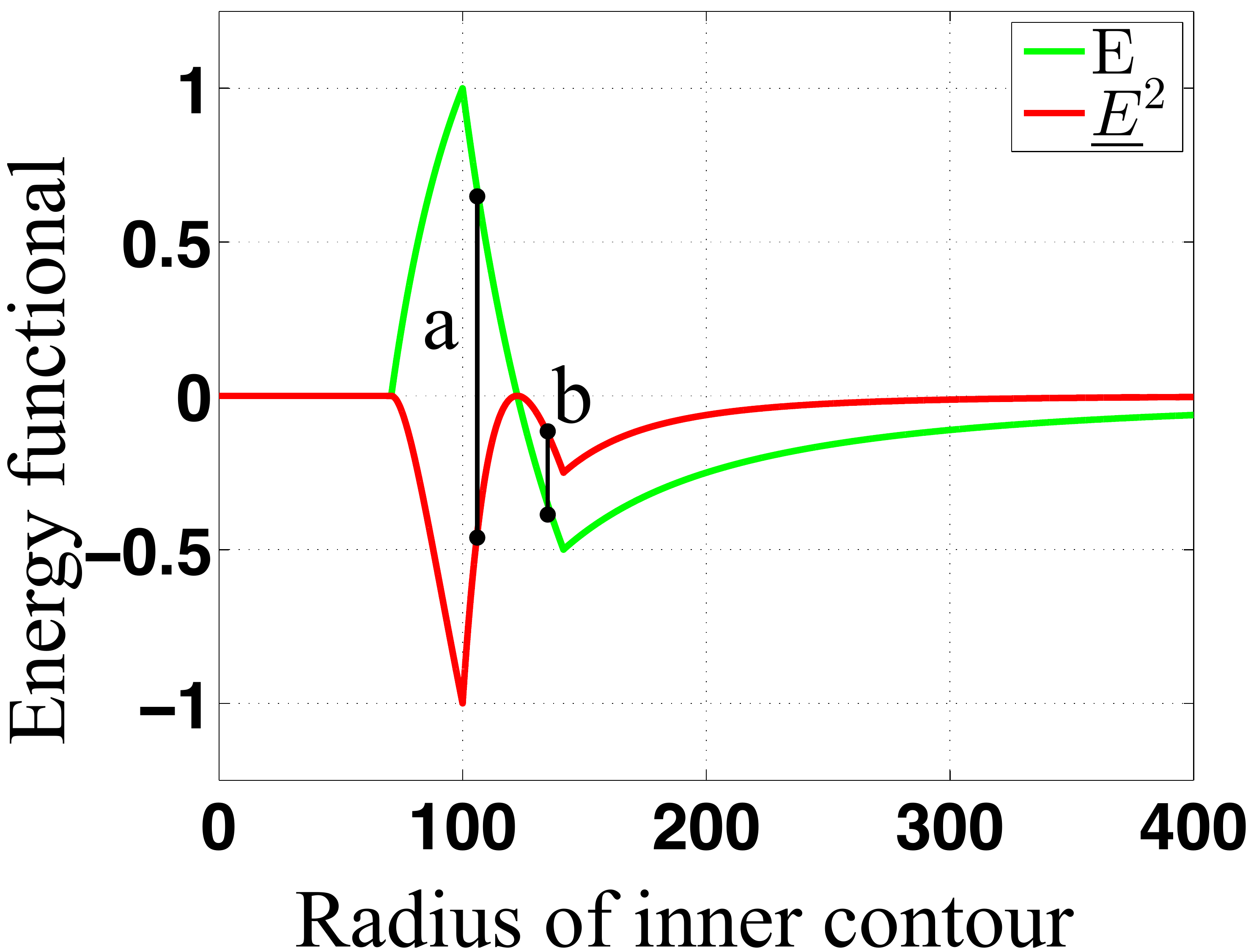}
\end{array}
$
\caption{Comparison of energy functions, red color represents $\underline{E}^2$ functional and green represents $E$ functional. Some of the local maxima of $E$ functional become local minima for $\underline{E}^2$ functional.}
\label{SnakeEnergyComparison}
\end{figure}

\indent In the first experiment, we consider a bright ring shaped structure against a dark background as shown in Figure~\ref{Contrast}(a) having an inner and outer radius of 100 and 141 ($\approx$ 100$\sqrt{2}$) pixels, respectively. We choose a circular template (ratio of inner to outer radius of template fixed at 1:$\sqrt{2}$) for segmentation and initialize the snake as shown in Figure~\ref{Contrast}(a). We plot the energies $E$ and $\underline{E}^2$ as the template is uniformly scaled (i.e. $A=B$) between (0, 400) about the center of the bright ring. The plot is shown in Figure~\ref{SnakeEnergyComparison}, where the green curve corresponds to $E$ and the red curve to $\underline{E}^2$. The energies are plotted against the radius of the inner contour of the shape template on the horizontal axis. From the plot we observe that for a snake initialized as shown in Figure~\ref{Contrast}(a) and having an initial inner contour radius $\mathrm{a} \in (100,100\sqrt{2})$, minimization of $E$ causes the snake to converge to the result shown in Figure~\ref{Contrast}(b). Maximization causes the snake to converge to the result shown in Figure~\ref{Contrast}(c). However, for $\underline{E}^2$ snake energy, the final result is  dependent on the initialization. Specifically, the result in Figure~\ref{Contrast}(c) requires an initialization between 100 and 120 pixels while the result in Figure~\ref{Contrast}(b) requires an initialization beyond 120 pixels, shown as `b' in Figure~\ref{SnakeEnergyComparison}.\\ 
\indent In second experiment, we consider an image containing two overlapping circular objects as shown in Figure~\ref{Contrast}(d). The pixel intensity value of the two objects are chosen differently while the region of overlap and the background are set to the average value. We consider a snake (derived from a circular template) initialized on the intersection of the two circular objects as shown in Figure~\ref{Contrast}(d). Minimizing $E$ gives the result in Figure~\ref{Contrast}(e), while maximizing $E$ gives the result in Figure~\ref{Contrast}(f). However, the relative area of overlap with each object at the time of initialization determined the final output for $\underline{E}^2$ optimizing snake. In both experiments, $E$ had a larger basin of attraction as compared to $\underline{E}^2$, which resulted in better control of the final segmentation output of the snake. \\
\indent We further substantiate that $E$ snakes will have larger basin of attraction compared to $\underline{E}^2$ snakes by showing that some spurious local minima are present in $\underline{E}^2$ snakes which are not local minima for $E$ snakes. Consider the local maxima for $E$ snakes (for $E >$ 0). These local maxima will become local minima for $\underline{E}^2$ as $\displaystyle \frac{\partial^2 }{\partial \tau^2}\underline{E}^2 > 0$ ($\tau=A,\,B,\,\theta,\,x_c,\,y_c$). These additional minima locations correspond to sub-optimal solutions and decrease the basin of attraction. Also, the snakes optimizing the local contrast carry more information (sign of energy function) and hence, their segmentation performance is superior to $\underline{E}^2$ snakes. To conclude, local square contrast function is more sensitive to initialization and less controllable compared to local squared contrast function. We use local contrast ($E$) as energy functional in the rest of the paper.


\section{Active Contour Optimization}
\label{Affineopt}
\indent We next proceed with the optimization of the snake energy $E$ with respect to the RAT parameters $A, B, \theta, x_c,$ and $y_c$. We need the partial derivatives of $E$ with respect to the parameters to enable gradient-descent optimization. The energy $E$ involves double integrals, and the parameters define the boundaries. These, in turn, define the regions ${\Re_0}$ and ${\Re_1}$ over which the integrals are evaluated. Direct optimization seems to be a difficult task; however, thanks to Green's theorem the surface integral can be expressed conveniently as a contour integral. This methodology simplifies the calculations of the partial derivatives and consequently, we have to deal with contour integrals.\\
\indent We first perform a coordinate-axes transformation from $(x,y)$ to $(X,Y)$ such that
\begin{eqnarray} 
\left(\begin{array}{cc} 
X \nonumber \\
Y\nonumber 
\end{array}\right) 
=
\left(\begin{array}{ccc}
\quad A\cos\theta & B\sin\theta & x_c\nonumber \\
-A\sin\theta & B\cos\theta & y_c \nonumber 
\end{array}\right) \nonumber
\left(\begin{array}{ccc}
x \nonumber \\ 
y\nonumber \\
1
\end{array}\right),
\end{eqnarray}
which converts the image function $f(x,y)$ to
\begin{multline}
\displaystyle F(X,Y)=f\left(\frac{(X-x_c)\cos \theta}{A}\right.-\frac{(Y-y_c)\sin \theta}{A}, \nonumber \\ 
\frac{(X-x_c)\sin \theta}{B}+\left.\frac{(Y-y_c)\cos \theta}{B}\right). \nonumber
\end{multline}
\normalsize
We could have preserved the image as it is, and performed the axes transformation on the inner and outer contours, but we found that transforming the image to the new coordinate system simplifies the calculations and also results in brevity of notation. The resulting energies $E_0$ and $E_1$ take the form:
\begin{align}
E_0&=\int\int_{\Re_0}{F(X, Y)\,\mathrm{d}x\,\,\mathrm{d}y}, \mbox{and} \nonumber \\
E_1&=\int\int_{\Re_1}{F(X, Y)\,\mathrm{d}x\,\,\mathrm{d}y}, \mbox{respectively.} \nonumber
\end{align}
\normalsize
\indent For brevity of notation, we have dropped the parameter $t$ and denoted $(X(t), Y(t)), (x(t), y(t))$ as $(X, Y), (x,y)$, respectively. We next invoke Green's theorem for computation of $E_0$ and $E_1$. We show the calculations only for $E_0$, and those for $E_1$ follow analogously. Applying Green's theorem to $E_0$ leads to the following expression:

\begin{align}
E_0=-\oint_{\Re_0}F^{Y}\,\mathrm{d}x=\oint_{\Re_0}F^{X}\,\mathrm{d}y, \text{where} 
\label{Green's theorem}
\end{align}
\begin{align}
F^{Y}(X,Y)&=\int_{-\infty}^{Y}F(X,\zeta)\,\mathrm{d}\zeta,\,\mbox{and} \nonumber \\ 
F^{X}(X,Y)&=\int_{-\infty}^{X}F(\zeta,Y)\,\mathrm{d}\zeta. \nonumber 
\end{align}
\normalsize
$E_0$ is a function of $(X, Y)$, which are in turn functions of the parameters of the contours. The partial derivative of $E_0$ with respect to $A$  is given by the chain rule as follows:
\begin{align}
\label{totalderivative}
\frac{\partial E_0}{\partial A}&=\frac{\partial E_0}{\partial X}\frac{\partial X}{\partial A}+\frac{\partial E_0}{\partial Y}\frac{\partial Y}{\partial A}.
\end{align}
Substituting (\ref{Green's theorem}) in (\ref{totalderivative}), we get that 
\begin{align}
\frac{\partial E_0}{\partial A}&= \oint_{\Re_0}\frac{\partial F^{X}}{\partial X}\frac{\partial X}{\partial A}\mathrm{d}Y-\oint_{\Re_0}\frac{\partial F^{Y}}{\partial Y}\frac{\partial Y}{\partial A}\mathrm{d}X \nonumber \\
&=\oint_{\Re_0}F(X,Y) x\cos \theta \left( -A \sin \theta\, \mathrm{d}x+B\cos \theta\, \mathrm{d}y\right) \nonumber \\
&\,\,\,\,\,\,\,\,\,\,\,\,\,\,\,\,+\oint_{\Re_0}F(X,Y)x\sin \theta \left( A \cos \theta\, \mathrm{d}x+B\sin \theta\, \mathrm{d}y \right) \nonumber \\
&=B\oint_{\Re_0}F(X,Y)\,x\, \mathrm{d}y. 
\label{withA} 
\end{align}
Similarly, the partial derivatives of $E_0$ with respect to the parameters $B,\,\theta,\,x_c,$ and $y_c$ are given as,
\begin{align}
\frac{\partial E_0}{\partial B}&=-A\oint_{\Re_0} F(X,Y)\,y\, \mathrm{d}x, \nonumber\\
\frac{\partial E_0}{\partial \theta}&=\oint_{\Re_0} F(X,Y)(A^2 x\,\mathrm{d}x+B^2y\,\mathrm{d}y), \nonumber \\
\frac{\partial E_0}{\partial x_c}&=\oint_{\Re_0} F(X,Y)(-A \sin\theta \,\mathrm{d}x+B \cos\theta\, \mathrm{d}y), \mbox{and}\nonumber  \\
\frac{\partial E_0}{\partial y_c}&=\oint_{\Re_0} F(X,Y)(-A \cos\theta \,\mathrm{d}x-B \sin\theta\, \mathrm{d}y).\nonumber
\end{align}
\indent The generalized expressions for the partial derivatives of $E$ in (\ref{Energy Formulation}) with respect to the RAT parameters are given below:
\begin{align}
\frac{\partial \mbox{E}}{\partial \tau}&=\frac{a_0 \displaystyle\frac{\partial \mbox{E}_1}{\partial \tau}-a_1 \displaystyle\frac{\partial \mbox{E}_0}{\partial \tau}}{AB}-\frac{E}{\tau\,  AB }, \,\,\,\, \tau=A,B, \text{ and} \nonumber \\
\frac{\partial \mbox{E}}{\partial \tau}&=\frac{a_0 \displaystyle\frac{\partial \mbox{E}_1}{\partial \tau}-a_1\displaystyle \frac{\partial \mbox{E}_0}{\partial \tau}}{AB} , \label{final2} \,\,\,\, \tau=x_c,y_c,\theta .
\end{align}
We simplified (\ref{final2}) by substituting the values of partial derivatives of $E_0$ and $E_1$ with respect to RAT parameters. Expressing them in terms of $x_0(t)$, $y_0(t)$, $x_1(t)$, $y_1(t)$ and the image $f$, we have  
\begin{align} 
\label{GS:partial derivatives}
\frac{\partial E}{\partial A}&=\frac{a_0}{A}\oint_{\Re_1}f(X_1,Y_1)x_1(t)y'_1(t)\,\mathrm{d}t \nonumber\\
&- \frac{a_1}{A}\oint_{\Re_0}f(X_0,Y_0)x_0(t)y'_0(t)\,\mathrm{d}t - \frac{1}{A^2B} E, \nonumber \\
\frac{\partial E}{\partial B}&=-\frac{a_0}{B}\oint_{\Re_1}f(X_1,Y_1)x'_1(t)y_1(t)\,\mathrm{d}t \nonumber\\
&+\frac{a_1}{B}\oint_{\Re_0}f(X_0,Y_0)x'_0(t)y_0(t)\,\mathrm{d}t - \frac{1}{AB^2}E, \nonumber \\
\frac{\partial E}{\partial \theta}&=\frac{a_0}{AB}\oint_{\Re_1}f(X_1,Y_1)\left(A^2x_1(t)x'_1(t)+B^2y_1(t)y'_1(t)\right)\,\mathrm{d}t \nonumber \\ 
                                                                     & -\frac{a_1}{AB}\oint_{\Re_0}f(X_0,Y_0)\left(A^2x_0(t)x'_0(t)+B^2y_0(t)y'_0(t)\right)\,\mathrm{d}t,\nonumber\\
\frac{\partial E}{\partial x_c}&=\frac{a_0}{AB}\oint_{\Re_1}f(X_1,Y_1)(-A \sin\theta x'_1(t)+B \cos\theta y'_1(t))\,\mathrm{d}t\nonumber\\
&-\frac{a_1}{AB}\oint_{\Re_0}f(X_0,Y_0)(-A \sin\theta x'_0(t)+B \cos\theta y'_0(t))\,\mathrm{d}t,\nonumber\\
\text{and}\nonumber \\
\frac{\partial E}{\partial y_c}&=-\frac{a_0}{AB}\oint_{\Re_1}f(X_1,Y_1)(A \cos\theta x'_1(t)+B \sin\theta y'_1(t))\,\mathrm{d}t\nonumber\\
&+\frac{a_1}{AB}\oint_{\Re_0}f(X_0,Y_0)(A \cos\theta  x'_0(t)+B \sin\theta  y'_0(t))\,\mathrm{d}t. \nonumber 
\end{align}
The set of preceding equations is valid for any parametric representation of the template. We consider B-spline parametrization and provide simplified expressions in matrix form in (\ref{matrixform}) and (\ref{matrixform2}).

\section{Computational complexity}
\label{optimization}
\begin{figure*}[t]
\normalsize
\begin{eqnarray}
\left(\begin{array}{c}\partial E_i/\partial A \\ \partial E_i/\partial B \\ \partial E_i/\partial \theta\end{array}\right) &=&  \displaystyle \frac{\alpha_i}{AB} \sum \limits_{k, l =0}^{M_i-1} R_{f}(k,l) \left(\begin{array}{c}B\,c_{x_i,k}c_{y_i,l} \\-A\,c_{x_i,l}c_{y_i,k} \\ A^2c_{x_i,k}c_{x_i,l}+B^2c_{y_i,k}c_{y_i,l} \end{array}\right) - \left(\begin{array}{c}\displaystyle \frac{1}{A^2B} \\  \displaystyle \frac{1}{AB^2} \\ \displaystyle 0\end{array}\right) E_i, 
\label{matrixform}
\end{eqnarray}
\begin{eqnarray}
&\left(\begin{array}{c}\partial E_i/\partial x_c  \\ \partial E_i/\partial y_c \end{array}\right)=\displaystyle \frac{\alpha_i}{AB}\,\,\, \sum \limits_{k=0}^{M_i-1}\left(\begin{array}{c}-A\sin\theta c_{x_i,k}+B\cos\theta\ c_{y_i,k} \\ -A\cos\theta c_{x_i,k}-B\sin\theta\ c_{y_i,k} \end{array}\right) \times S_f(k),  
\label{matrixform2}
\end{eqnarray}
where $R_{f_{i}}(k,l) = \displaystyle\int_{0}^{M_i} f(X_i(t),Y_i(t)) b_{k,l}(t)\,\mathrm{d}t$, and $b_{k,l}(t)$= $\beta_{P}(t-k) \beta'_{P}(t-l),\,M_i$ denotes the number of knots. $S_{f_{i}}(k) = \displaystyle{\int_{0}^{M} f(X_i(t),Y_i(t)) \beta'_{P}(t-k)\,\mathrm{d}t.}$ and $\alpha_{i}=\begin{cases}
a_{1} & \text{for }i=0,\\
a_{0} & \text{for }i=1.
\end{cases}$\\
\vspace{0.2in}
\line(1,0){500}
\normalsize 
\end{figure*}
\indent The main steps involved in the computation of partial derivatives are given below.
\begin{enumerate}
\item The continuous-domain expressions for $R_{f,0}(k,l)$ and $R_{f,1}(k,l)$ in (\ref{matrixform}) have to be discretized for the purpose of computation. This is achieved by sampling the contour finely, which leads to the approximation:
\begin{eqnarray}
R_{f,i}(k,l)\approx&\displaystyle\frac{1}{R}\sum_{j=0}^{M_{i}R-1}f\left(X_i\left(\frac{j}{R}\right),Y_i\left(\frac{j}{R}\right)\right)\nonumber \\
&\times\beta_P\left(\frac{j-kR}{R} \right)\beta'_P\left(\frac{j-lR}{R} \right), \mbox{and}\label{discrete-1}\nonumber\\ 
S_{f,i}(k)\approx&\displaystyle\frac{1}{R}\sum_{j=0}^{M_{i}R-1}f\left(X_i\left(\frac{j}{R}\right),Y_i\left(\frac{j}{R}\right)\right)\nonumber \\
& \times \beta'_P\left(\frac{j-kR}{R} \right), \nonumber 
\label{discrete-2}
\end{eqnarray}
where $M_i$ is the number of knots, $R$ is the number of discretization steps over $[0,1]$, leading to an overall number of $M_iR$ points along the closed contour. The terms involving $\beta_P$ are spline-dependent and can be precomputed. Also, $R_{f,i}(k,l)$ is common in the partial derivatives with respect to $A$, $B$, and $\theta$; $S_{f,i}(k)$ is common in the partial derivatives with respect to $x_c$ and $y_c$.
\item The computational complexity in (\ref{matrixform}) and (\ref{matrixform2}) is ${\cal O}\left(RM^3\right)$ and ${\cal O}\left(RM^2\right)$ operations, respectively. Further, since the B-splines are compactly supported, look-up tables for $\displaystyle \beta_P\beta'_P$ and $\beta'_P$ are precomputed
\end{enumerate}
The use of gradient descent for optimization warrants a stopping criterion, for which we chose to monitor the snake energy $E$.
Let $E_{i}$ be the value of $E$ in the $i^{th}$ iteration during snake optimization, and $E_{i+1}$ the value at ${(i+1)}^{th}$ iteration. Denote $E_{diff,i}=E_{i+1}-E_{i}$. If the number of zero-crossings of $E_{diff}$ crosses a predefined threshold, then we stop the gradient descent. Zero crossings is a good indicator as RAT parameters oscillate about the optimal solution, a behavior typical to gradient descent.   
\section{Experimental Results}
\label{validation}
\indent In this section, we examine the performance of the proposed technique on multiple images, and the effect of initialization, noise, and occlusion.
\subsection{Effect of initialization}
\begin{figure*}[t]
\centering
$
\begin{array}{ccccccc}
\includegraphics[width=0.85in]{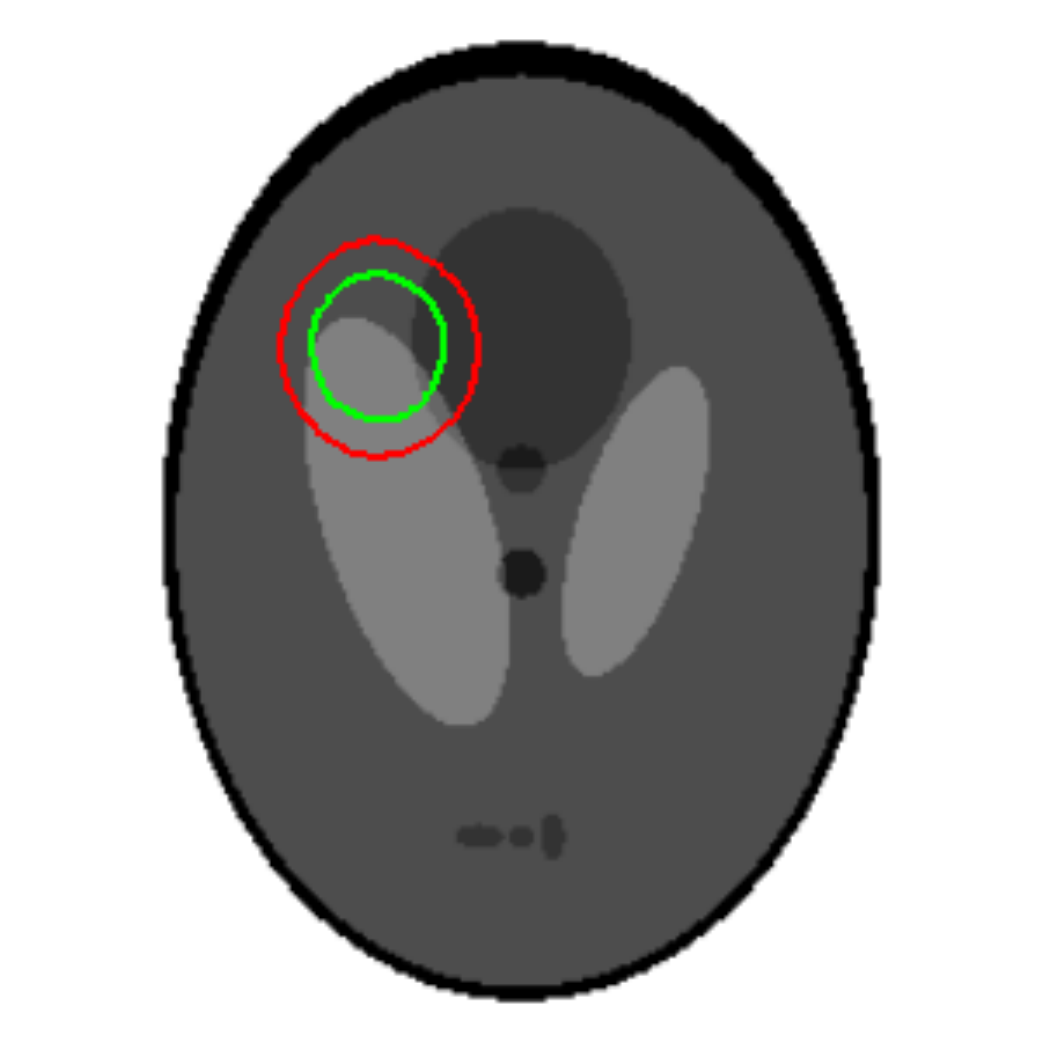}&
\includegraphics[width=0.85in]{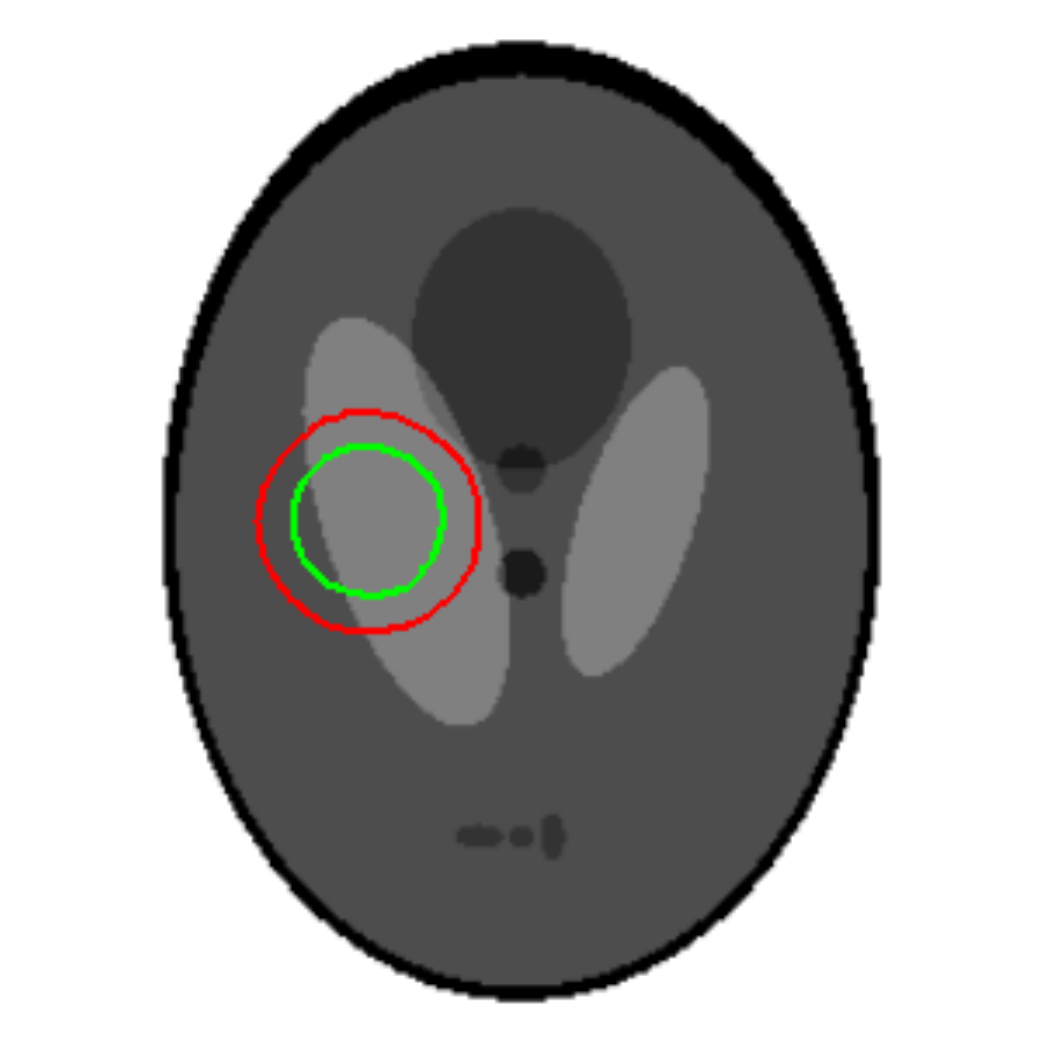}&
\includegraphics[width=0.85in]{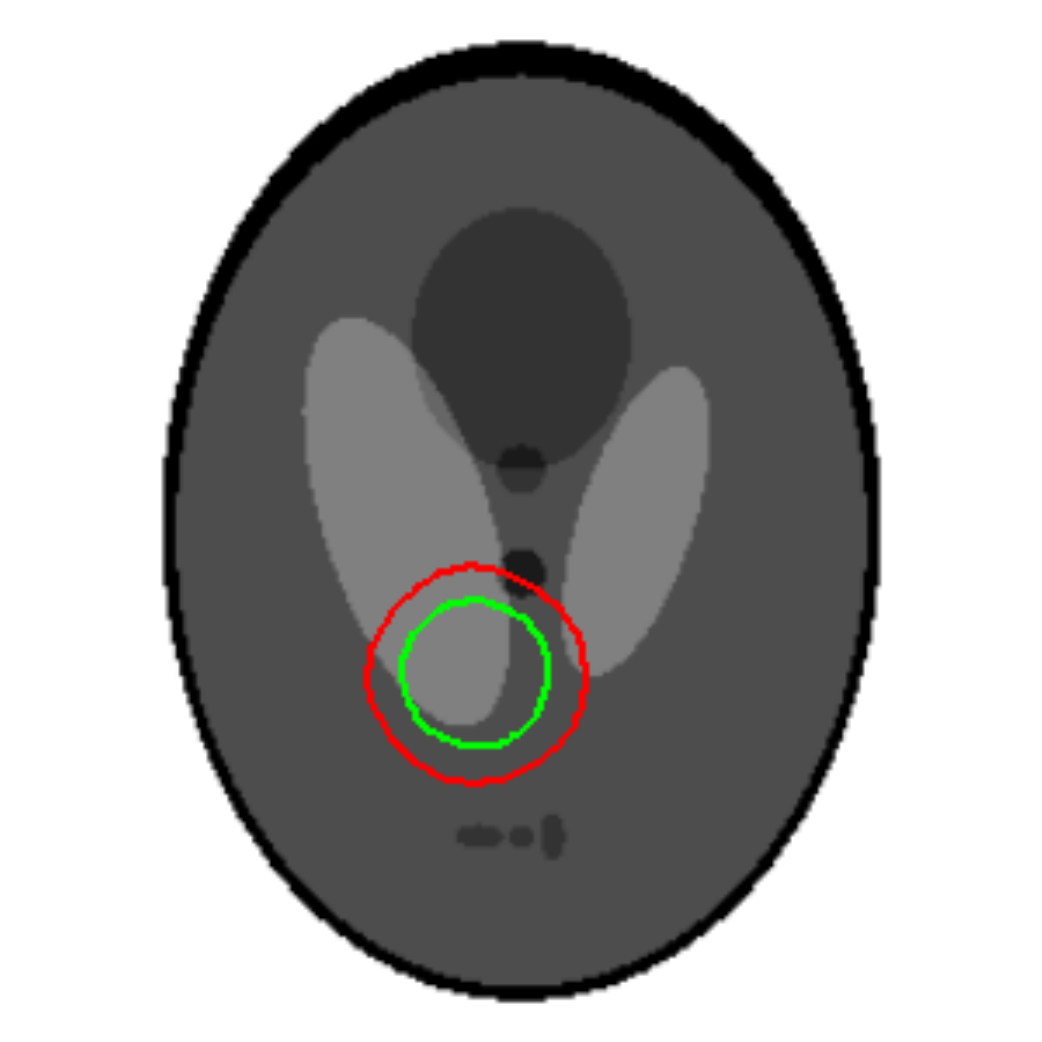}&
\includegraphics[width=0.85in]{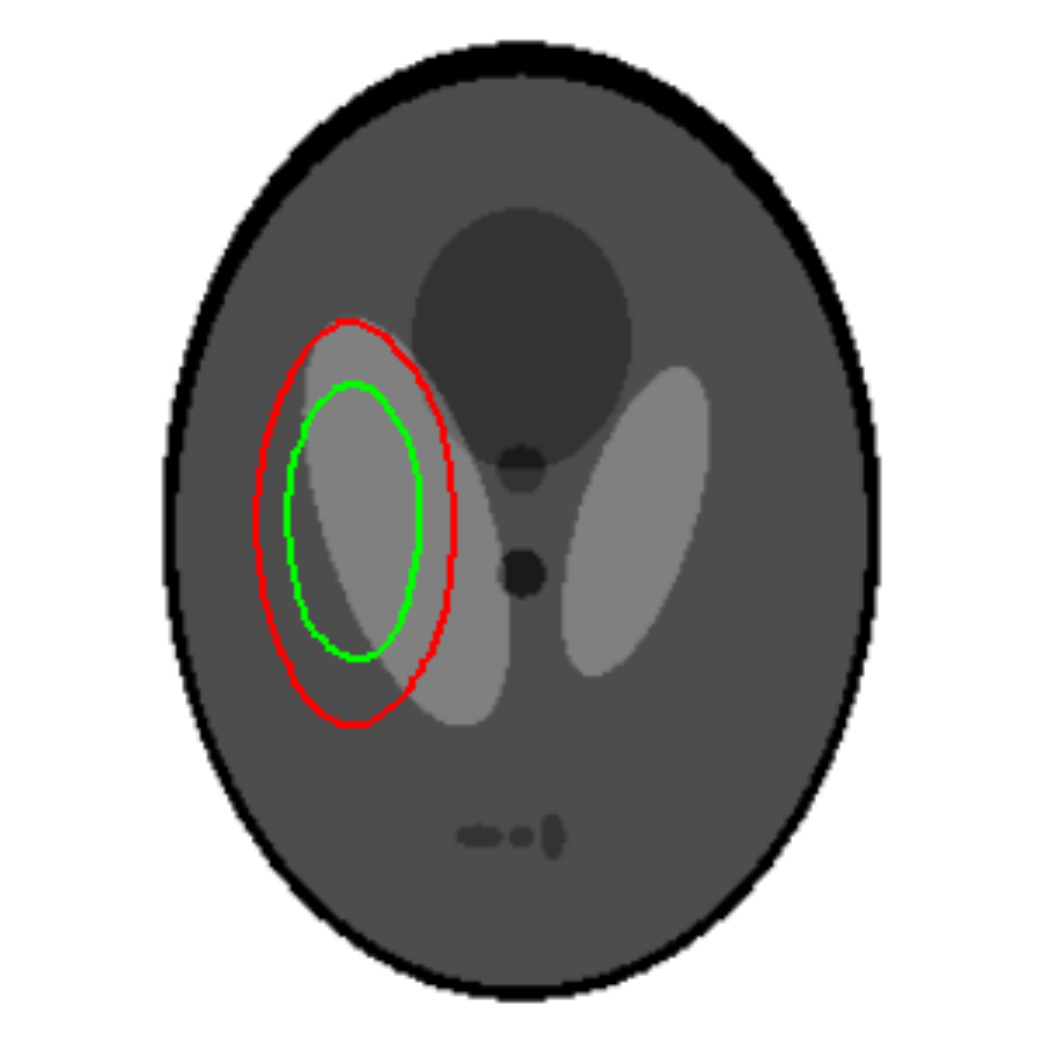}&
\includegraphics[width=0.85in]{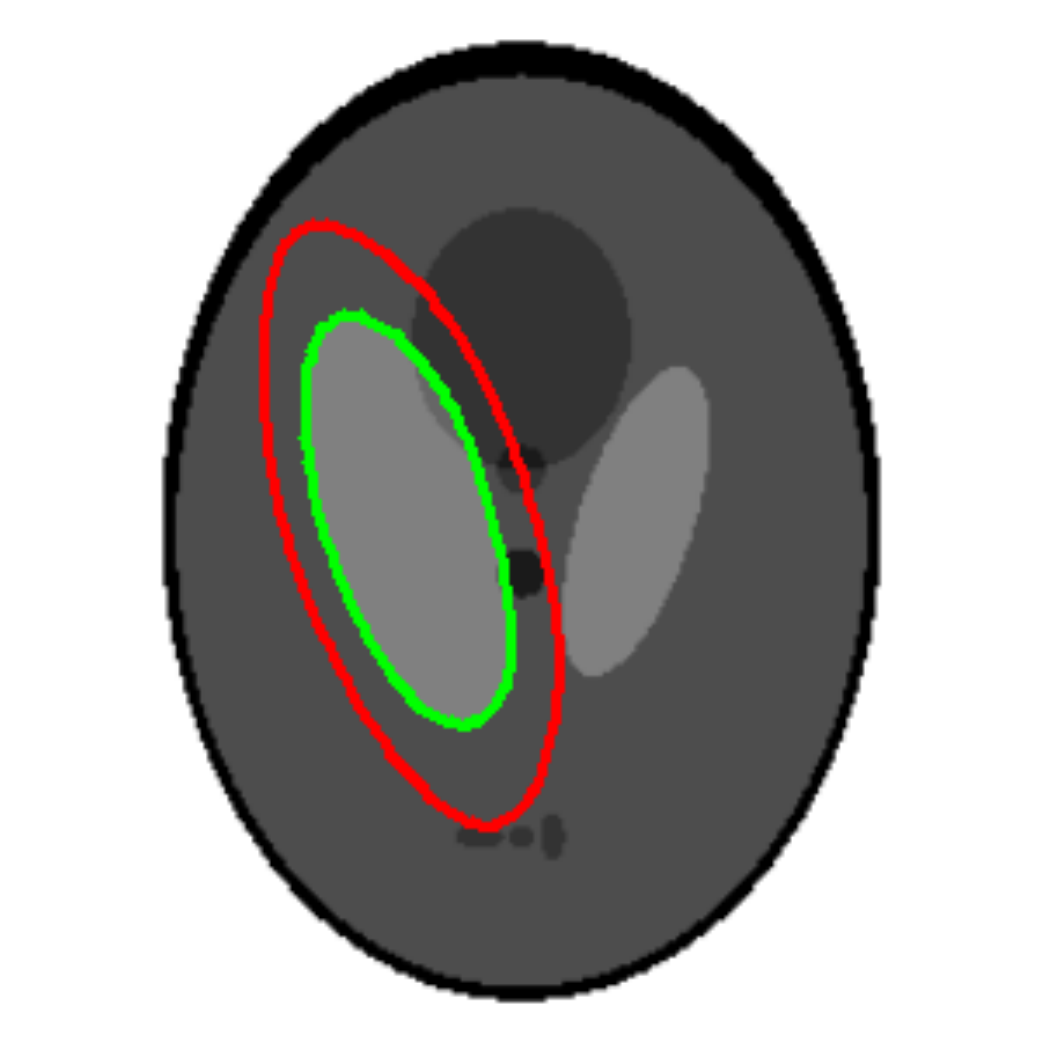}&
\includegraphics[width=0.85in]{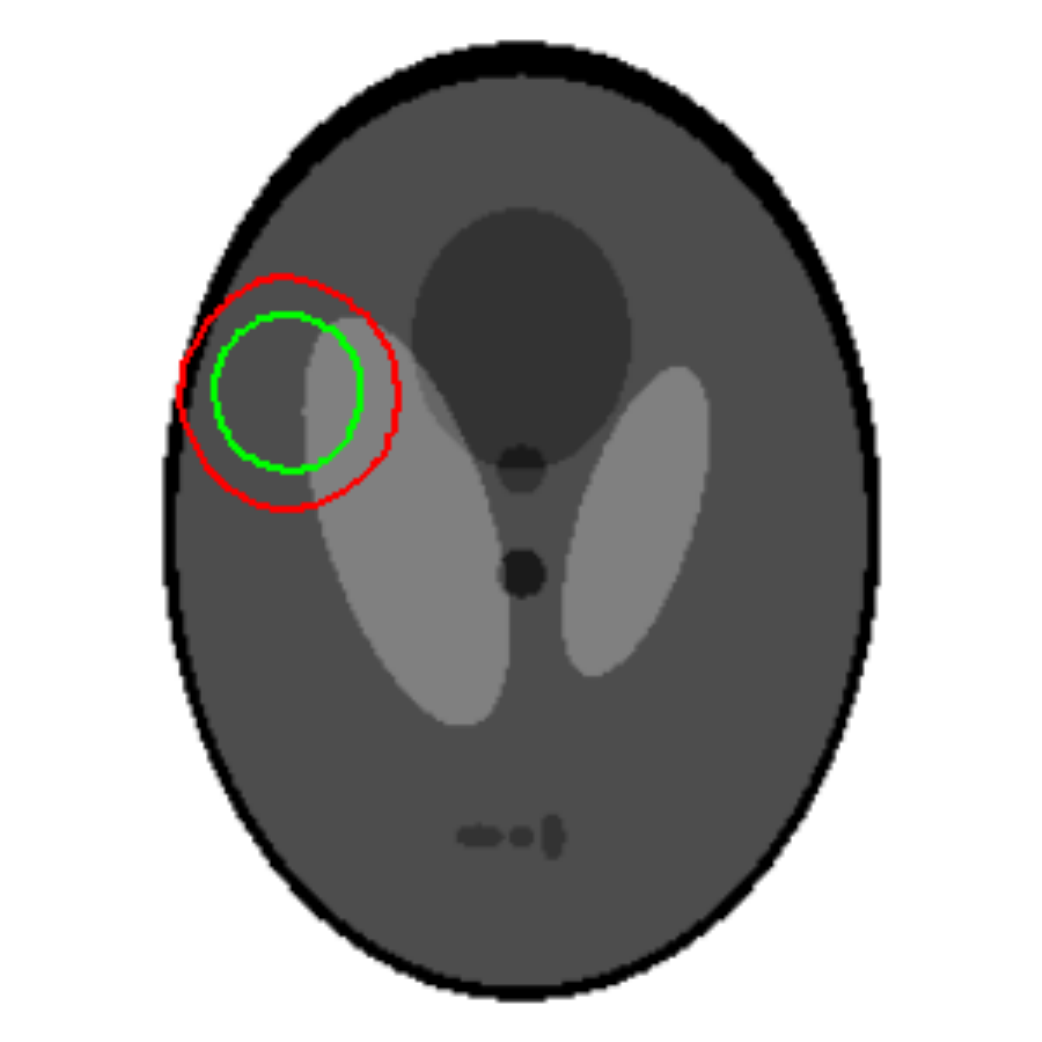}&
\includegraphics[width=0.85in]{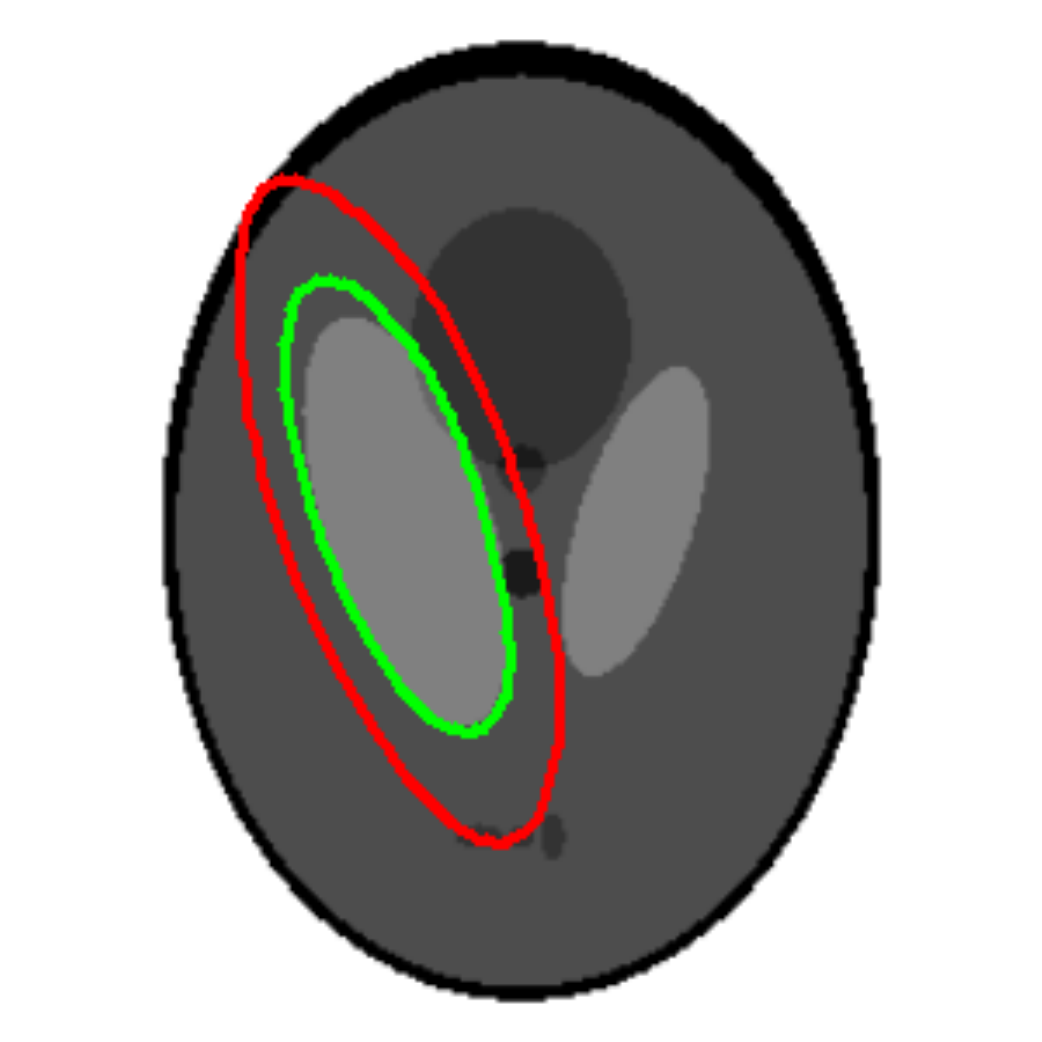}\\
\includegraphics[width=0.85in]{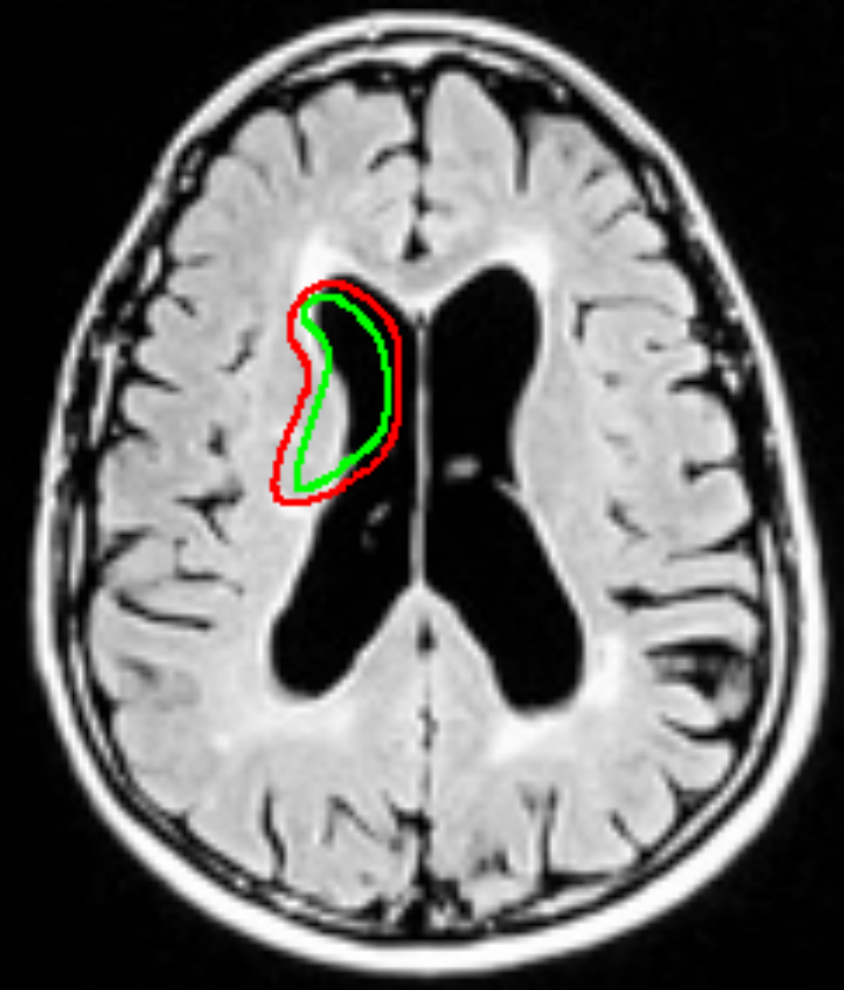}&
\includegraphics[width=0.85in]{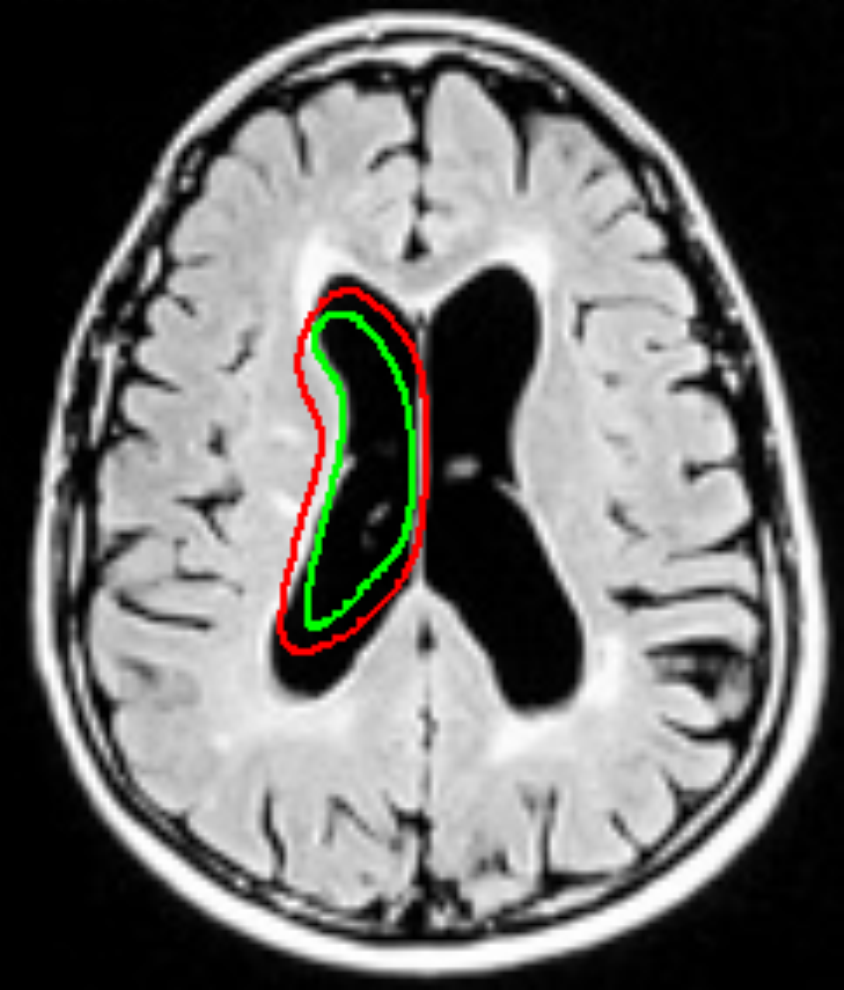}&
\includegraphics[width=0.85in]{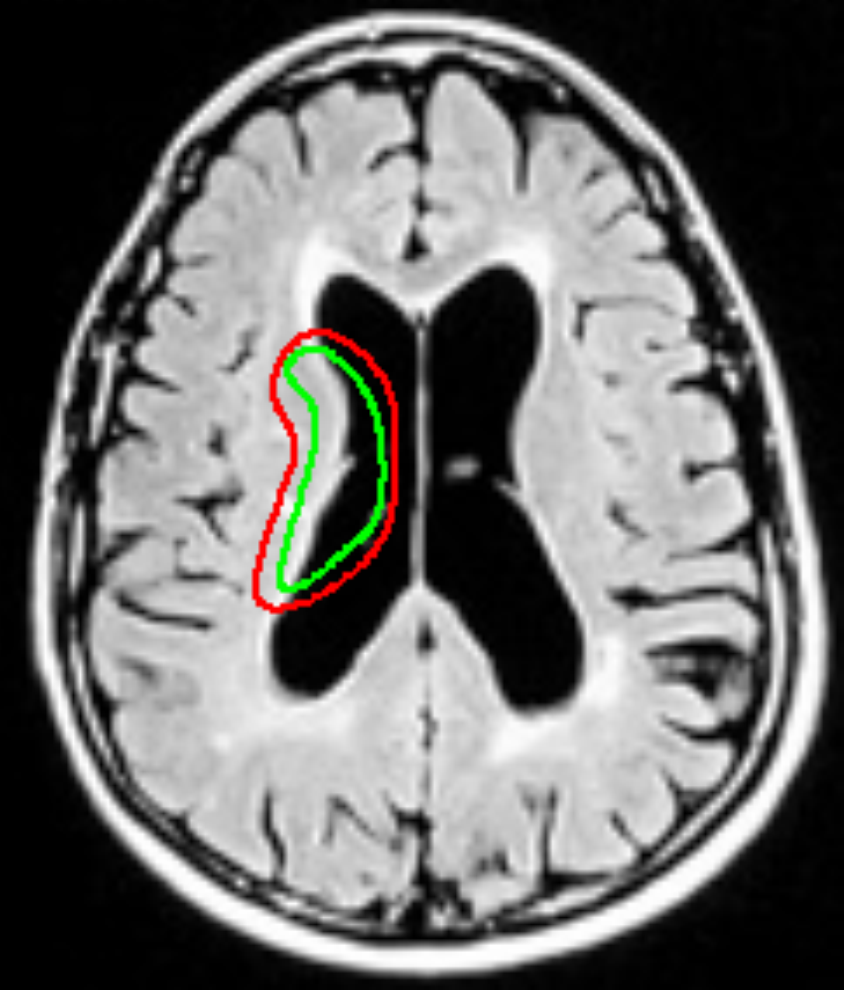}&
\includegraphics[width=0.85in]{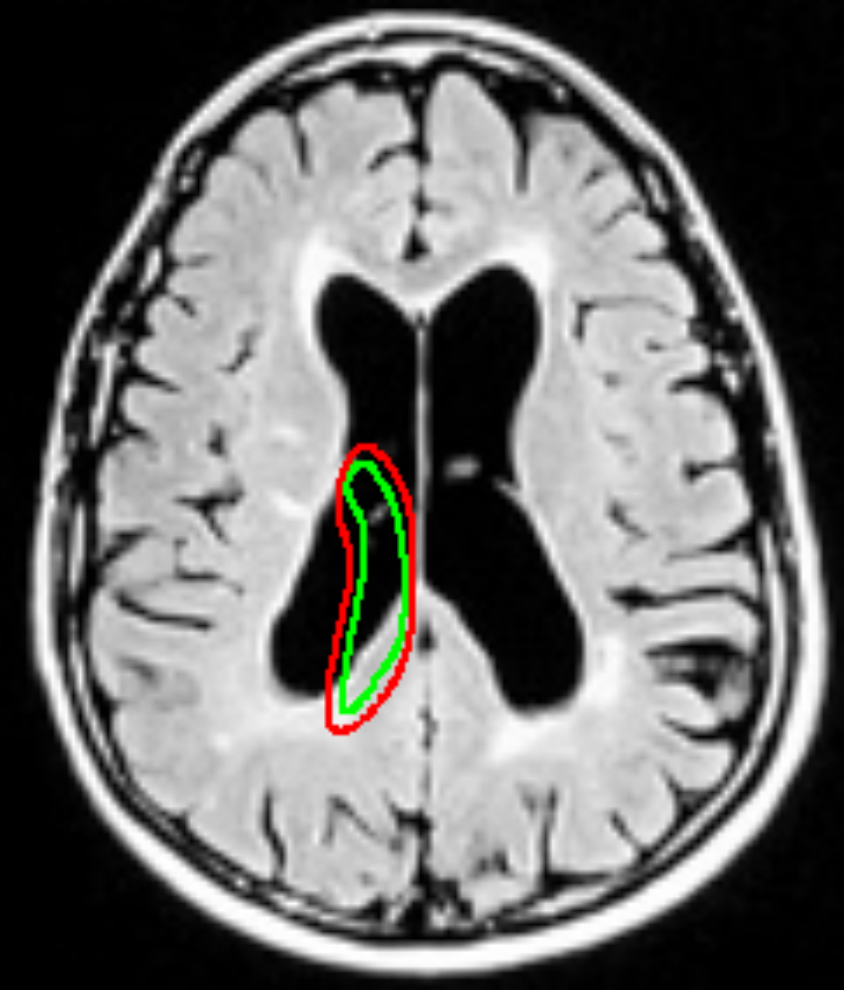}&
\includegraphics[width=0.85in]{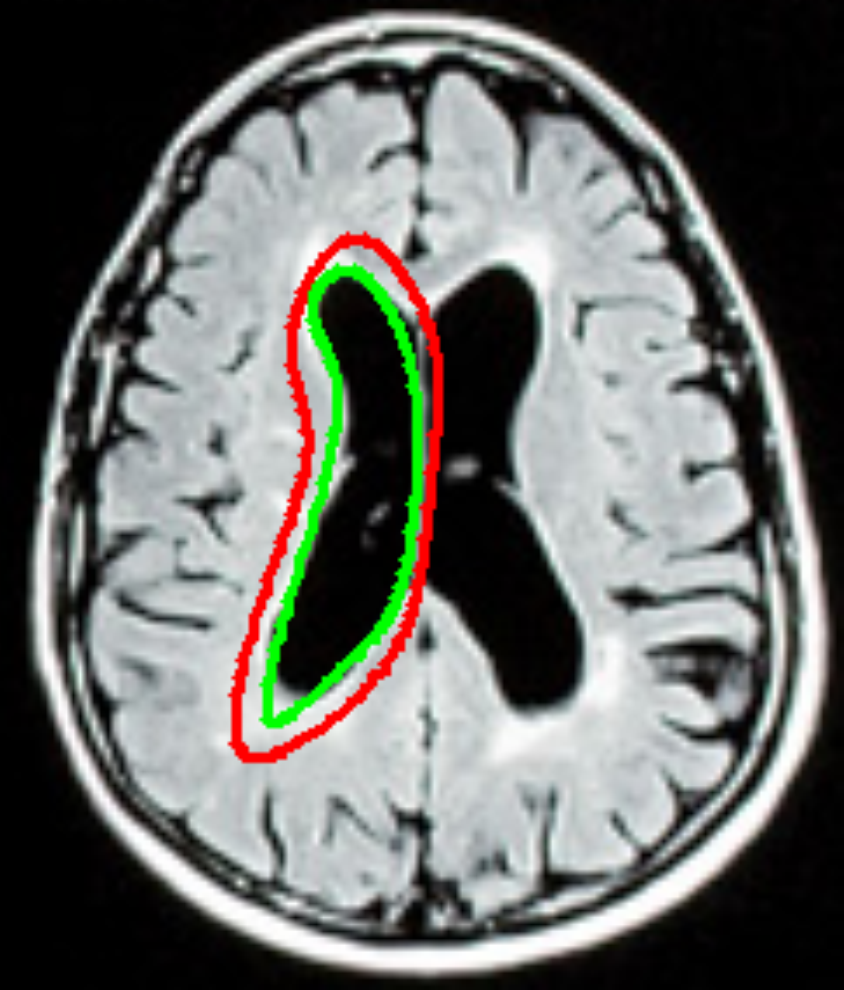}&
\includegraphics[width=0.85in]{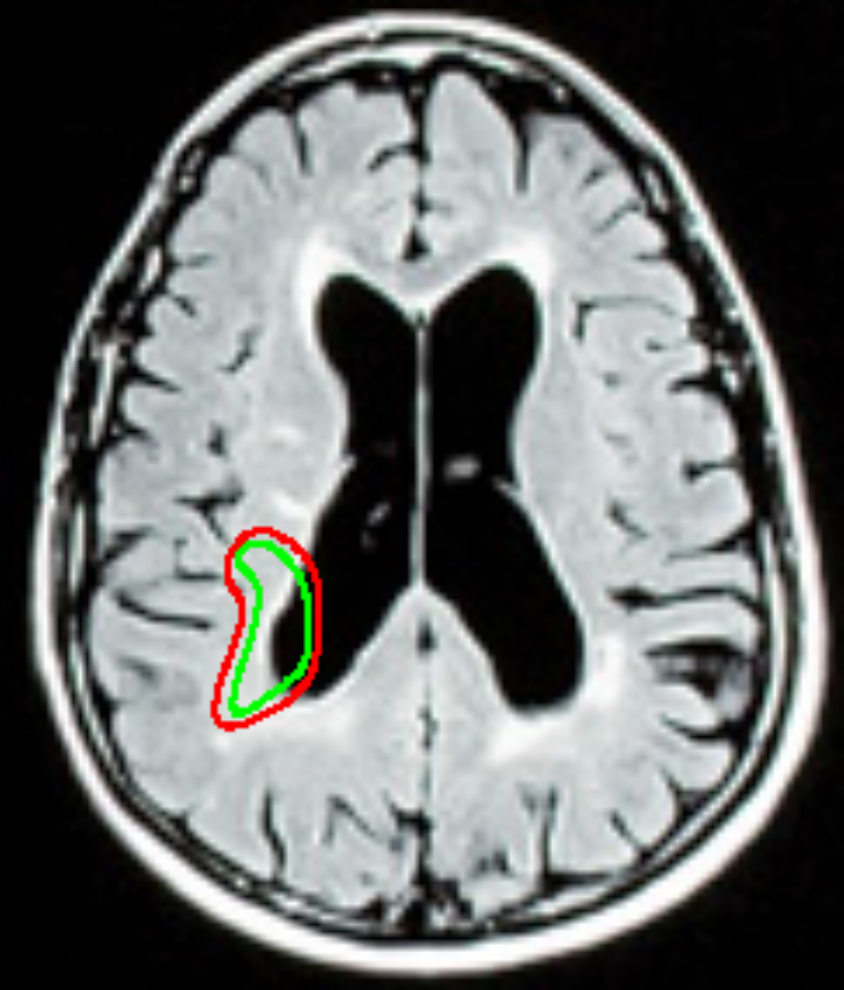}&
\includegraphics[width=0.85in]{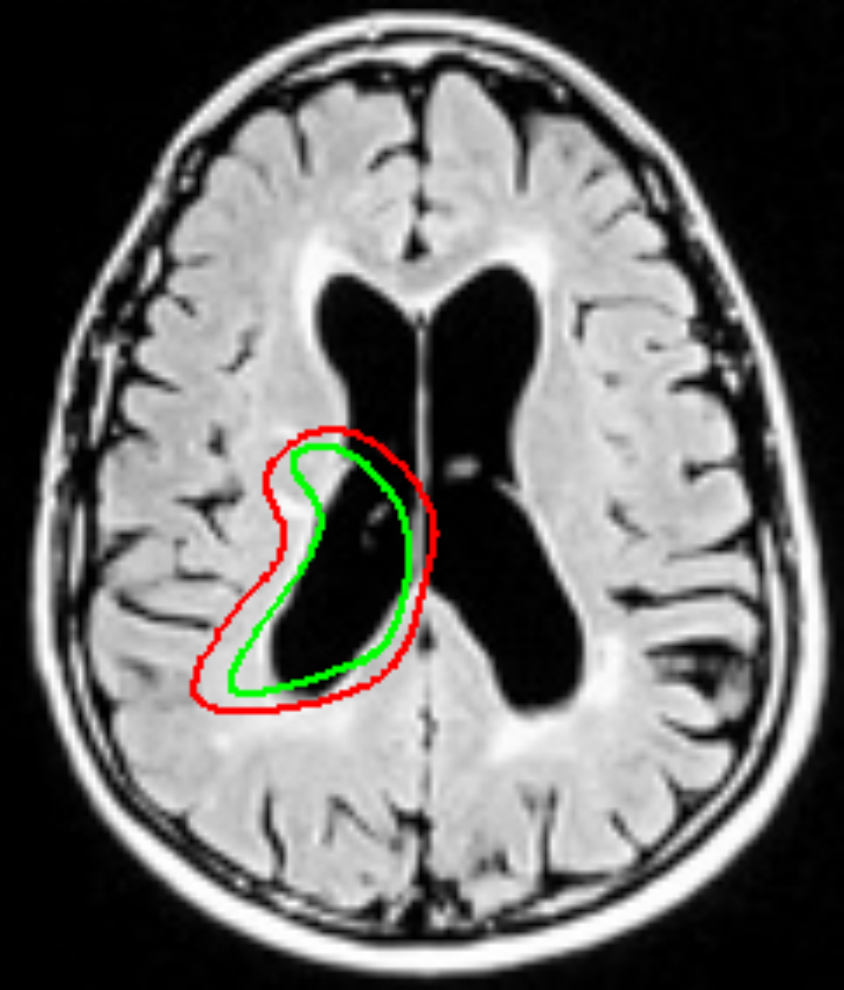}\\
 (a) & (b) &(c)&(d)&(e)&(f)&(g)\\
\end{array}
$
\caption{Robustness to initialization: These images illustrate that reasonably good initializations such as those shown in $(a)-(d)$ converge to $(e)$, whereas a poor initialization such as the one shown in $(f)$ converges to a suboptimal solution $(g)$.}
\label{initphantom}
\end{figure*}

\indent Consider a Shepp-Logan phantom image \cite{Shepp} shown in Figure~\ref{initphantom}, which is widely used for validation of computed tomography reconstruction algorithms. The initializations shown in Figure~\ref{initphantom}(a)--(d) converge to the optimal solution shown in (e), whereas a poor initialization (as shown in Figure~\ref{initphantom}(f), where the outer contour overlaps with the dark boundary) leads to slightly suboptimal solution shown in (g). Also, as expected, a good initialization (Figure~\ref{initphantom}(b)) results in faster convergence. The initializations shown in (a)--(d) converged to (e) in 121, 71, 74, and 77 milliseconds, respectively. The execution times are given with respect to an ImageJ implementation of the algorithm running on Mac~OSX 2.66 GHz, Intel Pentium Dual CPU T3400 @2.16GHz.
A similar set of experiments for a non-convex shape was carried out using MR images shown in the second row of Figure~\ref{poissonnoise}. Since the desired object of interest in this case is dark and the surrounding background is bright, we negate the energy function $E_g$. The convergence times in milli-seconds for the initializations shown in (a)-(d) are 110, 78, 95, and 83, respectively. We attribute the fast convergence to the local nature of snake energy and direct computation of line integrals in (\ref{matrixform}) and (\ref{matrixform2}).
\subsection{Robustness to Poisson and Gaussian noise}
\begin{figure*}[t]
\centering
$
\begin{array}{cccccc}
\begin{turn}{90}{Gaussian\,Noise} \end{turn}&
\includegraphics[width=1.2in]{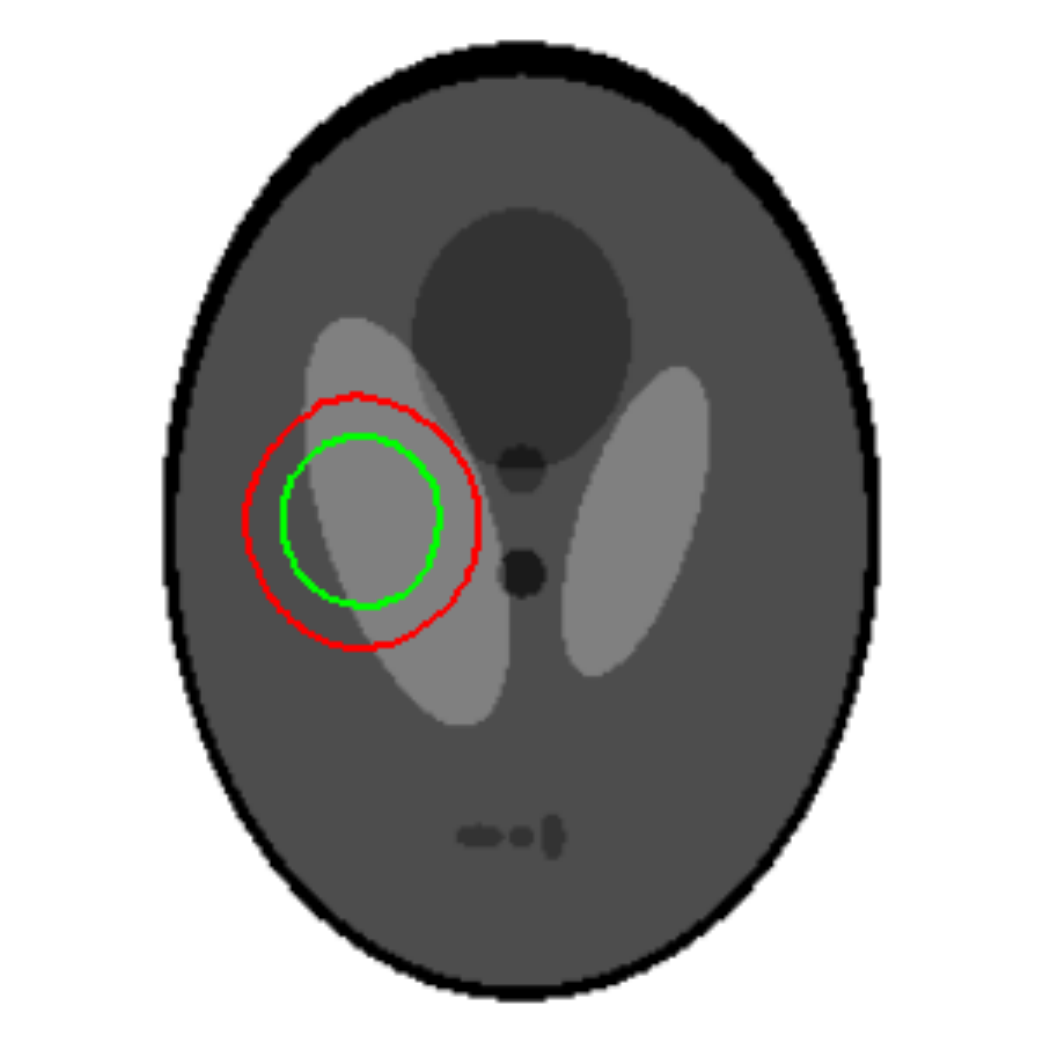} &
\includegraphics[width=1.2in]{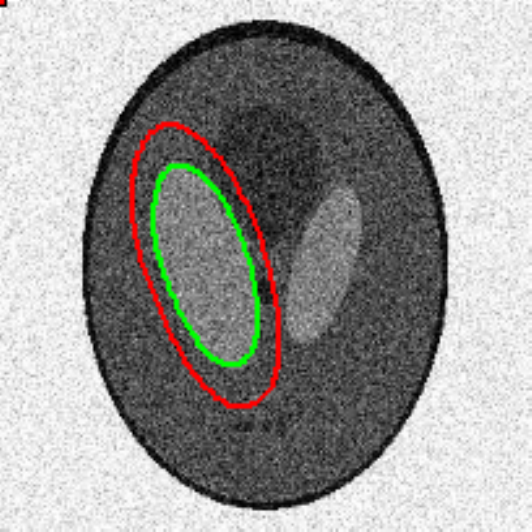} &
\includegraphics[width=1.2in]{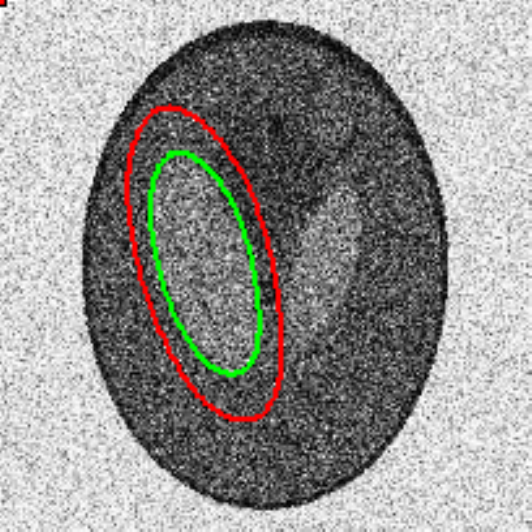} &
\includegraphics[width=1.2in]{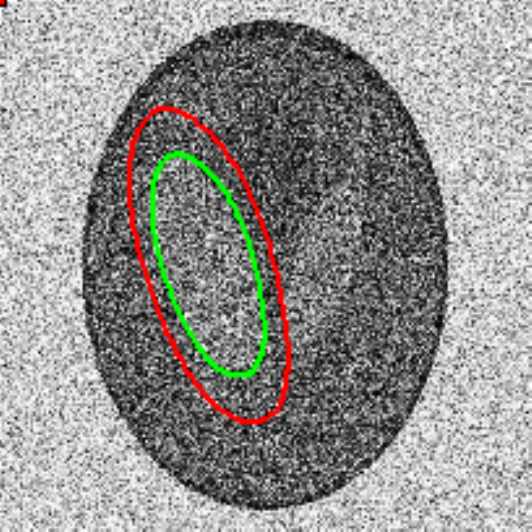} &
\includegraphics[width=1.2in]{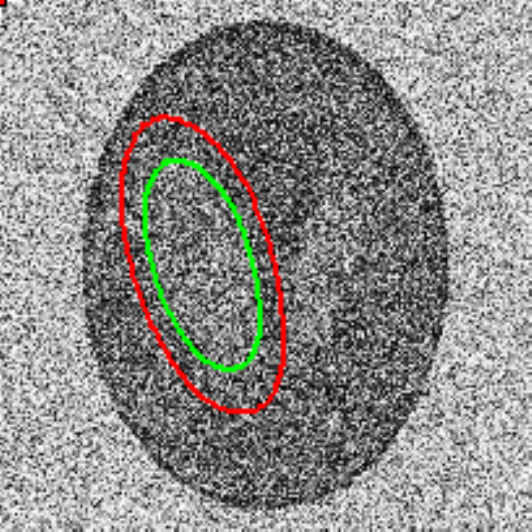}\\
& \mbox{Initialization} & 20.17\,\mbox{dB} & 14.15\,\mbox{dB} & 10.63\,\mbox{dB} & 8.13\,\mbox{dB} \\
\begin{turn}{90}{Poisson\,Noise} \end{turn}&
\includegraphics[width=1.2in]{poisson_init.pdf} &
\includegraphics[width=1.2in]{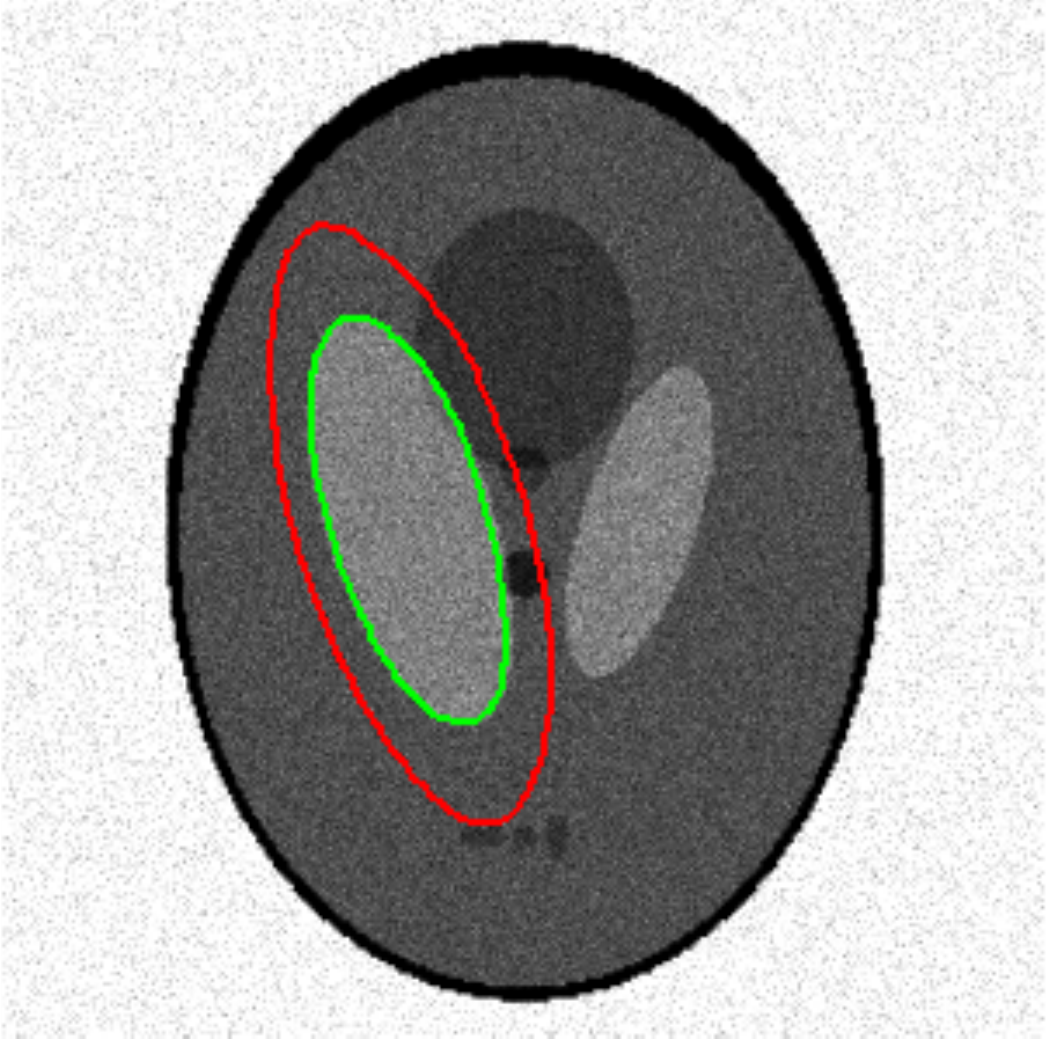} &
\includegraphics[width=1.2in]{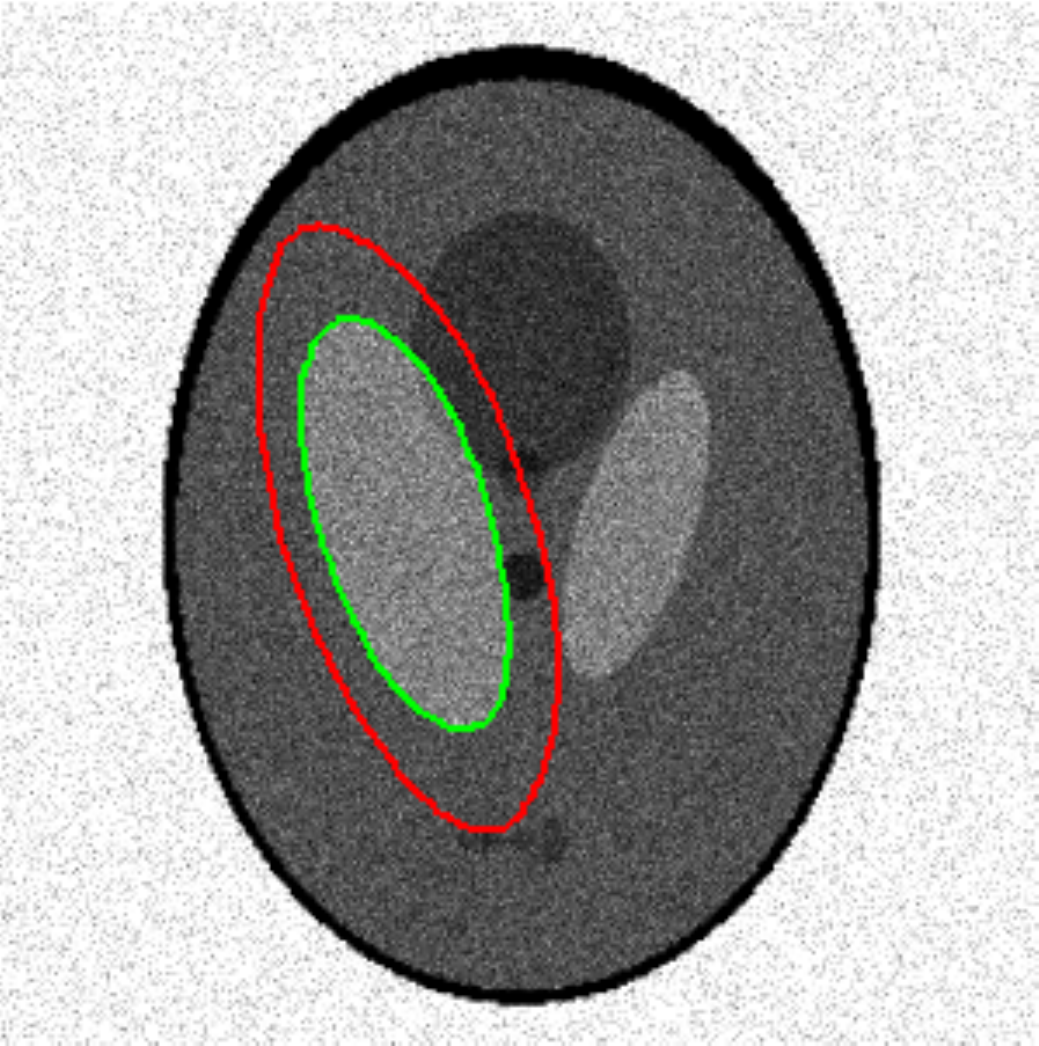} &
\includegraphics[width=1.2in]{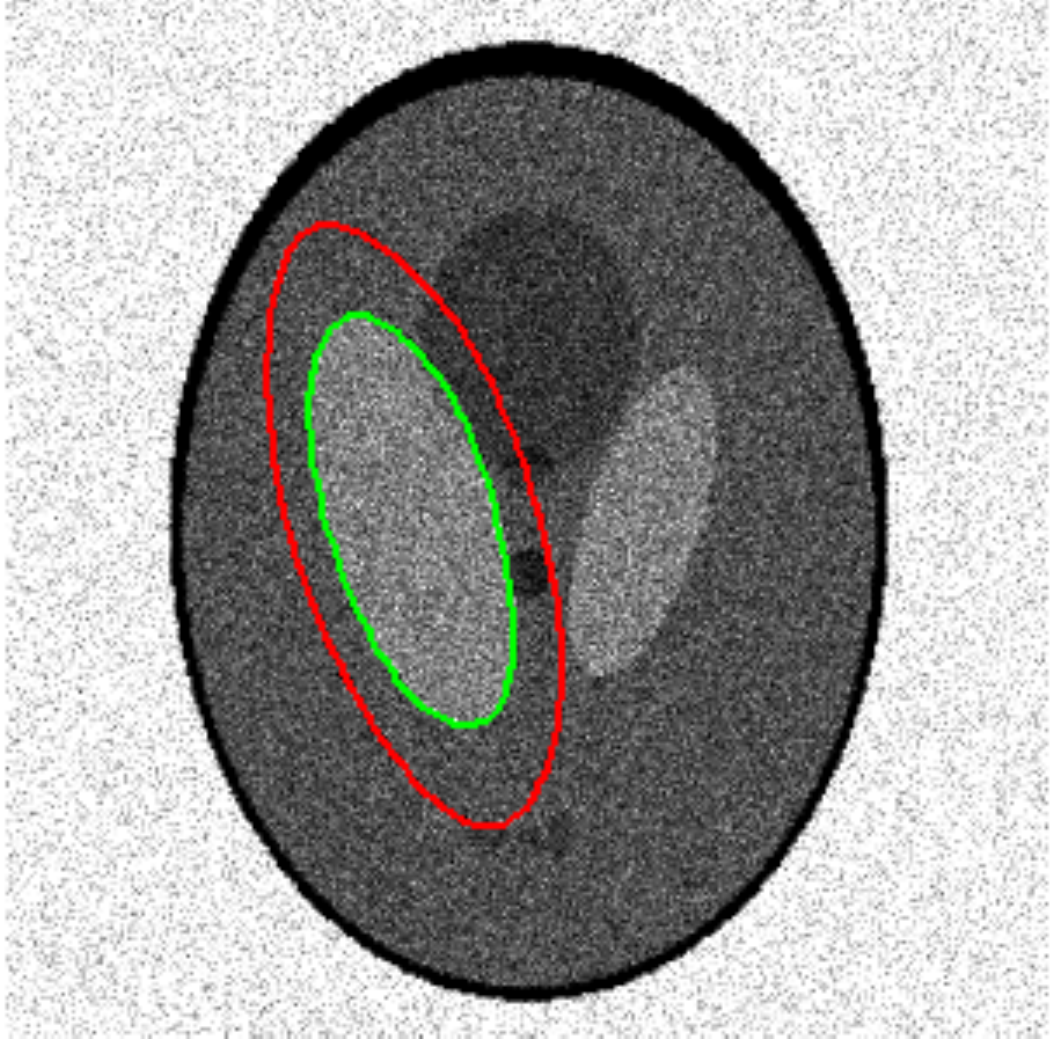} &
\includegraphics[width=1.2in]{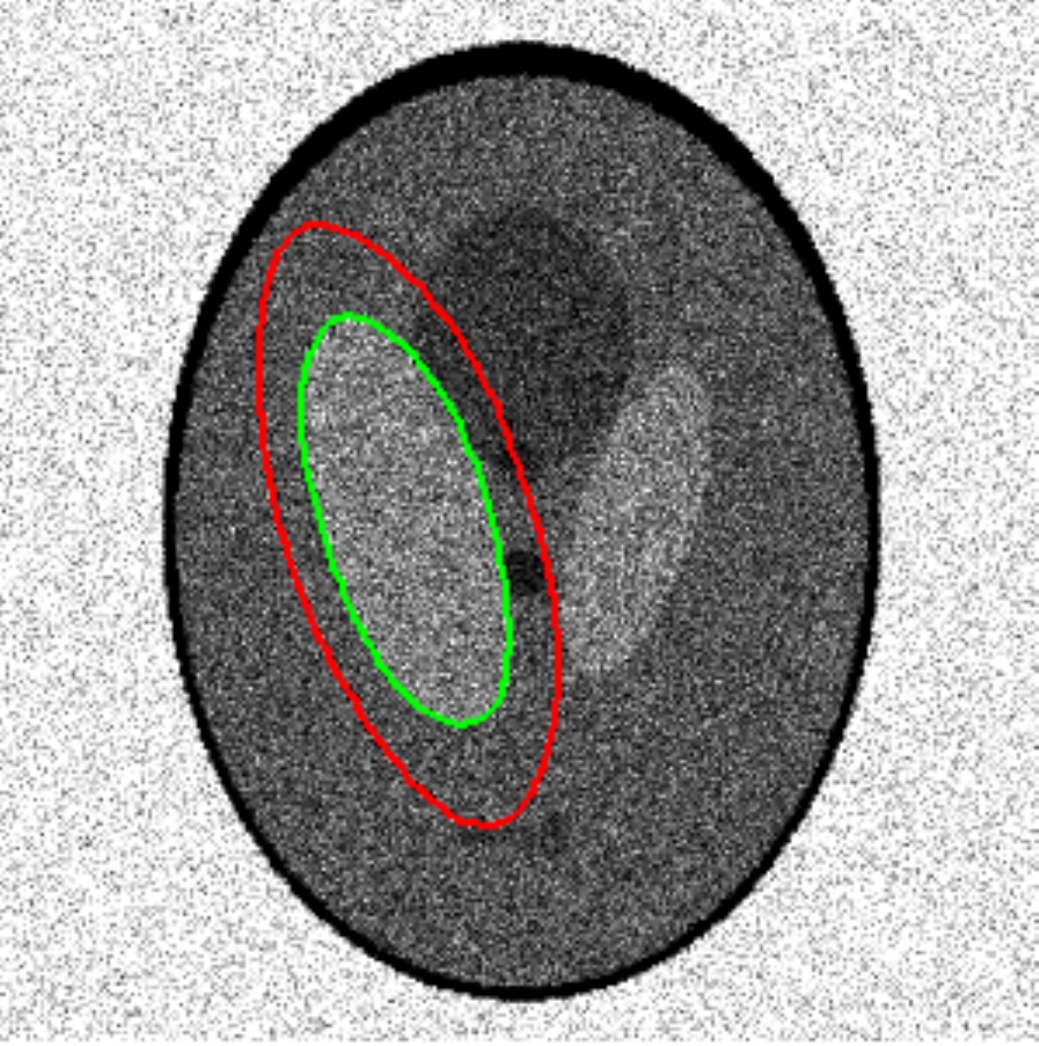}\\
& \mbox{Initialization} & 26.92\,\mbox{dB} & 23.32\,\mbox{dB} & 19.70\,\mbox{dB} & 17.78\,\mbox{dB} 
\end{array}
$
\caption{Robustness to noise: Top row: Gaussian noise; bottom row: Poisson noise. The numbers indicate PSNRs.}
\label{poissonnoise} 
\end{figure*}

\indent We validate the performance of the proposed technique on noisy images, by considering two types of noise distributions --- Gaussian noise and Poisson noise. Gaussian noise is independent of the image data, additive and stationary, whereas Poisson noise is dependent on the image data and is multiplicative. Images in the top row of Figure \ref{poissonnoise} (Shepp-Logan phantom \cite{Shepp}) are corrupted with Gaussian noise and the images in the bottom row with Poisson noise. Standard deviations of the Gaussian noise added to the four images are 25, 50, 75, and 100 respectively. Peak signal-to-noise ratios (PSNRs) computed therefrom are indicated in the figure captions. For Gaussian noise images, we used the standard PSNR definition given in \cite{gonzalezwoods}, and for Poisson noise images, the definition given in \cite{florian}. Section \ref{FischerInterpretation} gives theoretical guarantees for gaussian noise corrupted images, but from the experimental results, we observe that performance of active contour technique is robust even to Poisson noise.

\begin{figure}
\centering
$
\begin{array}{cc}
\mbox{Initialization} & \mbox{Converged Contour} \\
\includegraphics[width=1.55in]{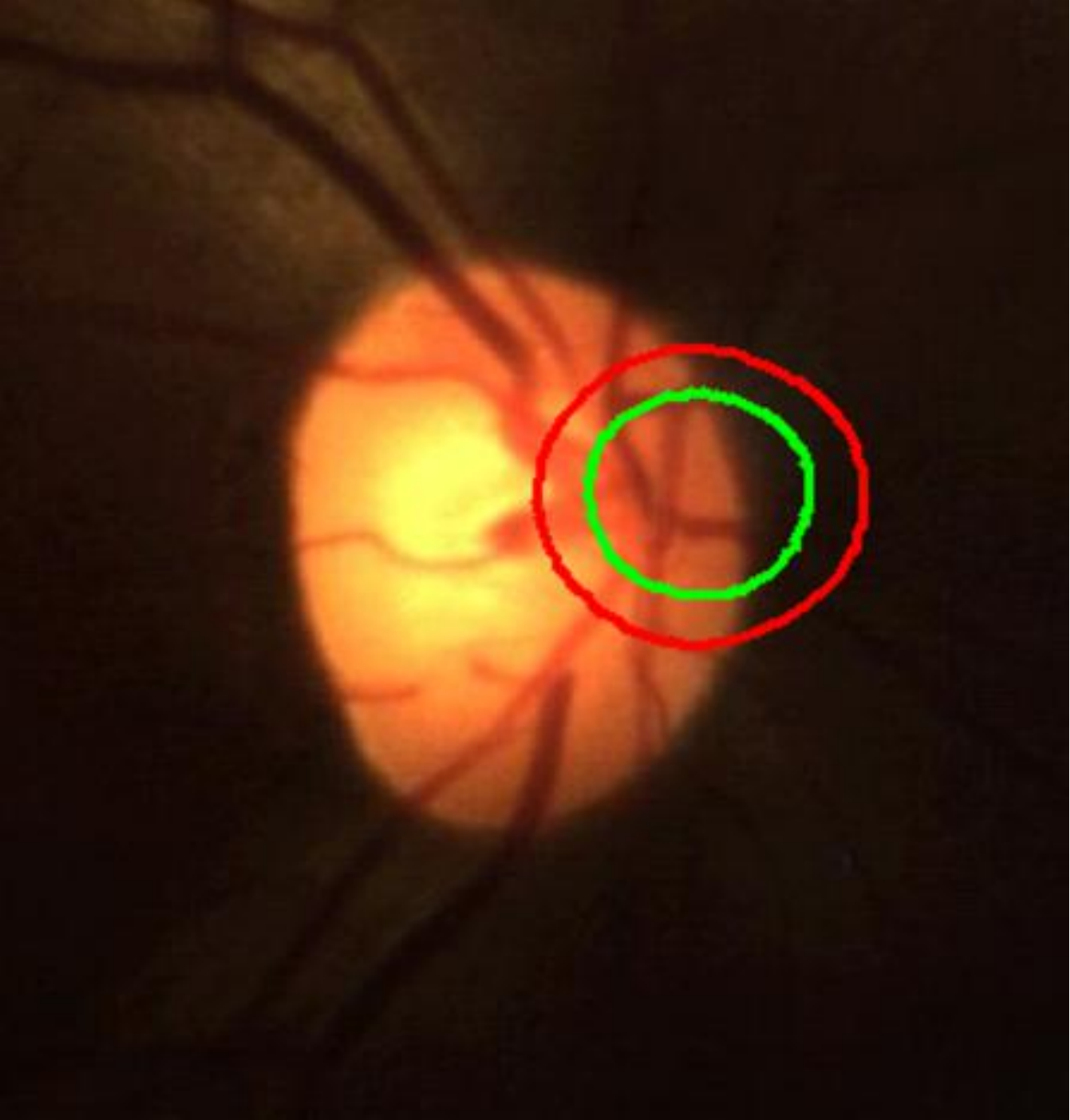} &
\includegraphics[width=1.55in]{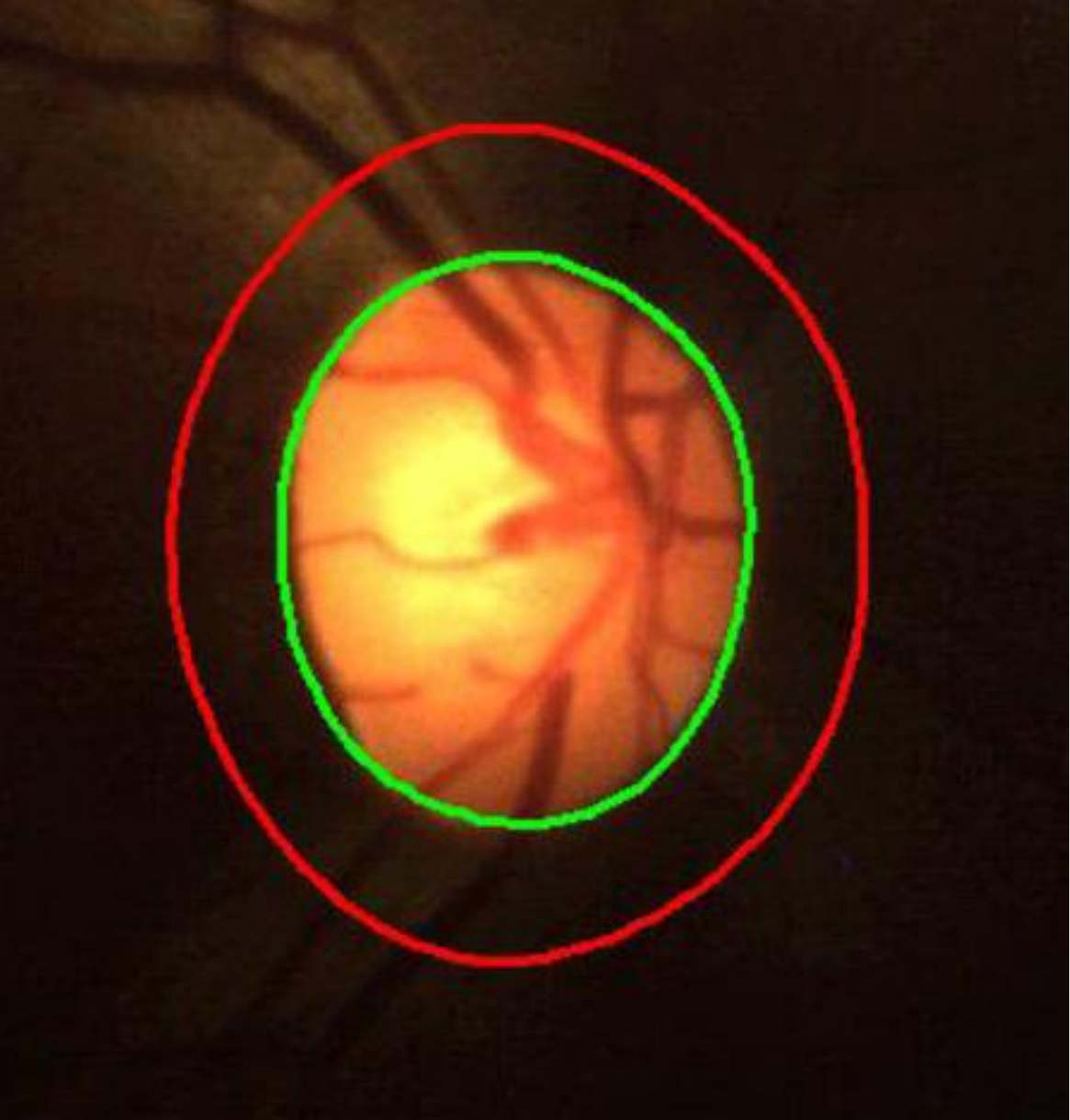}\\
\includegraphics[width=1.55in]{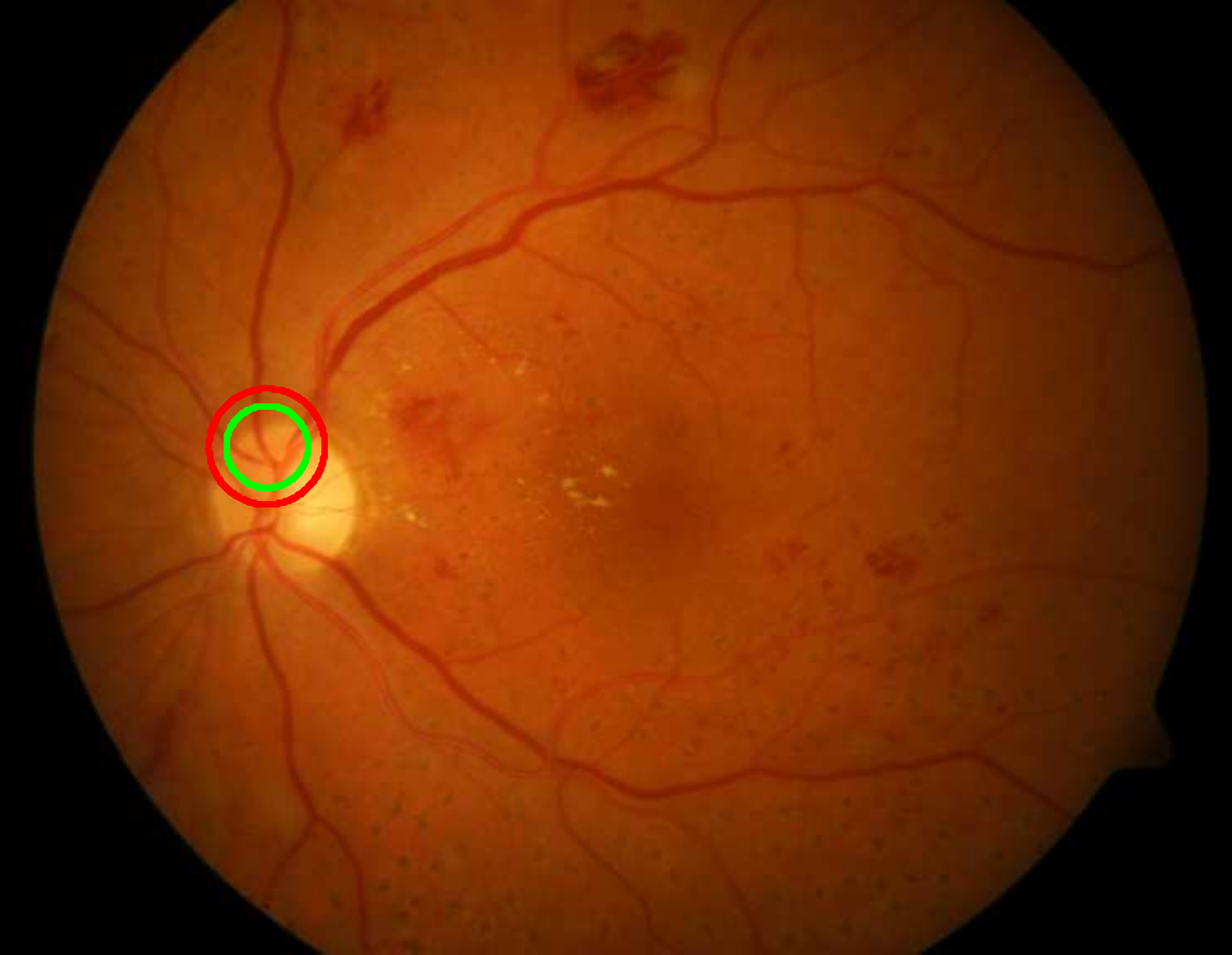} &
\includegraphics[width=1.55in]{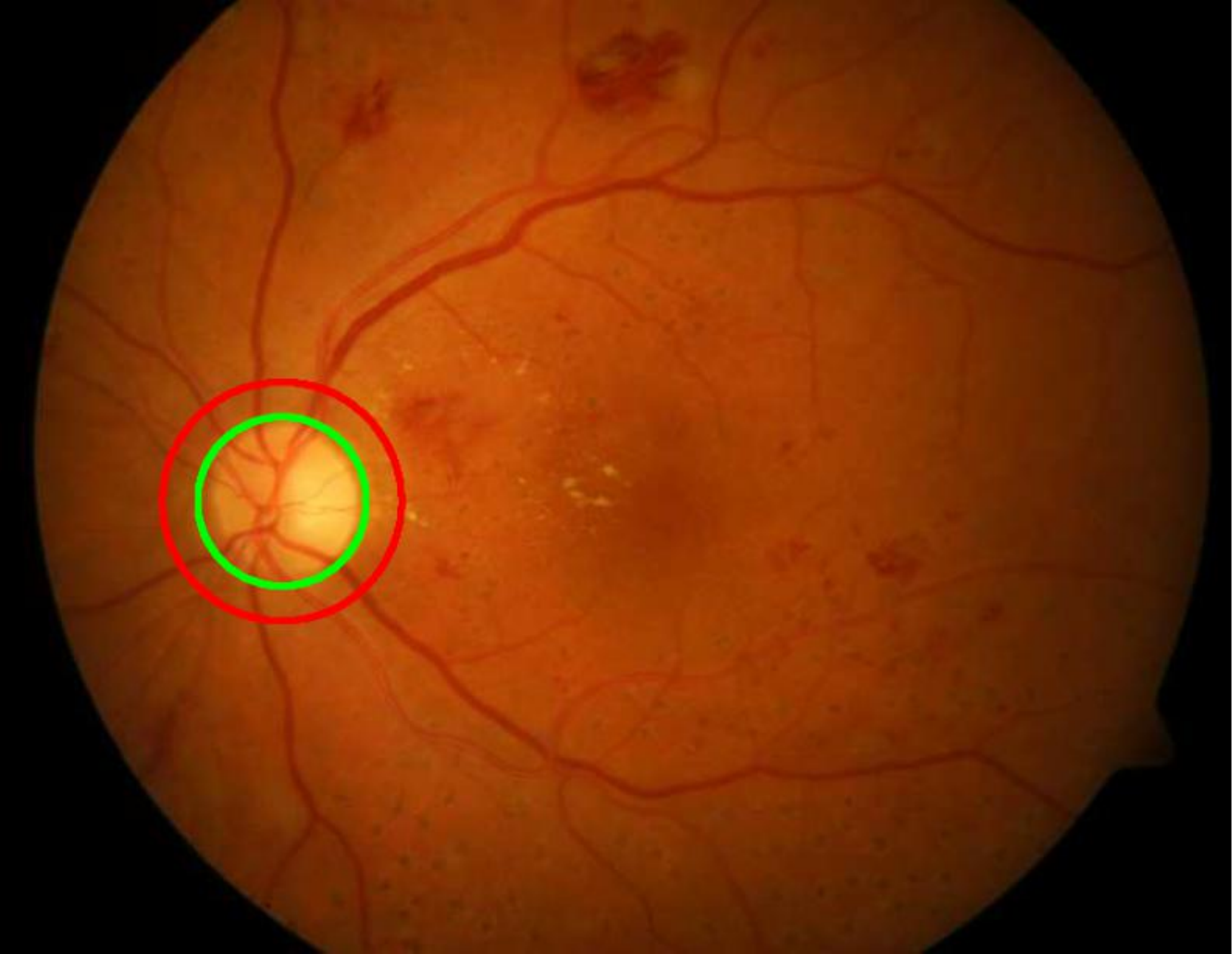}\\
\end{array}
$
\caption{Results on fundus images: Left column shows initialization and right column shows the converged contour.}
\label{retina}
\end{figure}
\begin{figure}
\centering
$
\begin{array}{cc}
\mbox{Initialization} & \mbox{Converged Contour} \\
\includegraphics[width=1.32in]{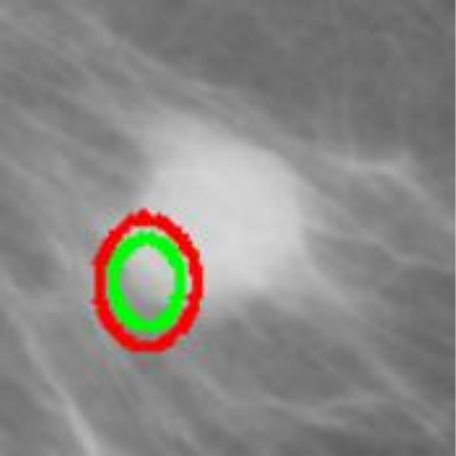} &
\includegraphics[width=1.32in]{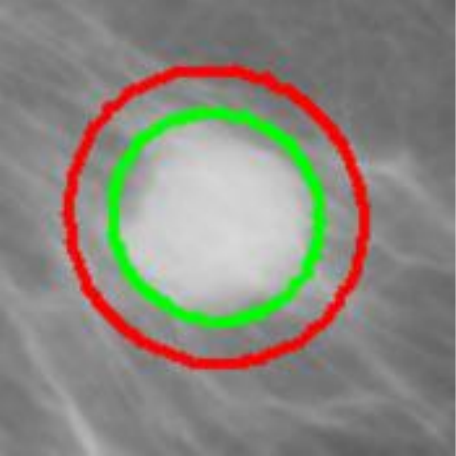} \\
\includegraphics[width=1.32in]{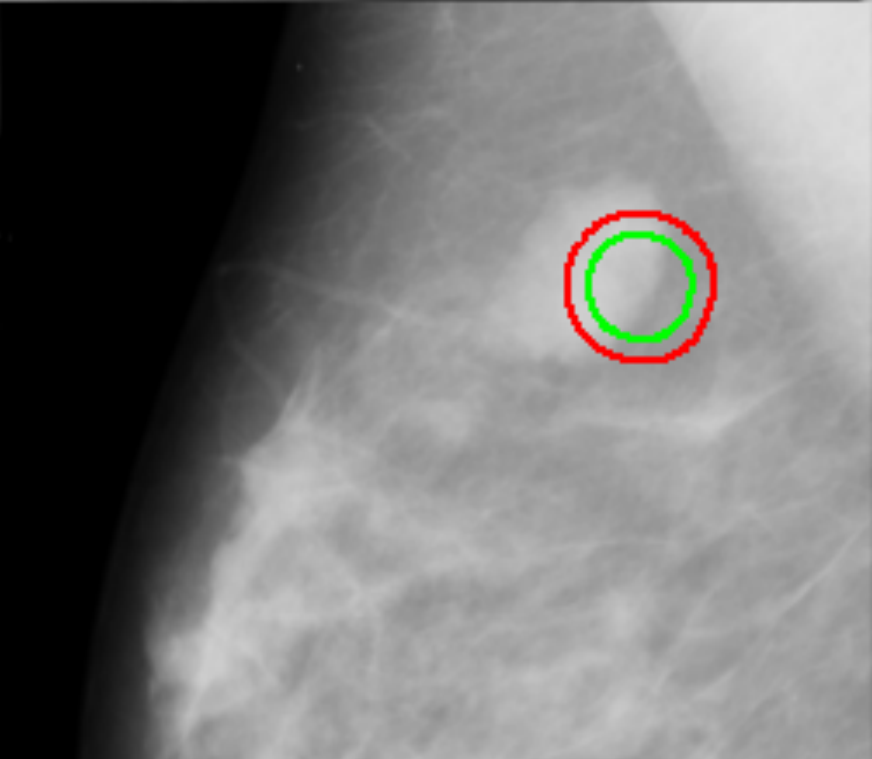} &
\includegraphics[width=1.32in]{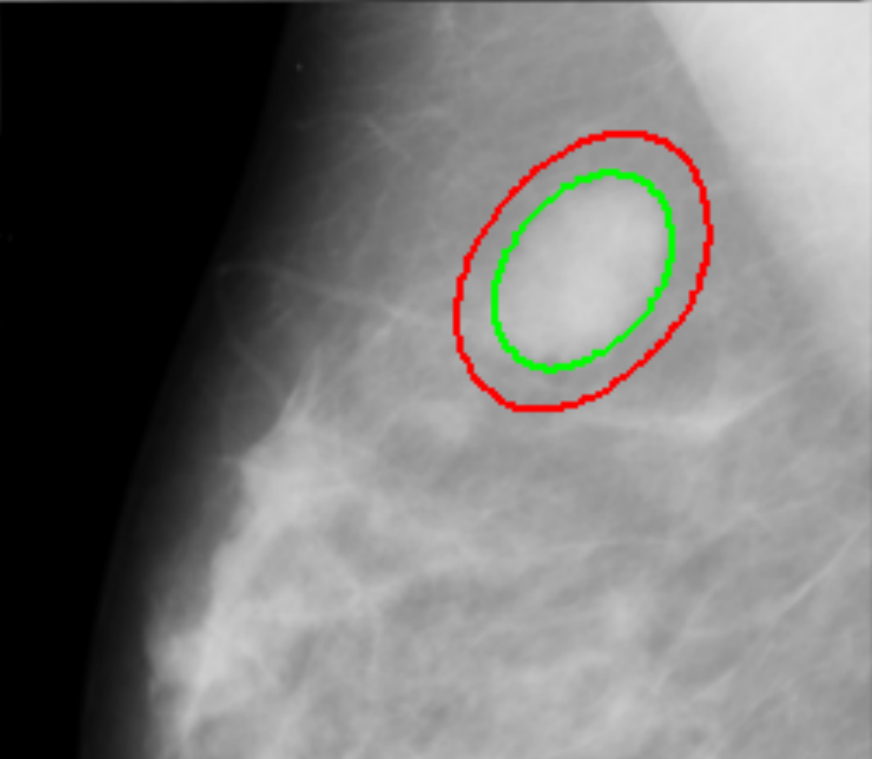} \\
\end{array}
$
\caption{Localization of micro-calcifications in mammogram images taken from \cite{minimias}. Left column shows initialization with elliptical templates, and right column shows the converged contours. Figures are cropped to show the region of interest.}
\label{DROSO} 
\end{figure}
\begin{figure*}[t]
\centering
$
\begin{array}{cccc}
\includegraphics[width=1.55in]{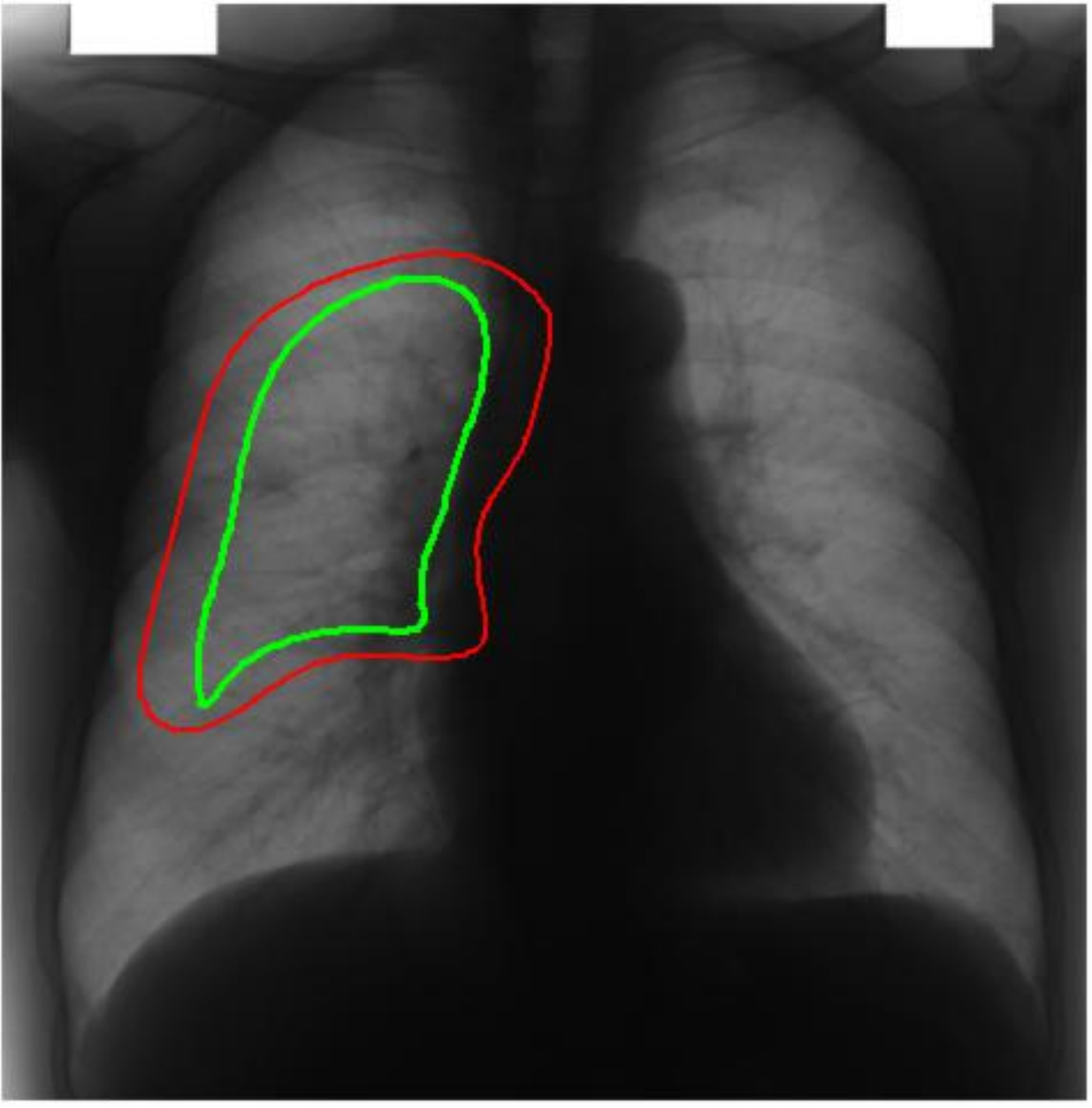} &
\includegraphics[width=1.55in]{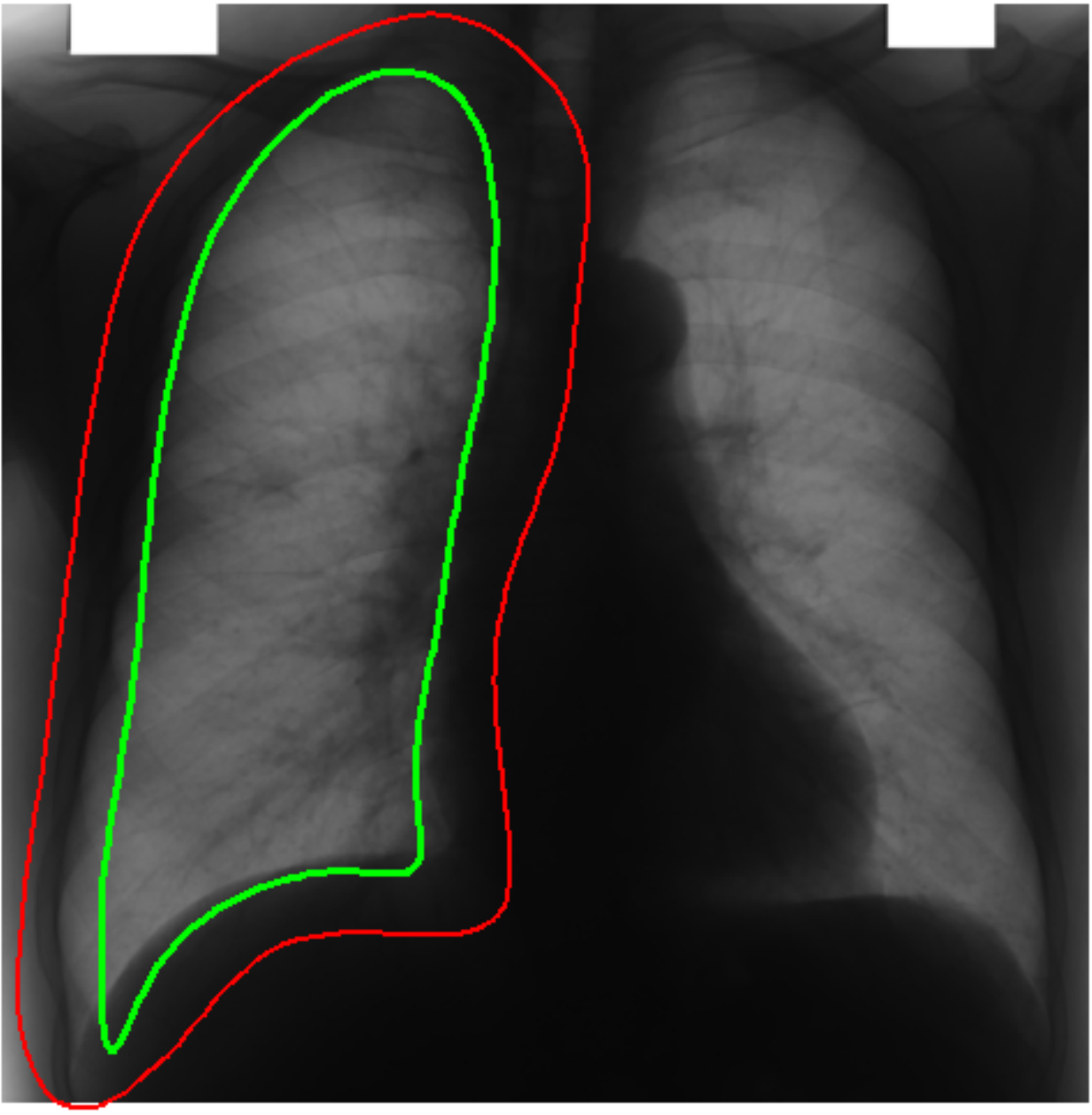}&
\includegraphics[width=1.55in]{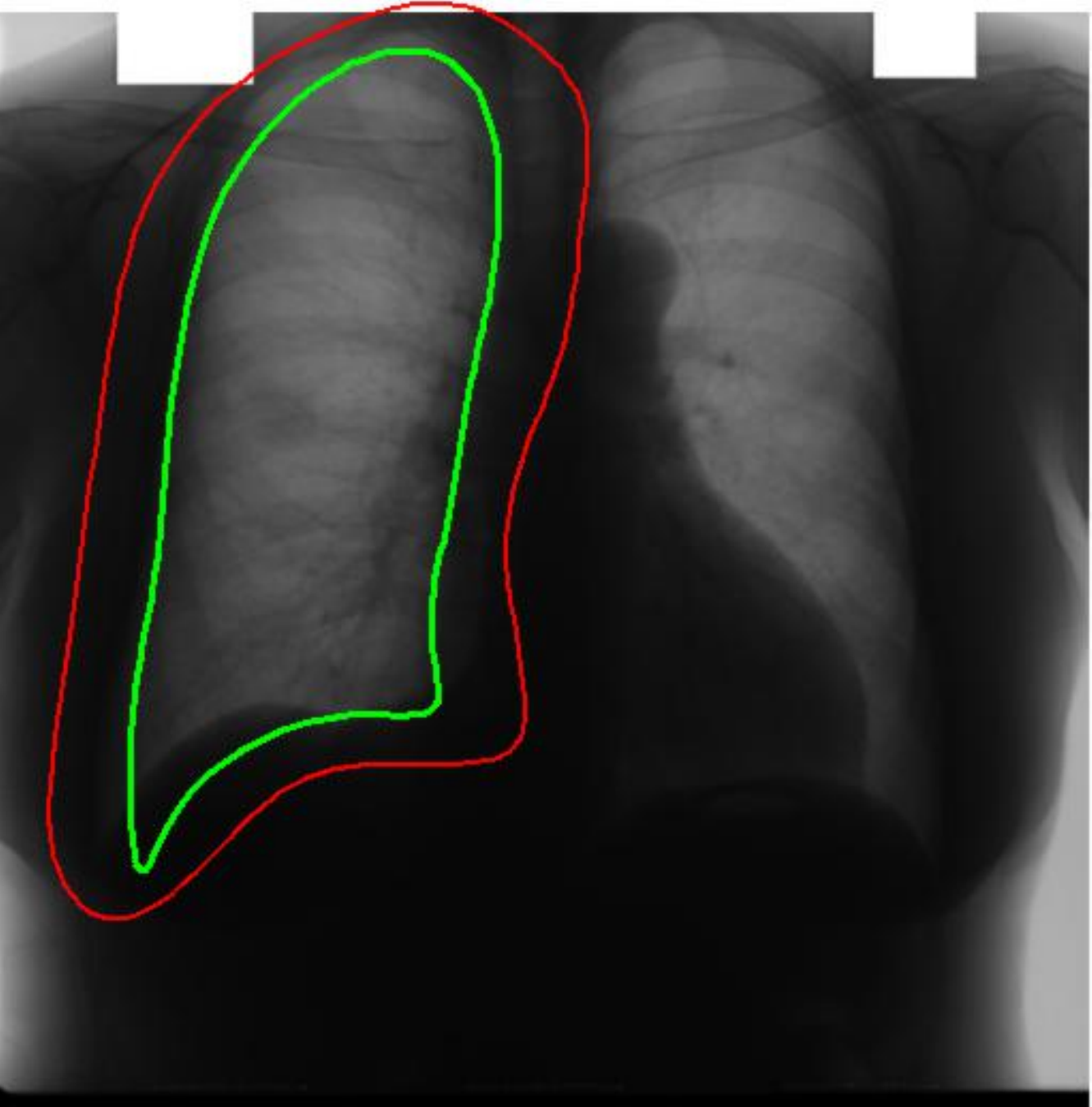} &
\includegraphics[width=1.55in]{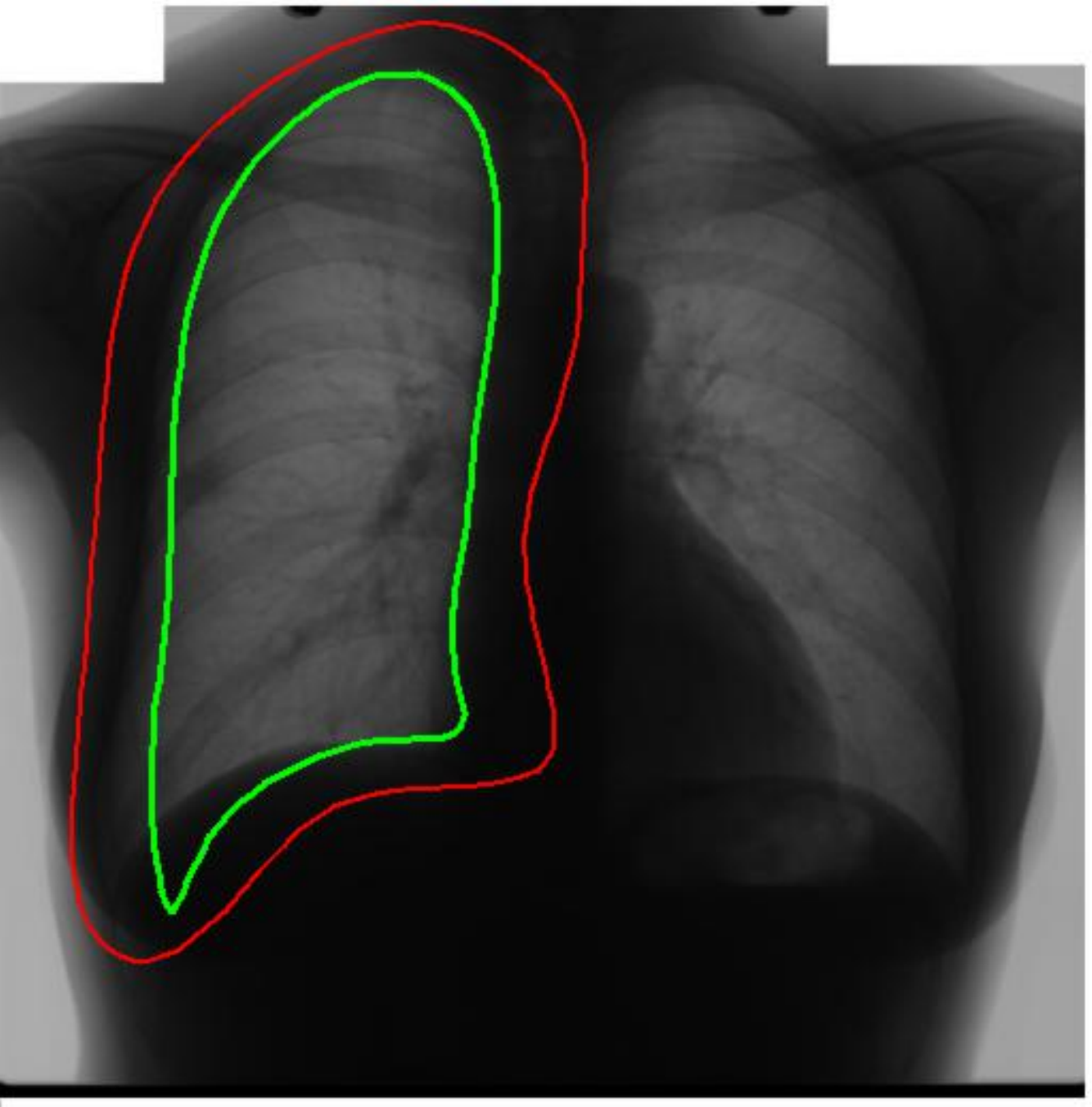}\\
(a) & (b) & (c) & (d)
\end{array}
$
\caption{Segmentation of lungs using the mean template: (a) Initialization of template, (b) converged result. For the result shown in (c) and (d), an initialization similar to that shown in (a) was provided.}
\label{Lungs}
\end{figure*}
\begin{figure*}[t]
\centering
$
\begin{array}{ccc}
\includegraphics[width=2.15in]{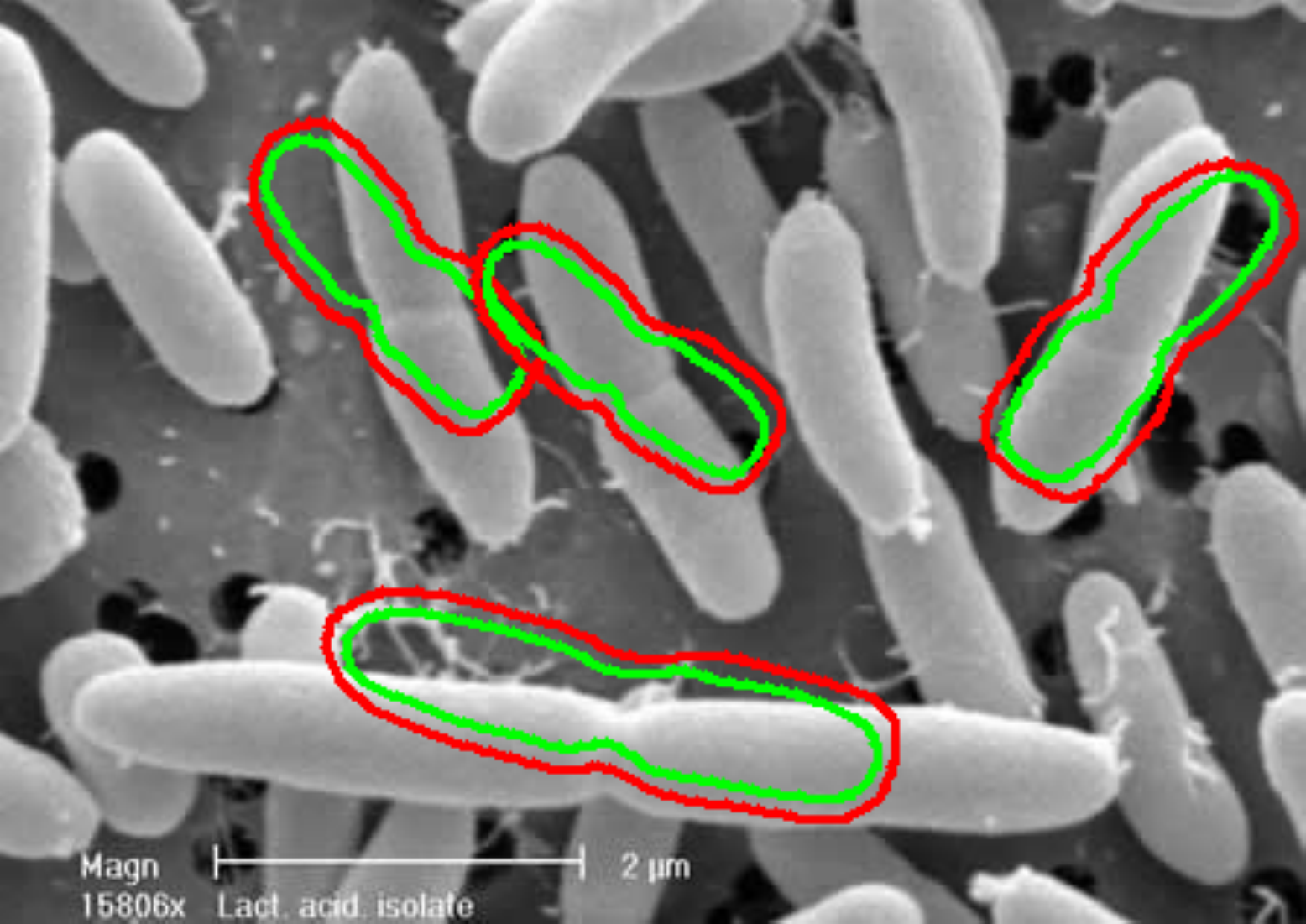} &
\includegraphics[width=2.15in]{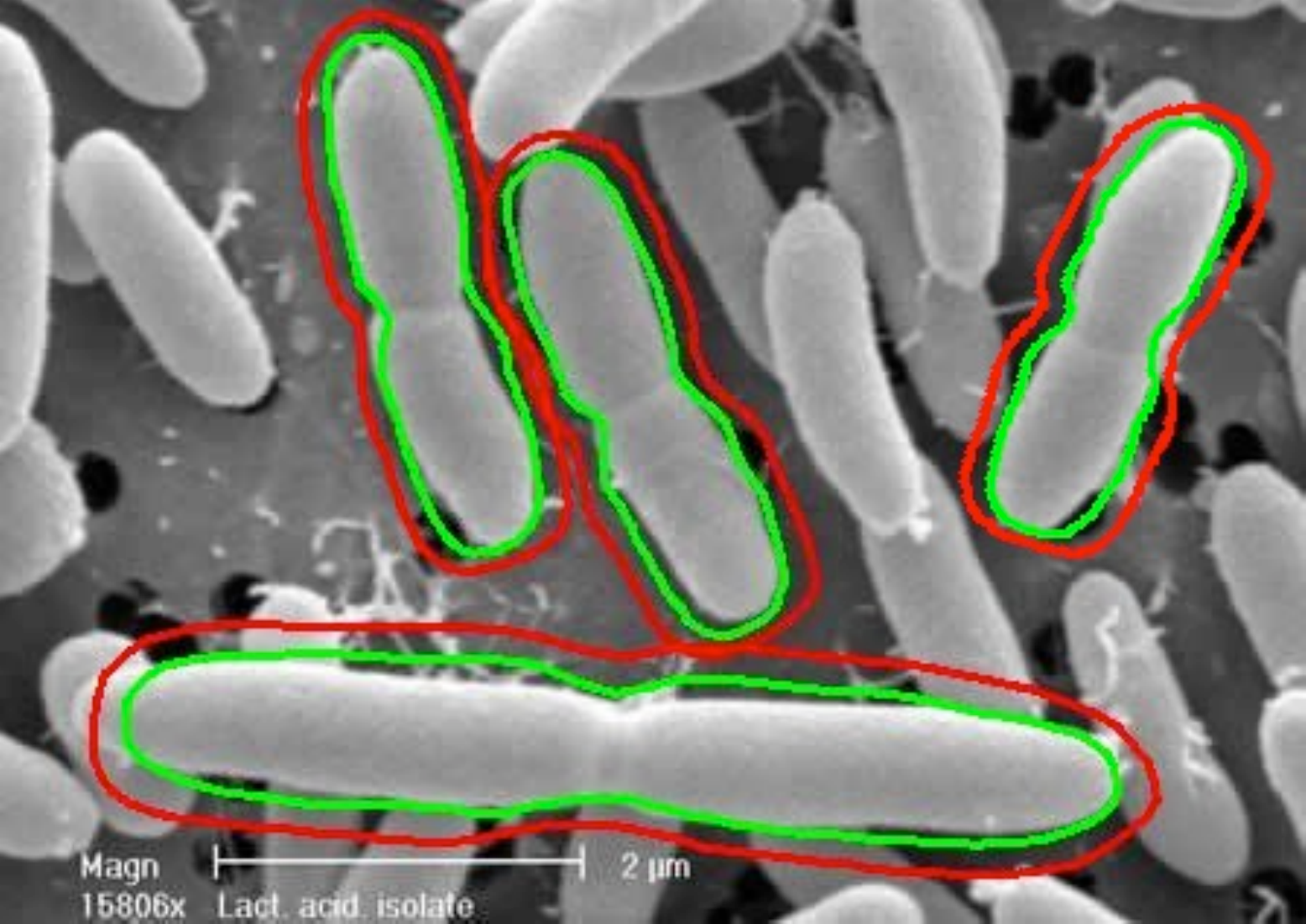} &
\includegraphics[width=2.15in]{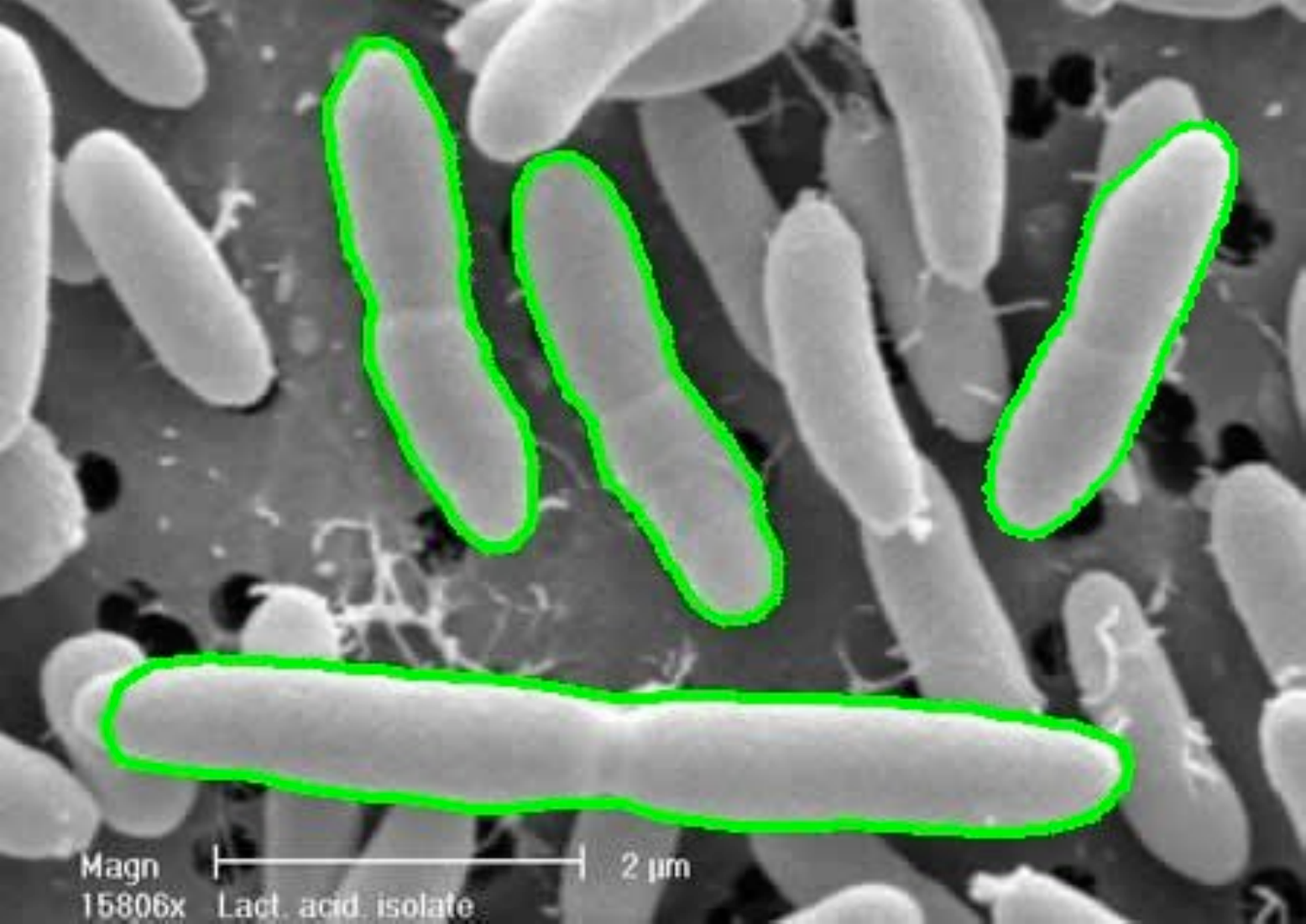} \\
(a) & (b) & (c)
\end{array}
$
\caption{Segmentation of lactobacilli taken from \cite{lactobacillus}: (a) Initialization (b) converged result using shape based optimization and (c) contour obtained by refining the result in (b) using gradient energies.}
\label{lacto}
\end{figure*}

\subsection{Elliptical templates}
\indent Many of the bio-medical cells can be approximated  with an ellipse \cite{ovuscule}. We present results using two elliptical shaped templates: a spline synthesized ellipse for which we use the partial derivatives given in equations (\ref{matrixform}) and (\ref{matrixform2}) directly, an exact ellipse for which we use the parameterization given in \cite{Adithya1} and use the equation (\ref{final2}). First row of Figure \ref{retina} corresponds to a hand drawn ellipse and the second row correspond to a parametric ellipse. We considered these two parameterizations to show that the algorithm performance is not critically dependent on the parameterization. These are fundus images taken from \cite{fundus}. The presence of veins in the fundus anatomy obscures snakes whose energies rely on image derivative functions. From the results shown in Figure \ref{retina}, we infer that the proposed technique is less affected by the vein structures or the parameterization and captures the near elliptical shape of the fundus outline. \\
\indent We next conducted an experiment on a mammogram taken from the mini-MIAS database \cite{minimias}, which contains some benignant and malignant tumor images. The images (filename: {\tt mdb028}) containing lumps as shown in Figure~\ref{DROSO}, were specifically classified in the database as circular in shape with a certain center and radius. We employed the elliptical shape template to localize the mass and measured the resulting dimensions of the fit. For the top row, the converged ellipse fit was found to have a semi-major and semi-minor axes of 58 and 56 pixels, respectively. The radius of the approximate circumscribing circle was given in the database as 56 pixels. The center of converged contour matches the one given in the database. For the bottom row, semi-major and semi-minor axes are 64 and 42 pixels, whereas the true radius is 68. The center of the converged snake differs by 4 pixels in both horizontal and vertical directions from that given in the database.

\begin{figure*}
\centering
$
\begin{array}{cccccc}
\includegraphics[width=1.00in]{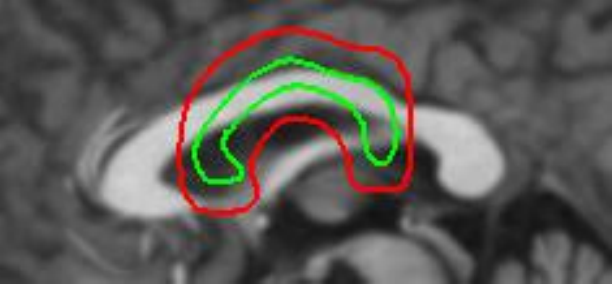} &
\includegraphics[width=1.00in]{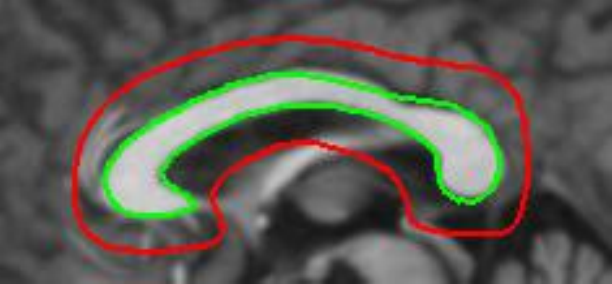} &
\includegraphics[width=1.00in]{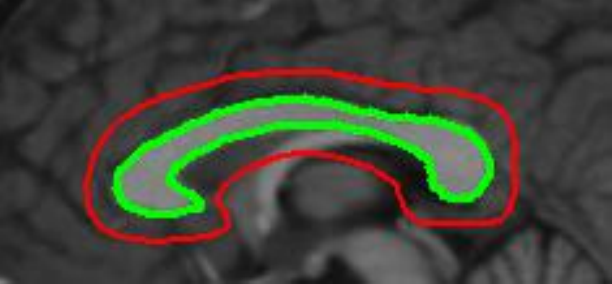} &
\includegraphics[width=1.00in]{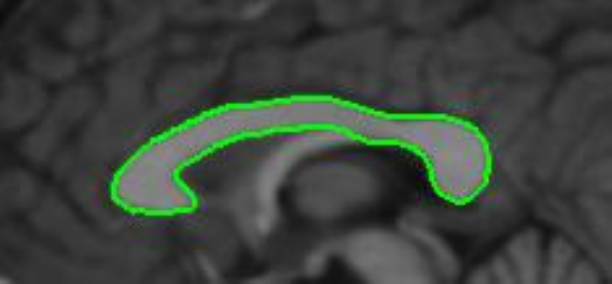} &
\includegraphics[width=1.00in]{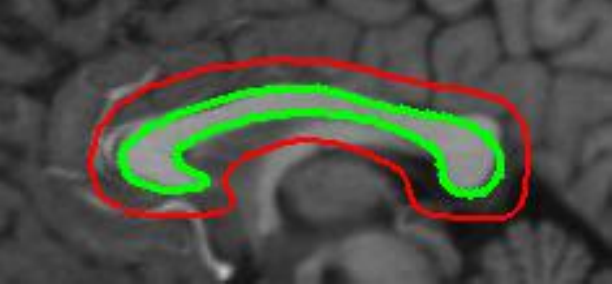} &
\includegraphics[width=1.00in]{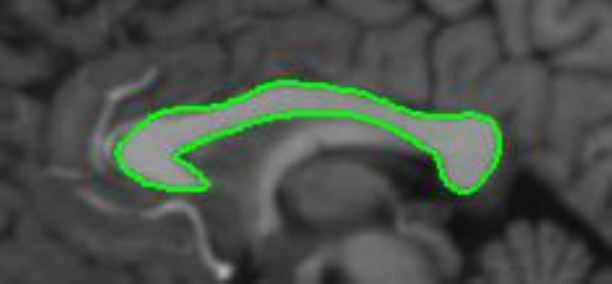} \\
\includegraphics[width=1.00in]{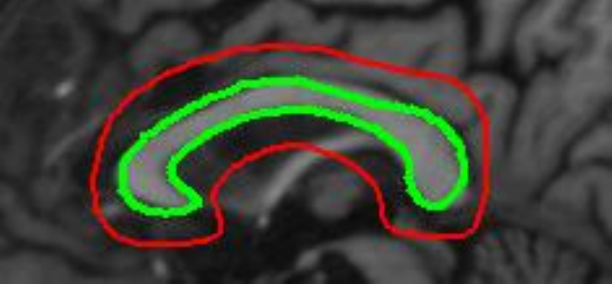} &
\includegraphics[width=1.00in]{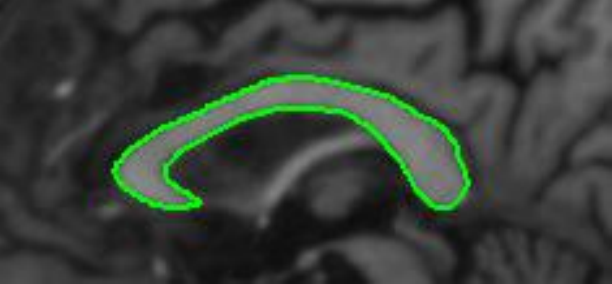} &
\includegraphics[width=1.00in]{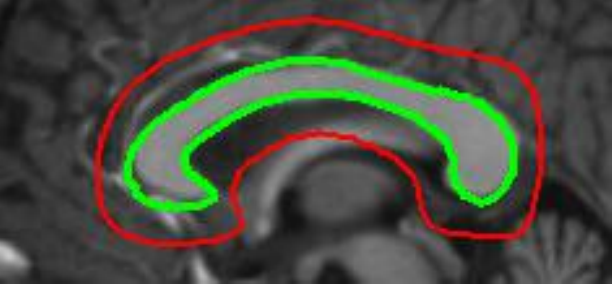} &
\includegraphics[width=1.00in]{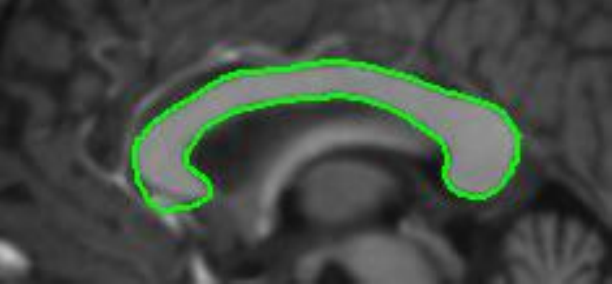} &
\includegraphics[width=1.00in]{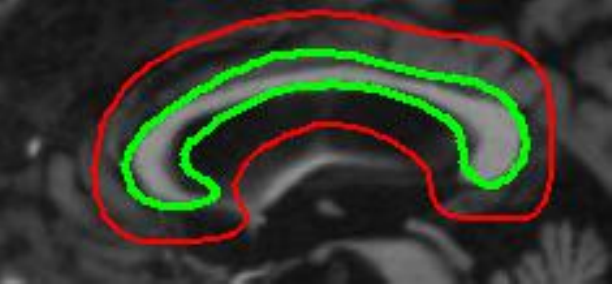} &
\includegraphics[width=1.00in]{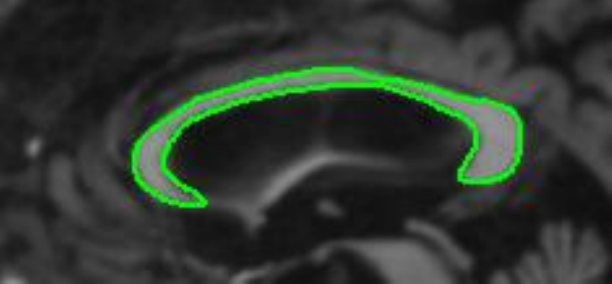} \\
\includegraphics[width=1.00in]{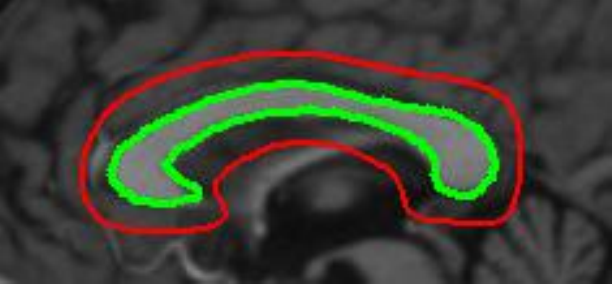} &
\includegraphics[width=1.00in]{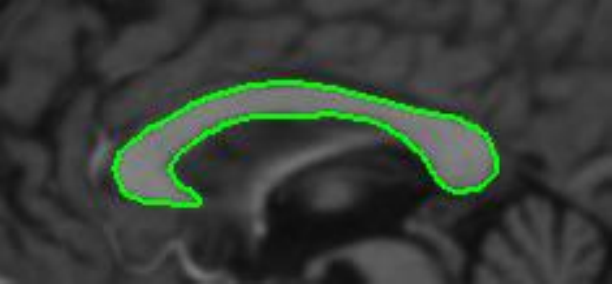} &
\includegraphics[width=1.00in]{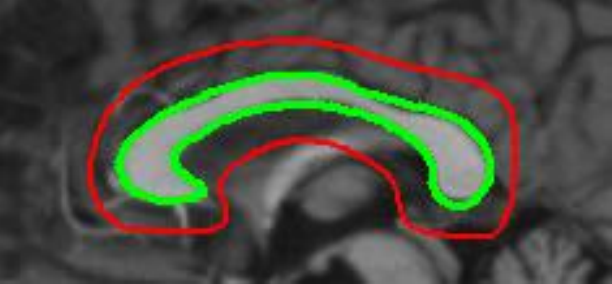} &
\includegraphics[width=1.00in]{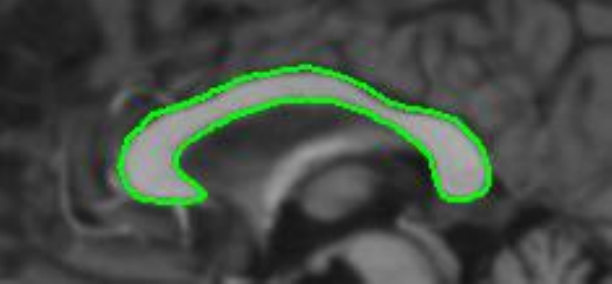} &
\includegraphics[width=1.00in]{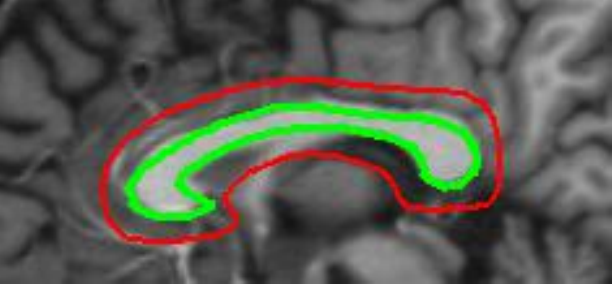} &
\includegraphics[width=1.00in]{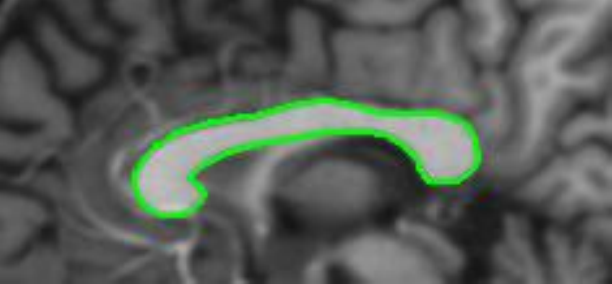} \\
\end{array}
$
\caption{Segmentation of corpus callosum images in sagittal plane T1-weighted brain magnetic resonance images taken from \cite{MarcusWPCMB07}. For all images, the same template was used and initialized in a manner similar to the way shown in the top left corner image. The converged results using the proposed shape template approach and the result obtained on subsequent refinement using snake energies\cite{Jacob} are displayed alongside.}
\label{corpus}
\end{figure*}
\begin{figure}[t]
\centering
$
\begin{array}{cc}
\includegraphics[width=1.55in, height=1.582in]{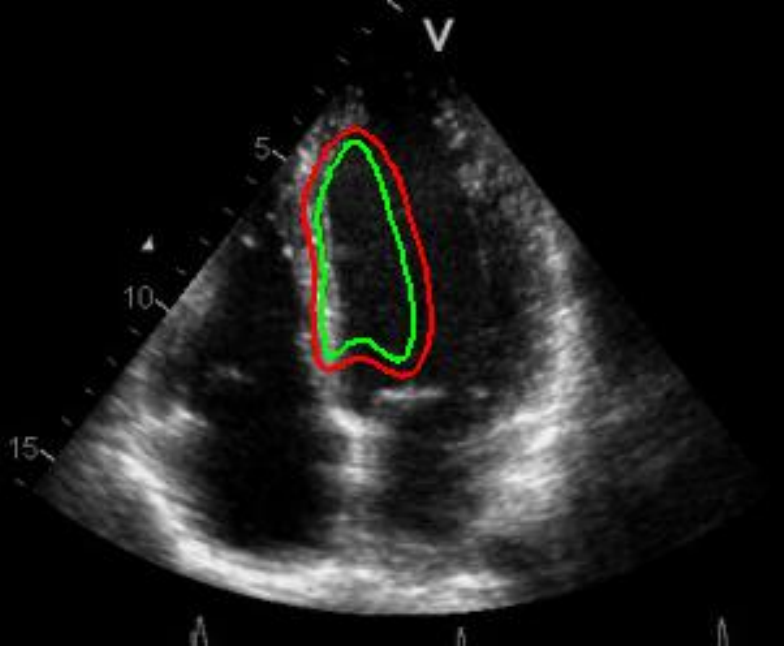} &
\includegraphics[width=1.55in, height=1.582in]{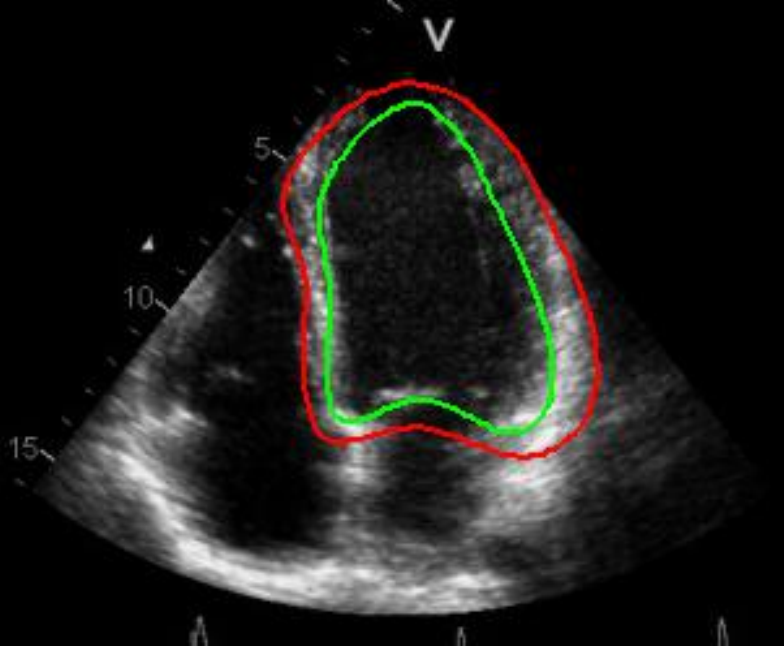} \\
(a) & (b)\\
\includegraphics[width=1.55in]{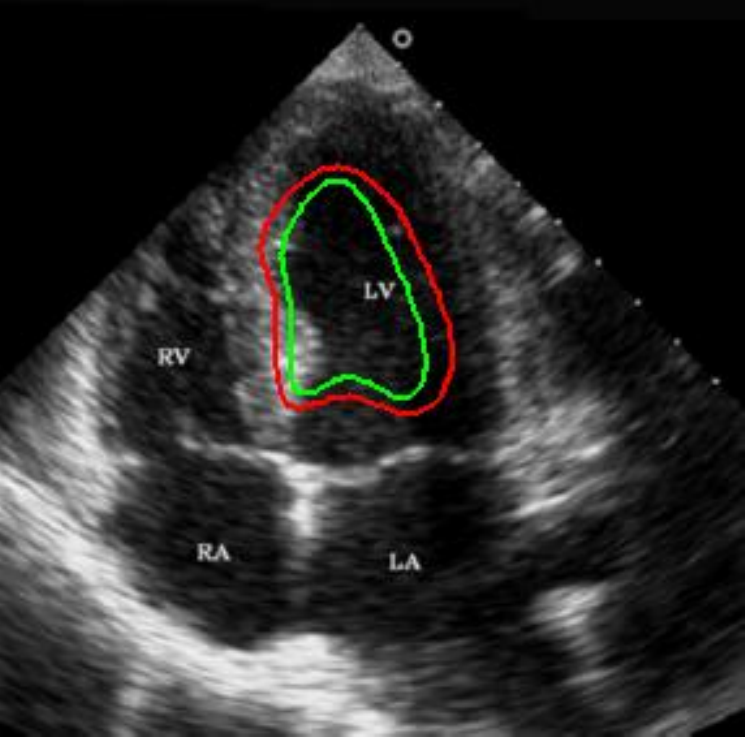} &
\includegraphics[width=1.55in]{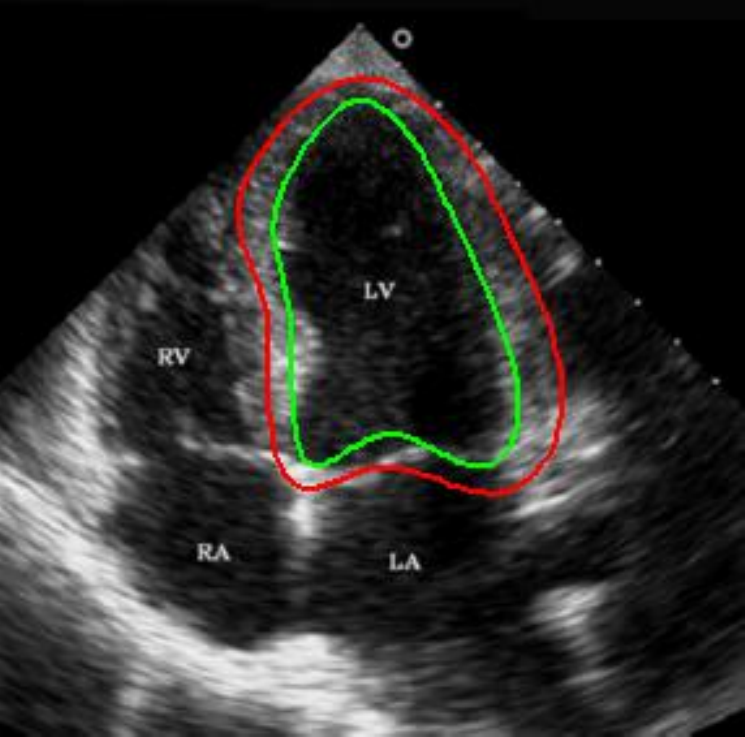}\\(c) & (d)
\end{array}
$ 
\caption{Segmentation of endocardium in B-mode ultrasound images. (a) and (c) show the initializations provided, (b) and (d) show the converged contours. Image source: (a) was taken from \cite{endocard-1} and (c) was taken from \cite{endocard-2} .}
\label{ultraendo}
\end{figure}
\begin{figure}
\centering
$
\begin{array}{cc}
\includegraphics[width=1.5in]{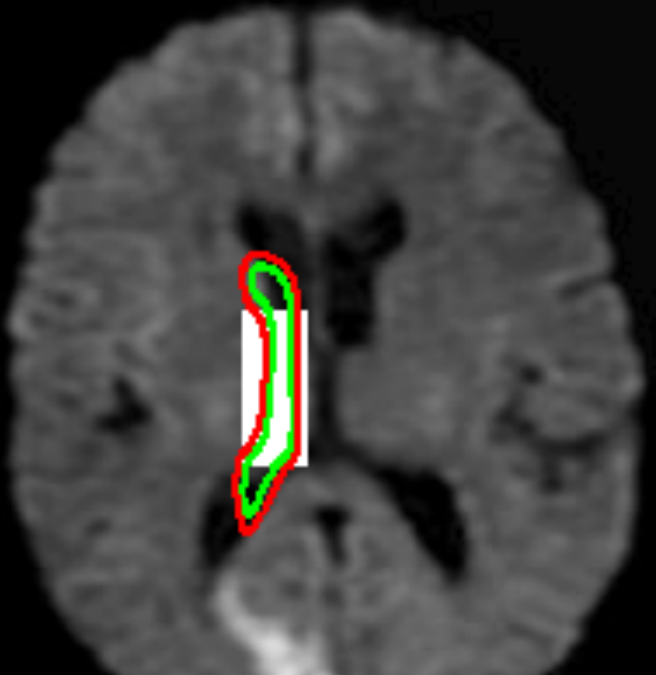} & 
\includegraphics[width=1.5in]{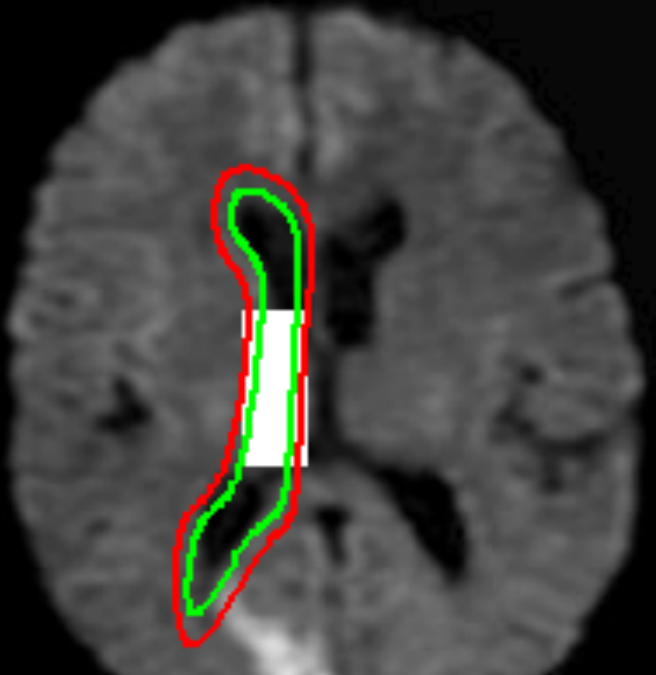} \\
(a) & (b)
\end{array}
$
\caption{Segmentation of left lateral ventricle: (a) Initialization with a template (b) converged contour. Image source: Intermountain Medical Imaging, Idaho.}
\label{mri}
\end{figure}
\subsection{Non-elliptical shape templates}
Yu et al. \cite{Tianli} reported good results for lung field and clavicle segmentation based on local differences of averages along the contour boundary. They start with an elliptical initialization and use dynamic programming to refine the contour. We also exploit the contrast between lungs and the rest of the body that X-ray imaging of the chest offers. We created a mean shape template from a set of training images using the methodology proposed in \cite{ICIP} and performed segmentation (refer Figure~\ref{Lungs}). The slight mismatch in shape between the mean template and the left lung in each image is evident from the results. \\
\indent Figure~\ref{lacto} shows segmentation on lactobacilli performed using a generic lactobacillus template. The template based snake, upon optimization provides a good fit to the cells which is apparent from Figure~\ref{lacto}(b). The result in Figure~\ref{lacto}(b) is refined further, by optimizing in the shape space of the snake. We used gradient energies \cite{Jacob} to refine the snake in Figure~\ref{lacto}(b) to obtain the result shown in ~\ref{lacto}(c).\\
\indent In Figure~\ref{corpus}, we show the results obtained for segmentation of corpus callosum in T1-weighted MR images. Many techniques on segmentation of corpus callosum are affected by the adjoining fornix structure due to the similarity in intensities. However, it can be observed from the results that the proposed snake is less affected by the fornix. We further refine results obtained using the proposed technique with gradient energies\cite{Jacob}. We display the results obtained on refinement alongside the output from the proposed shape-based approach. \\
\indent In Figure~\ref{ultraendo}, we show the results for segmentation of endocardium in B-mode ultrasound images, an imaging modality where shape prior information is popularly\cite{Hansson,Huang} used to counter the problem of broken/diffuse boundaries between the region of interest and its background. In Figure~\ref{mri}, we show an example where the template based formulation is able to overcome partial loss of signal due to occlusion and still segment the object reliably. We observe that, by incorporating prior knowledge of the shape, we can segment images with partial loss of structure and broken boundaries. \\
\begin{figure}
\centering
$
\begin{array}{cc}
\includegraphics[width=1.55in]{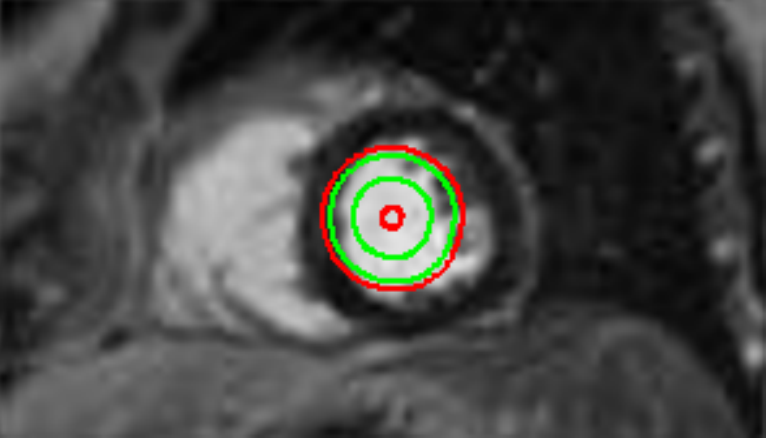}&
\includegraphics[width=1.55in]{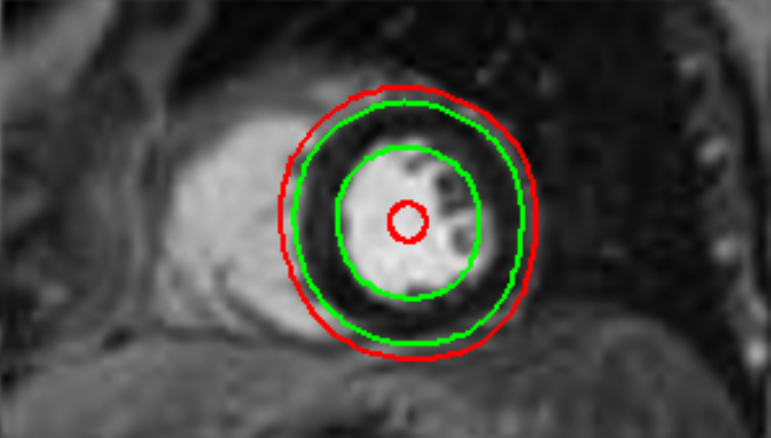}\\
\includegraphics[width=1.55in]{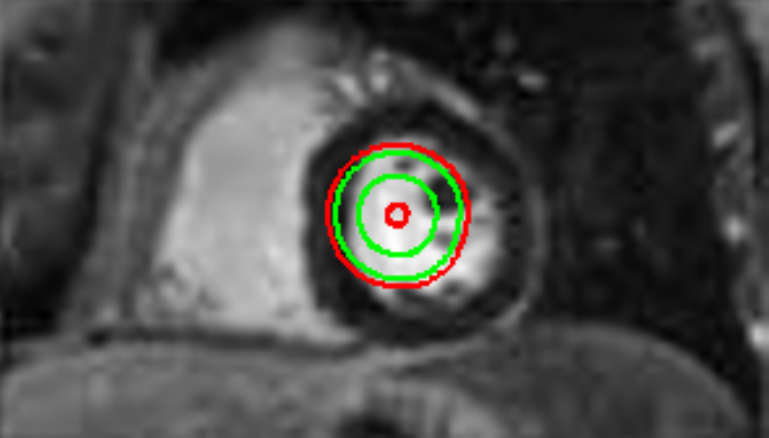}&
\includegraphics[width=1.55in]{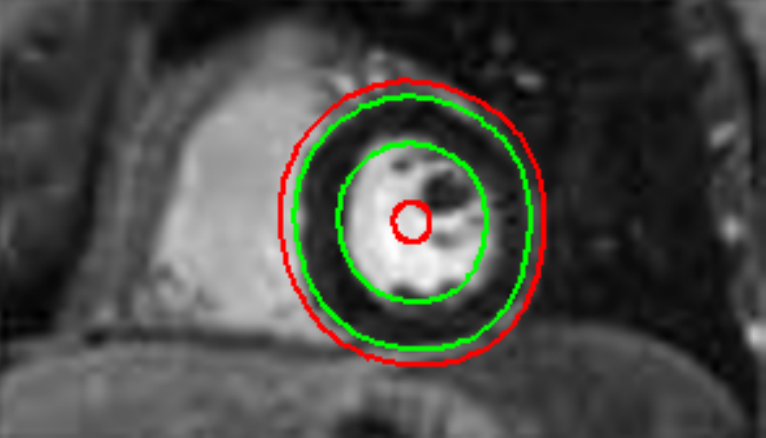}\\
\includegraphics[width=1.55in]{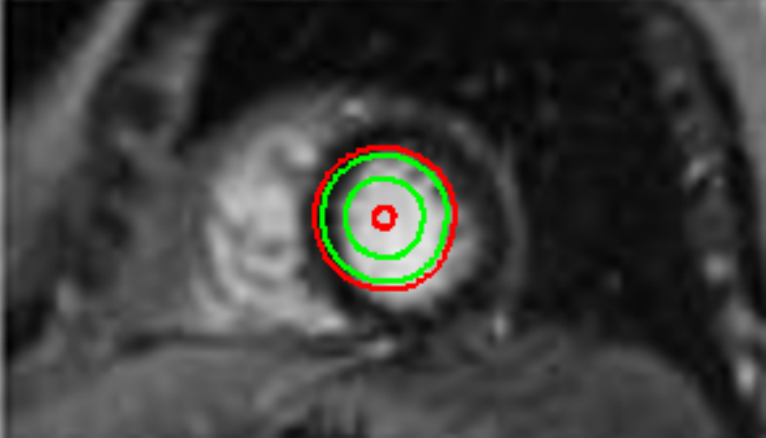}& 
\includegraphics[width=1.55in]{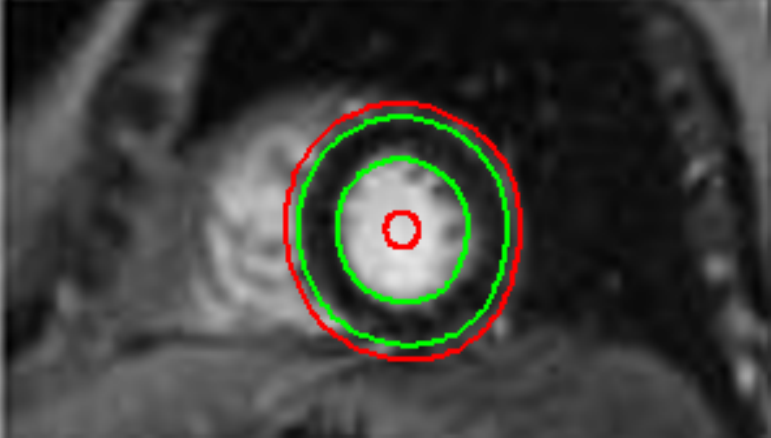}\\
\end{array}
$
\caption{Segmentation of left ventricle wall of the heart in an MR image taken from \cite{lvwalllink}. Left column shows initialization and right column shows converged result.}
\label{LV}
\end{figure}
\begin{figure}
\centering
$
\begin{array}{ccc}
\includegraphics[width=1.0in]{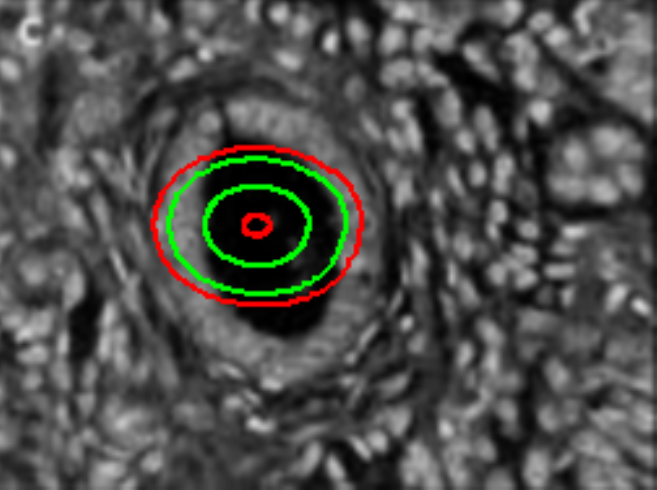} &
\includegraphics[width=1.0in]{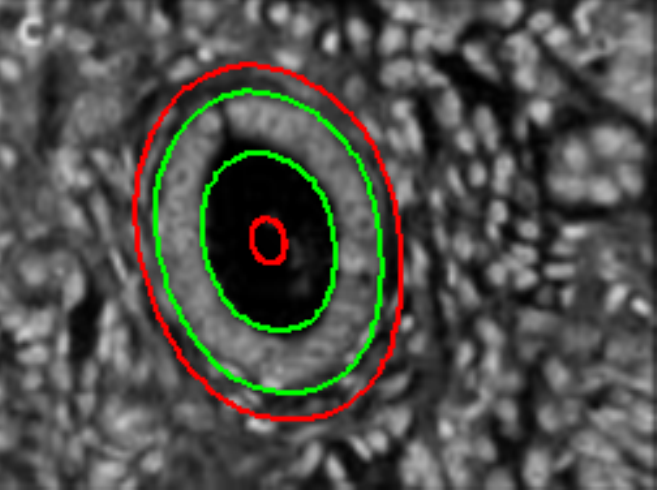}&
\includegraphics[width=1.0in]{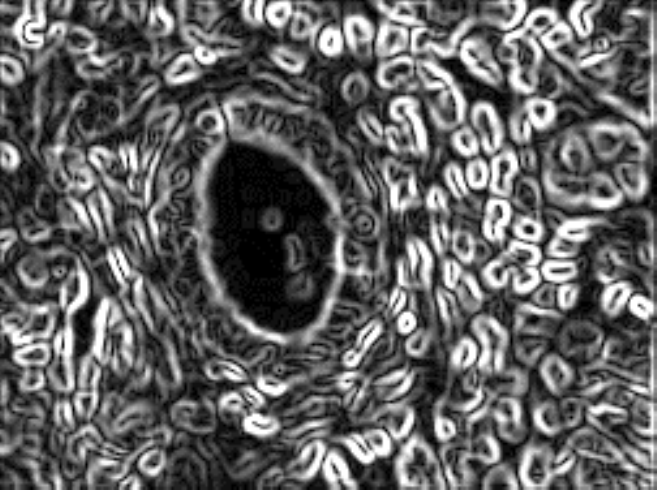}\\
(a) & (b) & (c)
\end{array}
$
\caption{(a) Initialization with ring template, (b) converged snake and (c) edge detector output of image. }
\label{tumor}
\end{figure}

\subsection{Ring templates}
\begin{figure}[t]
\centering
$
\begin{array}{c}
\includegraphics[width=3.0in]{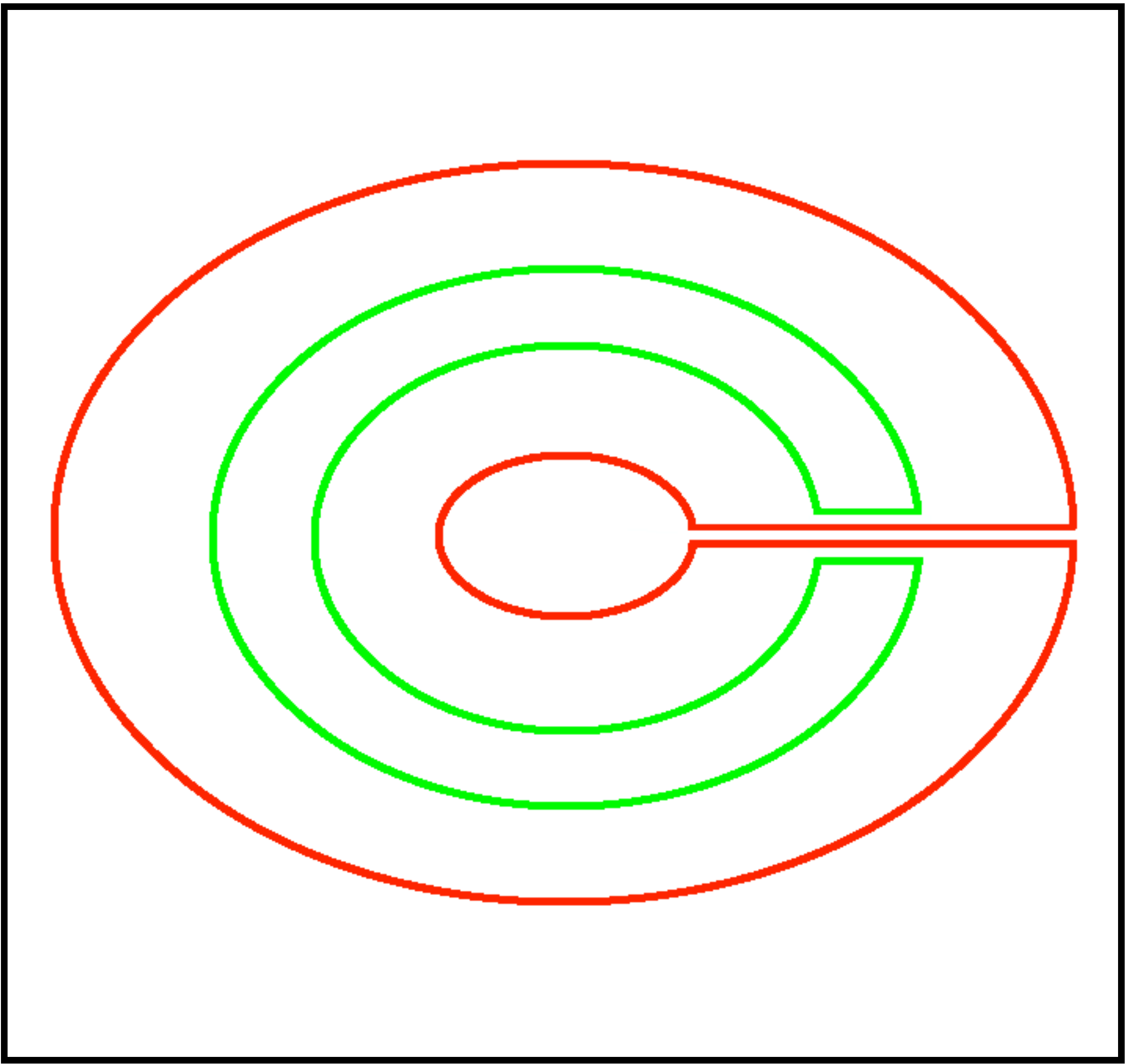} 
\end{array}
$
\caption{An elliptical ring template}
\label{RingContourExplanation}
\end{figure}
\indent We next develop templates for objects with hole(s) in their structure.  These templates are called ring templates. Figures~\ref{LV} and \ref{tumor} show examples of ring templates. The area enclosed between the red contours act as region $\Re_1$ and the area between the green contours is region $\Re_0$. Figure~\ref{RingContourExplanation} shows the construction of the ring contours. The two seemingly unconnected contours are actually one connected contour. Hence, partial derivative calculations in equation (\ref{final2}) are still valid. \\
\indent In Figure~\ref{LV}, the ring template has been used for segmenting the left ventricular (LV) wall of heart in cardiac MR images. Another example is shown in Figure~\ref{tumor}. The template used in both examples has a greater thickness in the inner ring-shaped annular region than the outer annular region. This was done to facilitate obtaining the desired object's structural information by means of the contrast (for example, between the endocardium and the LV wall) that a good initialization can offer. Hence, while designing a template, we can incorporate prior knowledge of the object and its surroundings.
\subsection{Guidelines for designing the annular region of the template}
\begin{figure*}
\centering
$
\begin{array}{cccc}
\includegraphics[width=1.5in]{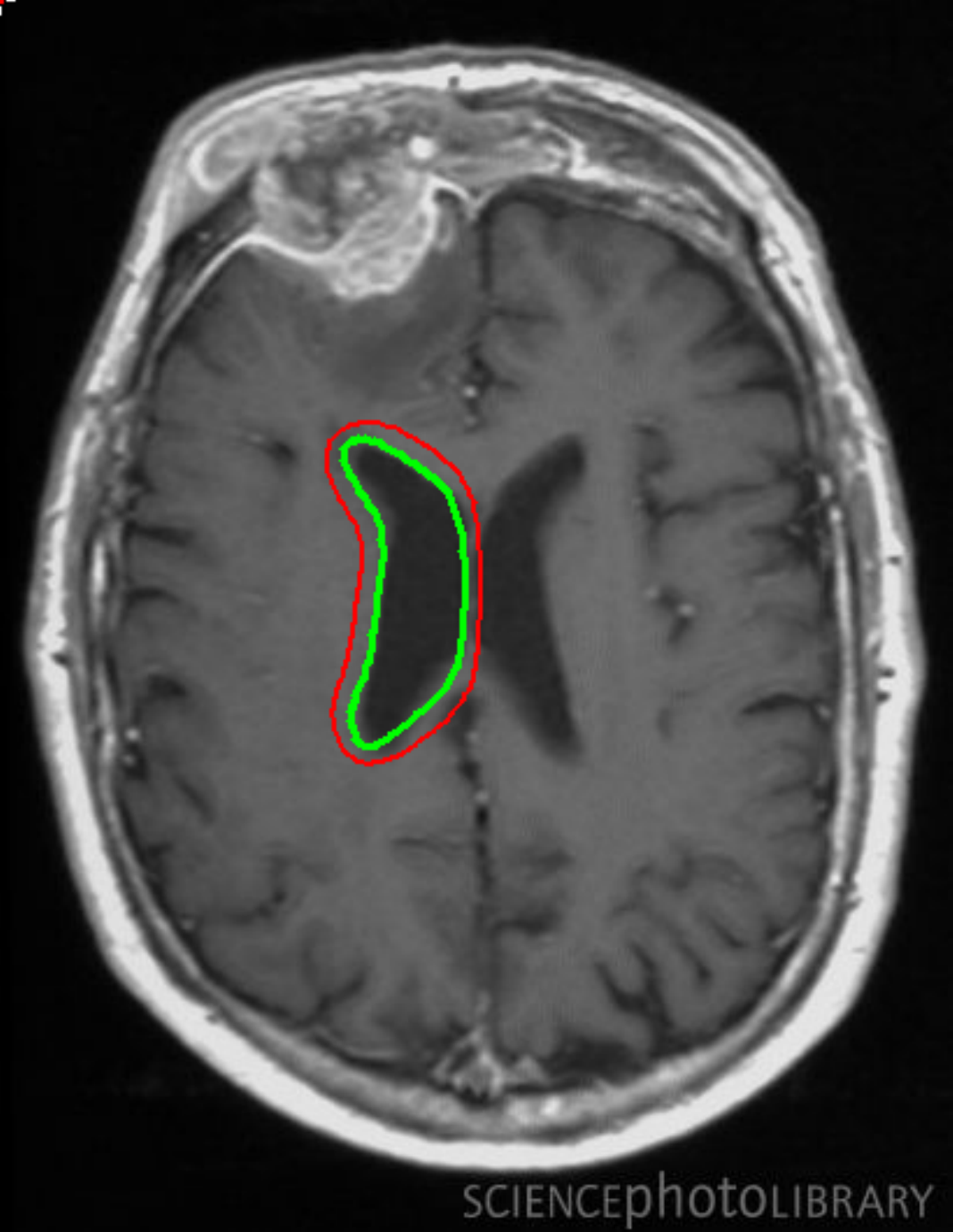}&
\includegraphics[width=1.5in]{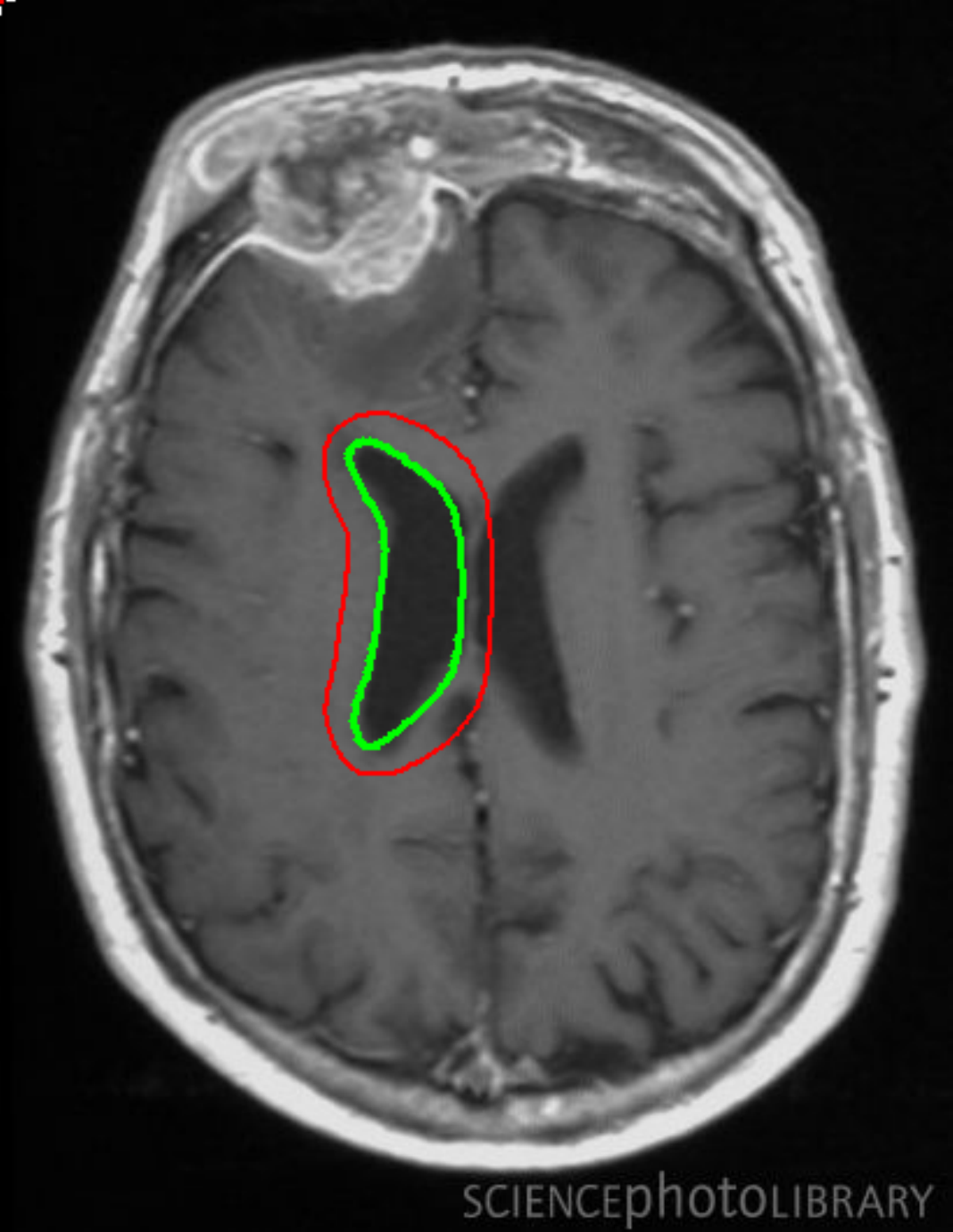}&
\includegraphics[width=1.5in]{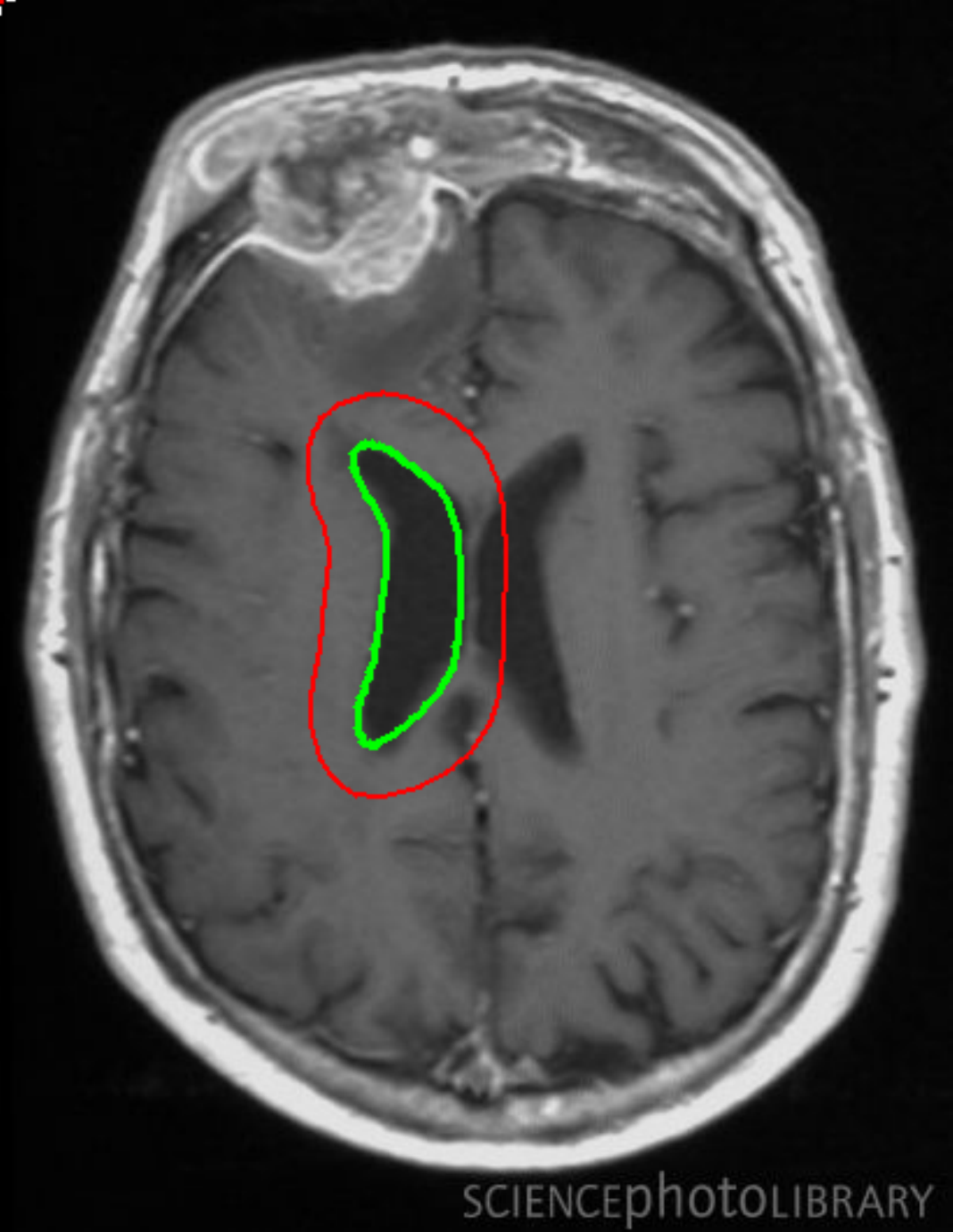}&
\includegraphics[width=1.5in]{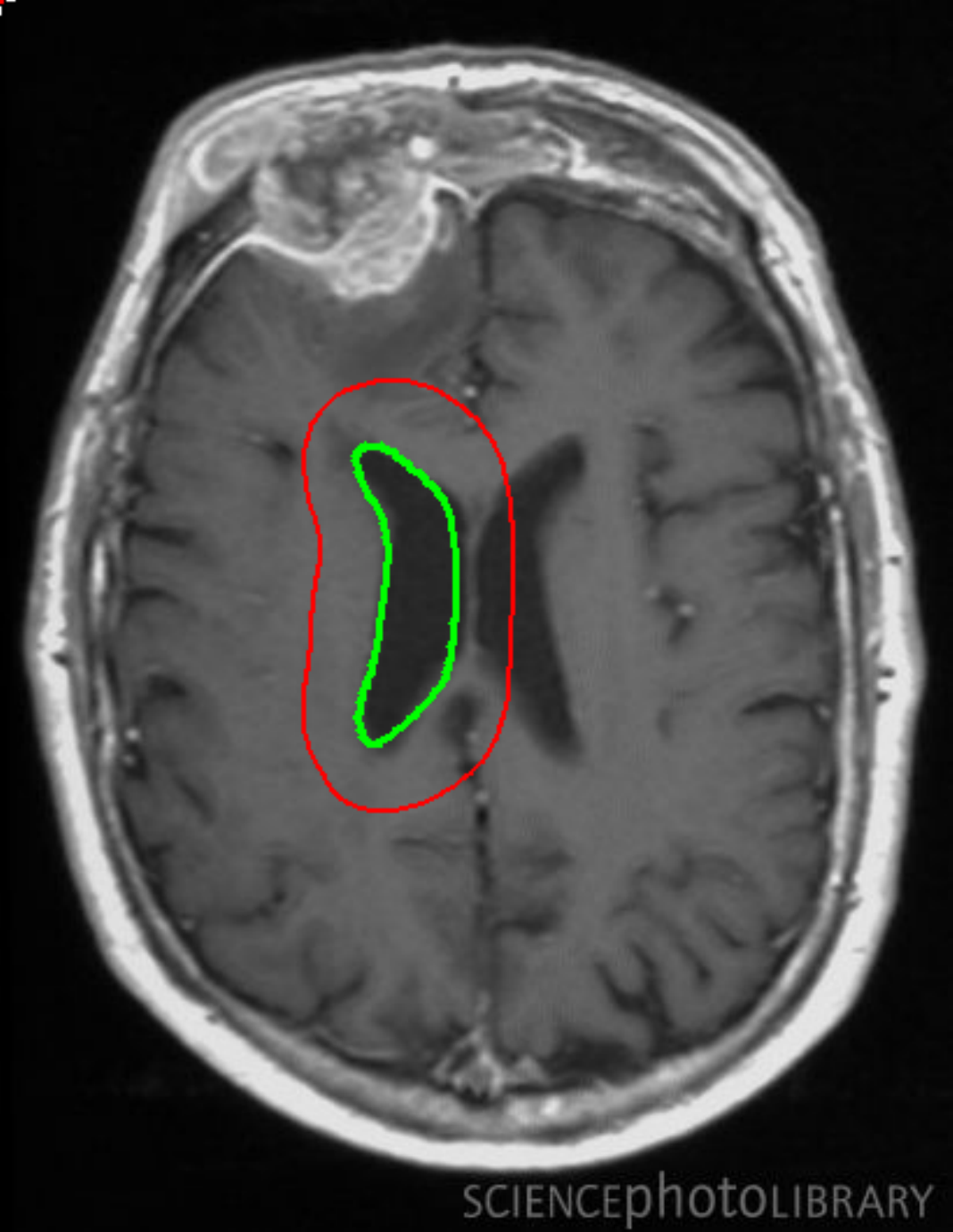}\\
(a) & (b) & (c) & (d)
\end{array}
$
\caption{(a), (b), (c), and (d) are the converged results using various templates that share the same contour for the inner template and differ only in their outer template. The annular widths of the template in (a), (b), (c), and (d) are 15, 25, 35 and 45 pixels, respectively. The MR scan is taken from \cite{anularwidth}.}
\label{annularwidthmri}
\end{figure*}

\indent We provide some key observations and infer some guidelines for designing the annular region of the template. A large annular region would make the snake less myopic but can cause overlap with other competing regions (regions having similar pixel intensity profile to that of the object) during snake optimization. The overlap may disturb the mean intensity characterization of the pixels between the inner contour and annular region of the evolving snake, thus resulting in a sub-optimal solution. On the contrary, choosing a thin annular region aimed at minimizing the overlap can cause the snake to become myopic, increase the time required for convergence or lead to a sub-optimal segmentation. Based on our experience, we suggest that areas of the annular region and inner contour of the template to be made approximately equal, a ratio between $0.5-1$ for the areas of the two regions. Hence, if the neighborhood of the object of interest is cluttered by competing regions, a smaller annular region may prove useful and vice versa when the clutter is less. Notice in Figure~\ref{tumor}, that the outermost annular region of the ring template used for segmentation has been deliberately made thin due to the large clutter of competing regions in the vicinity of the object. In Figure~\ref{retina}, the annular region of the circular template has been made sufficiently large so that the snake is not affected by blood veins in the fundus anatomy, thus avoiding convergence to a local minima from the initialization provided.\\ 
\indent When the object is sparsely surrounded by competing regions, we observed that the segmented output is not critically dependent on the width of the annular region. The energy functional is well behaved without local minima affecting its performance. Figure~\ref{annularwidthmri} illustrates a case where the left ventricle in a MR scan of the brain is segmented with templates differing in their annular widths. An inner template is chosen and the width of the annular region is changed to generate these four templates. The right ventricle has same intensity profile as that of the left ventricle. This causes the right ventricle to become a potentially competing region, but the results indicate identical segmentation outputs despite the outer contour overlapping with the right ventricle to varying degrees in Figures~\ref{annularwidthmri}(a), (b), (c), and (d) respectively. The experiment is repeated with multiple initializations and the consistency in final output is observed.

\section{Discussion and Conclusion}
\indent We proposed shape template based active contour formalism for image segmentation. Shape specificity is directly incorporated in these snakes  leading to constant number of parameters to be calculated. The parameters are independent of the shape of the template and hence, the complexity of algorithm. We proposed a contrast based energy function for the active contour formalism. The proposed energy function is stable and optimal under certain conditions. The energy function is optimized with respect to the RAT parameters. We used gradient descent for optimization and developed an efficient approach for calculating partial derivatives by invoking Green's theorem from vector calculus.  \\
\indent We evaluated the performance of the proposed algorithm systematically for various shape templates, initializations and Poisson/Gaussian noise conditions. The proposed formulation gave satisfactory results on templates ranging from simple shapes such as circles and ellipses to concave shapes and rings. The technique is also able to segment objects with partial occlusion, thanks to a priori knowledge incorporated in the template. We also discussed the effect of the thickness of annular region on the segmentation output. \\
\indent We have developed an ImageJ plugin based on the technique discussed in this paper \cite{pluginlink}. We have provided a library of shape templates mostly related to medical applications. The user also has an option to create a new template for segmentation. \\ 
\indent The proposed method primarily aimed to segment objects with a given shape prior. When training data is available, a shape model can automatically be learnt from it. The formalism proposed in this paper can incorporate such a shape model. Taking cue from the vast literature in this area, we proposed a segmentation scheme \cite{ICIP}, which alternates between shape template based snakes proposed in this paper and shape unconstrained snakes. The alternation is done efficiently using the shape model. Riesz basis property of splines is used to decrease the computational complexity. We validated the performance on Corpus Callosum segmentation in T-1 weighted brain MR images (cf. figure~\ref{corpus}) and observed good accuracy in the segmentation results. The technique was also observed to converge quickly due to fast computations of the proposed shape template based snakes.

\section*{Acknowledgment}
\small
The authors would like to thank Naren Nallapareddy for his help in generating some of the images and for proof-reading the manuscript. CSS would also like to thank Michael Unser and Philippe Th\'evenaz for introducing him to splines, snakes, and snakuscules. This work is sponsored by the Department of Science and Technology (DST) --- Intensive Research in High-Priority Areas (IRHPA) of the Government of India (Project code: DSTO$-$943).

\vspace*{-3\baselineskip}


\begin{thebibliography}{99}
\bibitem{Kass} M.~Kass, A.~Witkin, and D.~Terzopoulos, ``Snakes: Active contour models," \emph{Intl. J. on Comp. Vision}, vol.~1, pp.~321$-$331, 1988. 
\bibitem{GVF} C.~Xu, and J.~L.~Prince, ``Snakes, shapes, and gradient vector flow," \emph{IEEE Trans. on Image Process.}, vol.~7, no.~3, pp.~359$-$369, March 1988.
\bibitem{Pentland} A.~Pentland, and S.~Sclaroff, ``Closed-form solutions for physically based shape modeling and recognition," \emph{IEEE Trans. on Pattern Analysis and Machine Intelligence}, vol.~13, no.~7, pp.~715$-$729, January 1995. 
\bibitem{Brigger} P.~Brigger, J.~Hoeg, and M.~Unser, ``B-spline snakes: A flexible tool for parametric contour detection," \emph{IEEE Trans. on Image Process.}, vol.~9, no.~9, pp.~1484$-$1496, September 2000.
\bibitem{Staib} L.~H.~Staib, and J.~S.~Duncan, ``Boundary finding with parametrically deformable models," \emph{IEEE Trans. on Pattern Analysis and Machine Intelligence}, vol.~14, no.~11, pp.~1061$-$1075, 1992.
\bibitem{Cootes} T.~F.~Cootes, C.~J.~Taylor, D.~H.~Cooper, and J.~Graham, ``Active shape models--their training and application," \emph{Comp. Vision and Image Understanding}, vol.~61, no.~1, pp.~38$-$59, January 1995.
\bibitem{Blake} A.~Blake, R~Curwen, and A.~Zisserman, ``A framework for spatio-temporal control in the tracking of visual contours," \emph{Intl. J. on Comp. Vision}, vol.~11, no.~2, pp.~127$-$145, 1993.
\bibitem{Paragois} M.~Paragios, ``A level set approach for shape-driven segmentation and tracking of the left ventricle," \emph{IEEE Trans. on Medi. Imaging}, vol.~22, no.~6, pp.~773$-$776, 2003.
\bibitem{tsai} A.~Tsai~et~al., ``A shape-based approach to the segmentation of medical imagery using level sets," \emph{IEEE Trans. on Medi. Imaging}, vol.~22, no.~2, pp.~137$-$154, 2003.
\bibitem{Levelsetprior} T.~Chan, and W.~Zhu, ``Level set based shape prior segmentation," \emph{in Proc. CVPR 2005 IEEE Comp. Society Conf.}, vol.~2, pp.~1164$-$1170, 2005.
\bibitem{Levelprior} Y.~Chen~et~al., ``Using prior shapes in geometric active contours in a variational framework," \emph{Intl. J. Comp. Vision}, vol.~50, no.~3, pp.~315$-$328, 2002. 
\bibitem{Cremerslevelprior} D.~Cremers, and S.~Soatto, ``A pseudo-distance for shape priors in level set segmentation," \emph{in Proc. IEEE 11th Intl. Conf. on Comp. Vision}, 2007.
\bibitem{snakuscule} P.~Th\'evenaz, and M.~Unser, ``Snakuscules," \emph{IEEE Trans. on Image Process.}, vol.~17, no.~4, pp.~585$-$593, April 2008. 
\bibitem{ovuscule} P.~Th\'evenaz, R.~Delgado-Gonzalo, and M.~Unser, ``The Ovuscule," \emph{IEEE Trans. on Pattern Analysis and Machine Intelligence}, vol.~33, no.~2, pp.~382$-$393, February 2011.
\bibitem{Adithya} A.~K.~Pediredla, and C.~S.~Seelamantula, ``Active-contour-based automated image quantitation techniques for Western Blot Analysis," \emph{in Proc. 7th Intl. Symposium on Image and Signal Process. and Analysis (ISPA)}, pp.~331$-$336, September 2011. 
\bibitem{Adithya1} A.~K.~Pediredla, and C.~S.~Seelamantula, ``A unified approach for optimization of snakuscules and ovuscules," \emph{in Proc. IEEE Intl. Conf. on Acoustic, Speech, and Signal Process.}, March 2012. 
\bibitem{Werlberger} M.~Werlberger, T.~Pock, M.~Unger, and H.~Bischof, ``A variational model for interactive shape prior segmentation and real-time tracking," \emph{Scale Space and Variational Methods in Computer Vision}, pp.~200$-$211, 2009.
\bibitem{Cremers} T.~Schoenemann, and D.~Cremers, ``Globally optimal image segmentation with an elastic shape prior," \emph{in Proc. IEEE 11th Intl. Conf. on Comp. Vision}, 2007.
\bibitem{Chen} S.~Chen, D.~Cremers, and R.~J.~Radke, ``Image segmentation with one shape prior-A template-based formulation," \emph{Image and Vision Computing}, 2012.
\bibitem{duda} R.~Duda, P.~Hart, and D.~Stork, \emph{Pattern Classification}. 2nd ed. John Wiley and Sons, Inc, 2001. 
\bibitem{PerfectFit} M.~Unser, ``Splines: A perfect for for signal and image processing," \emph{IEEE Signal Process. Magazine}, vol.~16, no.~6, pp.~22$-$38, November 1999.
\bibitem{Jacob} M.~Jacob, T.~Blu, and M.~Unser, ``Efficient energies and algorithms for parametric snakes," \emph{IEEE Trans. on Image Process.}, vol.~13, no.~9, pp.~1231$-$1244, September 2004. 
\bibitem{Chakraborthy} A.~Chakraborthy, L.~H.~Staib, and J.~S.~Duncan, ``Deformable boundary finding in medical images by integrating gradient and region information," \emph{IEEE Trans. on Medi. Imaging}, vol.~15, pp.~859$-$870, June 1996.
\bibitem{VeseChan} T.~F.~Chan, and L.~A.~Vese, ``Active Contours without edges," \emph{IEEE Trans. on Image Process.}, vol.~10, no.~2, pp.~266$-$277, February 2001.
\bibitem{Chesnaud} C.~Chesnaud, P.~Refregier, and V.~Boulet, ``Statistical region snake-based segmentation adapted to different physical noise models," \emph{IEEE Trans. on Pattern Analysis and Machine Intelligence}, vol.~21, no.~11, pp.~1145$-$1157, February 1995.
\bibitem{Yezzi} J.~A.~Yezzi, A.~Tsai, and A.~Willsky, ``A fully global approach to image segmentation via coupled curve evolution equations," \emph{J. on Vision, Comm. and Image Repr.}, vol.~13, no.~1, pp.~195$-$216, 2002.
\bibitem{Mumford} D.~Mumford, and J.~Shah, ``Optimal approximation by piecewise smooth functions and associated variational problems," \emph{Comm. on Pure and Applied Mathematics}, vol.~42, pp.~577$-$685, 1989. 
\bibitem{kay} S.~Kay, \emph{Statistical Signal Processing: Estimation Theory}. EnglewoodCliffs, NJ:Prentice Hall, 1993.
\bibitem{Shepp} L.~Shepp, and B.~Logan, ``The Fourier reconstruction of a head section," \emph{IEEE Transaction on Nuclear Science}, vol.~21, pp.~21$-$43, June 1974.
\bibitem{gonzalezwoods} R.~C.~Gonzalez, and R.~E.~Woods, \emph{Digital Image Processing}. 2nd ed. Englewood Cliffs, NJ: Prentice Hall, 2002.
\bibitem{florian} F.~Luisier, C.~Vonesch, T.~Blu, and M.~Unser, ``Fast interscale wavelet denoising of Poisson-corrupted images," \emph{Signal Process.}, vol.~90, no.~2, pp.~415$-$427, Feb. 2010. 
\bibitem{fundus} \emph{http://www.parl.clemson.edu/stare/images2.htm}
\bibitem{minimias} J.~Suckling~et~al., ``The Mammographic Image Analysis Society Digital Mammogram Database," \emph{Exerpta Medica, Intl. Congress Series}, pp.~375$-$378, 1994.
\bibitem{lactobacillus} \emph{http://naturalabundancehealth.blogspot.com}
\bibitem{Tianli} T.~Yu, J. Luo, A. Singhal, and N. Ahuja, ``Shape regularized active contour based on dynamic programming for anatomical structure segmentation," \emph{in Proc. SPIE 5747, Medi. Imaging: Image Process.}, 2005. 
\bibitem{ICIP} J.~K.~Mogali, N.~Nallapareddy, C.~S.~Seelamantula, and M.~Unser, ``A shape-template based two-stage corpus callosum segmentation technique for sagittal plane T1-weighted brain magnetic resonance images," \emph{in Proc. IEEE Intl. Conf. Image Process. (ICIP)}, 2013.
\bibitem{MarcusWPCMB07} D.~S.~Marcus, T.~H.~Wang, J.~Parker, J.~G.~Csernansky, J.~C.~Morris, and R.~L.~Buckner, ``Open access series of imaging studies (OASIS): Cross-sectional MRI data in young, middle aged, nondemented, and demented older adults," {\it J. Cognitive Neuroscience}, vol. 19, no. 9, pp. 1498-1507, 2007.
\bibitem{endocard-1} \emph{http://folk.ntnu.no/stoylen/strainrate/Ultrasound/reject.jpg} 
\bibitem{endocard-2} \emph{http://ehjcimaging.oxfordjournals.org/content/9/3/401/F1.expansion}
\bibitem{Hansson} M.~Hansson, K.~Fundana, S.~S.~Brandt, and P.~Gudmundsson, ``Convex spatio-temporal segmentation of the endocardium in ultrasound data using distribution and shape priors" \emph{IEEE International Symposium on Biomedical Imaging: From Nano to Macro}, pp. 626$-$629, 2011.
\bibitem{Huang} Y.~Chen, W.~Guo, F.~Huang, D.~Wilson, and E.~A.~Geiser, ``Using prior shape and points in medical image segmentation" \emph{Energy Minimization Methods in Computer Vision and Pattern Recognition, Proceedings}, vol.~2683, pp. 291$-$305, 2003. 
\bibitem{lvwalllink} \emph{http://clinical.netforum.healthcare.philips.com/global/Explore/Case-Studies/MRI/Asymptomatic-patient-with-chronic-total-occlusion-of-RCA-m} 
\bibitem{pluginlink} \emph{http://sites.google.com/site/chandrasekharseelamantula/projects}.
\bibitem{anularwidth} \emph{http://www.sciencephoto.com/media/259763/view}
\end{thebibliography}
\end{document}